%% file: main.tex
\documentclass{article}
\PassOptionsToPackage{numbers, compress}{natbib}
\usepackage[preprint,eandd]{preprint_style}
\usepackage[skins,breakable]{tcolorbox}
\usepackage{enumitem}
\usepackage[utf8]{inputenc}
\usepackage[T1]{fontenc}
\usepackage{hyperref}
\usepackage{url}
\usepackage{booktabs}
\usepackage{amsfonts}
\usepackage{nicefrac}
\usepackage{microtype}
\usepackage{xcolor}
\usepackage{graphicx}
\usepackage{colortbl}
\usepackage{longtable}
\usepackage{tabularx}

\title{EEG Benchmarking Needs a Task Specification Layer: NeuroDoc for Rulebook-Guided, Executable Benchmark Construction}

\author{%
\normalfont
\begin{tabular}{c}
Chengxuan Qin$^{1,2,*}$, Zhige Chen$^{4,*}$, Shu Peng$^{4,*}$, Rui Yang$^{1,2}$, Jiping Cui$^{1,2,\dagger}$,\\
Yikai Dong$^{1,2,\dagger}$, Jun Li$^{1,2,\dagger}$, Liu Peng$^{1,2,\dagger}$, Zhida Shang$^{1,3,\dagger}$,\\
Mingze Tang$^{4,\dagger}$, Kay Chen Tan $^{4}$, Jibin Wu$^{4}$\\[0.5em]
{\small $^1$School of Advanced Technology, Xi'an Jiaotong-Liverpool University}\\
{\small $^2$School of Electrical Engineering, Electronics and Computer Science, University of Liverpool}\\
{\small $^3$School of Computer Science and Informatics, University of Liverpool}\\
{\small $^4$Department of Data Science and Artificial Intelligence, Hong Kong Polytechnic University}\\[0.2em]
{\small $^*$\,Equal contribution\quad $^\dagger$\,Equal contribution}\\[0.4em]
\parbox{0.9\textwidth}{\centering\footnotesize
  \texttt{c.qin8@liverpool.ac.uk},
  \texttt{zhige.chen@connect.polyu.hk},
  \texttt{shu.peng@polyu.edu.hk},
  \texttt{r.yang@xjtlu.edu.cn},
  \texttt{jiping.cui24@student.xjtlu.edu.cn},
  \texttt{yikai.dong24@student.xjtlu.edu.cn},
  \texttt{jun.li24@student.xjtlu.edu.cn},
  \texttt{liu.peng24@student.xjtlu.edu.cn},
  \texttt{zhida.shang25@student.xjtlu.edu.cn},
  \texttt{mason.tang@connect.polyu.hk},
  \texttt{kaychen.tan@polyu.edu.hk},
  \texttt{jibin.wu@polyu.edu.hk}
}%
\end{tabular}%
}

\begin{document}

\maketitle

\begin{abstract}
Electroencephalography (EEG) foundation models increasingly rely on multi-dataset training and evaluation, yet public EEG datasets still lack a shared task specification layer that can turn heterogeneous recordings into reusable benchmark units. Existing standards organize files, metadata, and provenance, but they do not specify EEG tasks under a common language and rulebook, leaving critical task semantics scattered across papers, code, and manual interpretation. We investigate whether heterogeneous public EEG datasets can be standardized through a structured task specification language paired with a shared rulebook. Our methodology represents each benchmark entry as a task document synchronized with an executable task kernel, with the rulebook defining task fields, evidence requirements, document-kernel alignment, review states, and machine-checkable constraints. Using this methodology, we release a community-reviewed EEG benchmark corpus centered on 53 completed and reviewed entries with 245 task definitions spanning diverse paradigms, and we introduce NeuroDoc and NeuroAudit as the operational support layer for rulebook-guided drafting, upgrading, review, amendment, and release management. We further examine whether the resulting benchmark units can be instantiated in a shared downstream setting across four EEG foundation model backbones, providing execution-based evidence for reusable, auditable, and executable EEG benchmarking infrastructure.
\end{abstract}

\section{Introduction}

Electroencephalography (EEG) provides a non-invasive way to monitor brain activity and has been widely used to study cognitive processes, behavioral states, and neurological conditions. Because EEG data are collected under diverse experimental paradigms for different scientific and clinical purposes, public EEG datasets have emerged with substantial variation in data acquisition, annotation, and organization conventions \cite{labram-iclr-2024, brainomni-2025, reve-2025, eegunity-tnsre-2025}. The heterogeneity has become increasingly consequential as modern EEG foundation models are trained on aggregated multi-dataset corpora and evaluated for transfer across datasets, subjects, recording setups, and downstream tasks.

Early self-supervised representation learning systems such as BENDR \cite{bendr-fnhum-2021} showed that useful EEG representations could be learned from large corpora spanning beyond a single downstream benchmark. More recent models (such as LaBraM \cite{labram-iclr-2024}, CBraMod \cite{cbramod-iclr-2025} and REVE \cite{reve-2025}) push this trend further toward large-scale aggregated pre-training. Several of these works rely on broad dataset aggregation: LaBraM reports pre-training data drawn from around 20 datasets \cite{labram-iclr-2024}, and REVE reports pre-training across 92 datasets \cite{reve-2025}. As large-scale dataset aggregation becomes central to EEG foundation model design, cross-dataset reuse is becoming an increasingly important practical requirement.

At the same time, emerging evidence \cite{eegfm-worth-it-iclr-2026} suggests that performance gains from EEG foundation models depend strongly on dataset choice, evaluation protocol, and downstream task definition. Model architecture therefore remains crucial, but it is not the only factor limiting progress. In practice, researchers who want to train, adapt, or evaluate models across many public EEG datasets still face repeated and labor-intensive task-definition work. This exposes a deeper question for the field: can EEG tasks be specified under a shared language and rulebook strongly enough to support reusable benchmarking across heterogeneous datasets?

Existing infrastructure addresses important adjacent parts of the problem, while task specification itself remains underspecified. BIDS \cite{bids-sdata-2016} and EEG-BIDS \cite{eeg-bids-sdata-2019} standardize directory organization and modality-specific metadata. Generic documentation frameworks such as Croissant \cite{croissant-2024} improve transparency, provenance tracking, and machine-readable dataset description. What remains missing is a shared specification layer for EEG tasks: a consistent way to state epoch derivation, label definition, evidence source, and executable realization for each dataset. In practice, this information is still distributed across dataset papers, supplementary notes, raw directory inspection, and code fragments, so cross-dataset EEG research continues to depend on substantial undocumented interpretation even when the raw data are publicly available.

\begin{figure}[b]
\centering
\includegraphics[width=\textwidth]{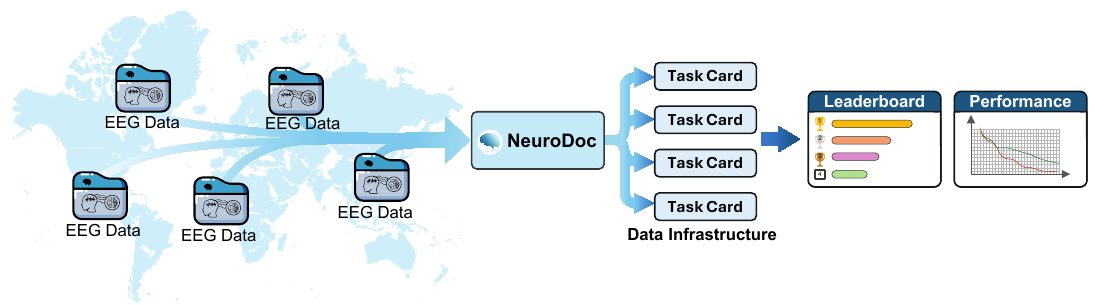}
\caption{Project vision. A rulebook-guided task specification layer converts heterogeneous EEG datasets into reusable benchmark units with auditable task semantics and executable kernels, enabling shared downstream evaluation without repackaging raw EEG recordings.}
\label{fig:section1_vision}
\end{figure}

That dependence on manual interpretation is not only a technical inconvenience. In EEG, raw recordings generally cannot be freely repackaged and redistributed because secondary sharing is constrained by research ethics, participant consent, privacy regulations such as GDPR and HIPAA, and repository-specific licensing or copyright conditions associated with dataset release \cite{open-brain-consent-hbm-2021, eeg-privacy-tnsre-2019}. This makes task specification a first-class scientific bottleneck: if the benchmark unit cannot be defined clearly at the task level, large-scale reuse remains fragile even when raw data are accessible. Figure~\ref{fig:section1_vision} illustrates the resulting design goal of this project.

Recent large language model (LLM)-based documentation systems can generate cards, metadata, or benchmark descriptions, but they are typically designed for generic documentation and do not incorporate EEG-specific semantic constraints or domain review \cite{cardgen-naacl-2024, dca-bench-kdd-2025,pre-meta-bioinformatics-2025, auto-benchmarkcard-arxiv-2025}. In EEG, task definitions often depend on paradigm-specific event interpretation, label construction, epoch configuration, and derived supervision routes. A useful solution therefore requires a structured task language, explicit domain rules, executable task logic, and evidence tracing.

In this paper, we study whether a task specification language and rulebook can make heterogeneous public EEG datasets explicit, executable, and auditable enough for reusable benchmarking. Therefore, we introduce an \emph{entry}-based representation in which each benchmark entry consists of a human-readable \emph{task document} and a dataset-specific \emph{task kernel}. The task document captures task semantics, assumptions, field-structured specifications, and supporting evidence, and the task kernel provides the executable logic that operationalizes the specification in a unified EEG processing pipeline.

To operationalize this methodology, we introduce \emph{NeuroDoc}, a rulebook-guided mechanism for preparing and updating \emph{entries}---task documents and paired task kernels---at scale, and \emph{NeuroAudit}, a reviewer-facing system for community review, amendment, and release-time status control. These tools support a growing benchmark database spanning public releases from repositories such as OpenNeuro \cite{openneuro-elife-2021}, PhysioNet \cite{physionet-circ-2000}, and other sources. The analysis in this paper centers on 53 completed and reviewed benchmark entries.

Overall, our contributions are as follows:
\begin{itemize}
\item \textbf{Methodology.} We propose a \emph{rulebook-guided EEG task specification language and methodology} for heterogeneous public EEG datasets. This specification layer defines task fields, evidence grounding, document-kernel synchronization, review states, and machine-checkable constraints for benchmark construction.

\item \textbf{Resource.} We present a \emph{community-reviewed EEG benchmark corpus} centered on 53 completed and reviewed benchmark entries and 245 task definitions. The released subset covers diverse EEG paradigms and task-interface patterns, and each entry is released as a reusable benchmark unit with paired task specification and executable kernel.

\item \textbf{Toolchain.} We introduce \emph{NeuroDoc} and \emph{NeuroAudit} as the operational toolchain for this methodology. \emph{NeuroDoc} provides rulebook-constrained, LLM-assisted drafting and document-kernel maintenance support. \emph{NeuroAudit} packages the rulebook into a reviewer-facing interface for validation, amendment, and release-time status control.
\end{itemize}

\section{Related Work}

Prior work comes from three aspects: 1) EEG standards and software infrastructure for dataset organization and processing; 2) generic frameworks for dataset documentation; and 3) large language model systems for automated documentation and scientific workflows. Table~\ref{tab:related_work_summary} summarizes the design gap addressed in this paper.

\begin{table*}[t]
\centering
\footnotesize
\setlength{\tabcolsep}{4pt}
\begin{tabular}{p{4.0cm}p{1.1cm}p{1.7cm}p{1.5cm}p{1.5cm}p{1.6cm}}
\toprule
Work & Scope & Core unit & Task  & Executable & Governance \\
\midrule
BIDS family~\cite{bids-sdata-2016,eeg-bids-sdata-2019,mne-bids-joss-2019} & mixed & layout+meta & none & validator & community  \\
HED / COBIDAS~\cite{hed-neuroimage-2021,cobidas-meeg-natneuro-2020} & EEG & events+report & event-level & tooling-lite & consensus \\
MOABB / EEGUnity~\cite{moabb-jne-2018,eegunity-tnsre-2025} & EEG & pipeline+data & loader-level & pipelines & none \\
Croissant~\cite{croissant-2024} & generic & metadata card & none & loader-ready & none \\
LLM curation/docs~\cite{dca-bench-kdd-2025,pre-meta-bioinformatics-2025,auto-benchmarkcard-arxiv-2025} & generic & cards+docs & none & text-check & none \\
EEGAgent~\cite{eegagent-aaai-2026} & EEG & analysis agent & none & tool orches. & none \\
proposed method & EEG & document+kernel & full & kernel-linked & community \\
\bottomrule
\end{tabular}
\caption{Comparison of adjacent lines of work. Columns summarize application scope, core standardized unit, task-specification depth, executable linkage, and governance mechanism.}
\label{tab:related_work_summary}
\end{table*}

\subsection{EEG Data Standards, Toolkits, and Documentation Frameworks}

Prior work has made EEG datasets easier to store, describe, benchmark, and process. BIDS \cite{bids-sdata-2016} and EEG-BIDS \cite{eeg-bids-sdata-2019} provide conventions for dataset organization and modality-specific metadata, while MNE-BIDS \cite{mne-bids-joss-2019} connects those conventions to practical analysis workflows. HED contributes a structured language for encoding event semantics \cite{hed-neuroimage-2021}, and COBIDAS-MEEG provides reporting guidance for reproducible EEG and MEG studies \cite{cobidas-meeg-natneuro-2020}. At the toolkit level, MOABB offers a reproducible benchmarking framework for EEG-based BCIs by standardizing datasets, paradigms, pipelines, and evaluation schemes across public benchmarks \cite{moabb-jne-2018}, while EEGUnity provides unified infrastructure for loading and processing heterogeneous EEG datasets \cite{eegunity-tnsre-2025}. In parallel, generic dataset documentation frameworks address transparency from a different angle: Croissant \cite{croissant-2024} defines machine-readable metadata for machine learning-ready datasets. Overall, these efforts standardize structure, metadata, reporting, event annotation, and high-level processing interfaces.

\subsection{LLM-based Documentation and Scientific Agent Systems}

Another literature explores whether documentation and scientific workflows can be partially automated with large language models and tool-augmented agents. In the EEG domain, EEGAgent focuses on automated EEG analysis, with emphasis on signal-level workflows \cite{eegagent-aaai-2026}. In the broader documentation space, DCA-Bench evaluates dataset curation agents \cite{dca-bench-kdd-2025}, Pre-Meta applies retrieval-augmented generation to metadata production \cite{pre-meta-bioinformatics-2025}, and Auto-BenchmarkCard targets automated benchmark documentation with factual validation \cite{auto-benchmarkcard-arxiv-2025}. These systems show that documentation-oriented generation can be grounded, validated, and partially automated. They do not provide a shared EEG task specification language, a rulebook that governs admissible task claims, or a community review process tied to executable benchmark entries.

\section{Methodology}

\subsection{Overview}

This section presents our rulebook-guided benchmark-construction methodology. We define the task specification language and entry-based representation, introduce the shared rulebook for review and machine-checkable validation, describe tools for entry generation, upgrade, and community review and provide the extensibility policy enabling corpus and rulebook evolution.

\subsection{Task Specification Language and Entry-based Representation}

Our methodology centers on a structured EEG task specification language implemented through benchmark entries. Each entry contains: a dataset Task Document (hereafter, \emph{Document}) and a dataset-specific task kernel (hereafter, \emph{Kernel}). The \emph{Document} provides the human-readable layer of the specification language, and the \emph{Kernel} provides the executable dataset-specific logic in a unified pipeline.

\begin{itemize}
\item First, the \emph{Document} includes \texttt{Paradigm}, \texttt{File Scan Result}, \texttt{Tasks}, \texttt{Data Quality Notes}, and \texttt{Kernel Responsibility}. \texttt{File Scan Result} stores program-derived facts such as scan metadata, event structure summaries, extracted information tables, and kernel fingerprints. \texttt{Tasks} stores the task specification itself, including task type, epoch method, epoch parameters, label route, target definition, and supporting evidence. Based on evidence sufficiency, we distinguish classical tasks from potential tasks. Together, these fields form the human-readable core of the task specification language.

\item Second, the \emph{Kernel} is the executable code for the entry, injecting dataset-dependent logic into a unified processing pipeline so that the tasks described in the \emph{Document} can be instantiated consistently. The entry-based representation is designed to keep these two synchronized elements aligned through generation, review, and later updates, thereby turning a dataset-specific task specification into a reusable benchmark unit.
\end{itemize}

To define EEG tasks consistently across heterogeneous datasets, we decompose each task specification into four coupled layers: 1) task type, 2) epoch method, 3) label route, and 4) meta level. Epoch anchoring and label routing are independent, so a task may use event-locked, long-event-locked, or continuous windows while drawing supervision from misc or info fields. We discuss the four layers below:

\begin{itemize}
\item Task Type. In the current specification, tasks are formulated as either classification or regression.

\item Epoch Method. We distinguish three cases: event-locked window, long-event-locked window, and continuous window. Event-locked windows use a discrete event as the anchor. Long-event-locked windows segment repeatedly with a specified overlap ratio until the boundary of a long event. Continuous windows segment repeatedly until the boundary of the recording session.

\item Label Route. We use three routes: event, misc, and info, which are also compatible with MNE \cite{mne-python-fnins-2013}. The event route uses discrete event semantics represented through annotations or stimulation-channel events; the misc route uses continuous targets stored in MISC channels; and the info route uses file-level attributes.

\item Meta Level. We distinguish raw meta and super meta. Raw meta refers to labels or fields directly available from raw metadata, such as original annotations and event tables. Super meta requires further derivation, such as binary contrasts collapsed from multiple event classes or response labels derived from event pairing.
\end{itemize}

Representative released \emph{Document} cases (8 out of 53) are provided in Appendix~\ref{appendix:document_cases}.

Public EEG datasets involve diverse paradigms as well as cognitive, clinical, sleep, and other applied settings, no single standard paradigm taxonomy cleanly covers the full range of tasks represented in the corpus \cite{moabb-jne-2018,cobidas-meeg-natneuro-2020,eegunity-tnsre-2025}. We therefore use a six-category taxonomy for corpus-level organization purposes: Complex Applied, High-level Mental, Action/Output, Sensory/Response, Biological State, and Others.

\subsection{Rulebook for Entries}

To make benchmark construction auditable and repeatable, our methodology centers on a single entry rulebook. This rulebook is the governance foundation of the task specification language and of the corpus built from it: it defines acceptable evidence, admissible task claims, document-kernel synchronization, and valid status transitions across the entry lifecycle. The full rulebook is reproduced in Appendix~\ref{appendix:entry_rulebook}.

Within the rulebook, some provisions are primarily normative and reviewer-facing, whereas others are structured enough to be checked automatically. The normative side defines evidence priority, data-over-description preference, conservative downgrade behavior, implementation-grounded task acceptance, and the responsibilities of human reviewers, automated assistants, and maintainers. This machine-checkable side protects program-derived content such as scan-derived identifiers, counts, information-table values, and kernel fingerprints from later model-side rewriting, restricts which status values may be assigned automatically, and checks section structure, legal field values, task-format compatibility, scanned-event consistency, and kernel fingerprints. We operationalize it through a detector-side rule encoding summarized in Appendix~\ref{appendix:machine_rule_subset}.

\subsection{NeuroDoc}

\subsubsection{NeuroDoc Overview}

NeuroDoc is a rulebook-guided benchmark-construction assistant that helps create and update drafts of the \emph{Document} and the \emph{Kernel} under the rulebook, with human review reserved for final acceptance. In the current implementation, NeuroDoc functions as an LLM-assisted, rulebook-constrained entry-construction workflow organized around two executable paths: a generation pipeline for initial entry drafting and an upgrading pipeline for revision based on human feedback. Functionally, NeuroDoc combines four capabilities: 1) context building from dataset-local and literature-side evidence, 2) constrained task synthesis under the task language and rulebook constraints, 3) kernel-oriented coding, and 4) validation against the machine-checkable subset of the rulebook. Project-scale statistics for the downloaded reference resources and the broader construction workflow are summarized in Appendix~\ref{appendix:reference_corpus_scale} and Appendix~\ref{appendix:project_scale}, while a small-scale quantitative characterization of NeuroDoc draft quality is summarized in Appendix~\ref{appendix:neurodoc_generation_quality}.

\subsubsection{Generation and Upgrade Pipelines}

The generation pipeline of NeuroDoc creates a first synchronized entry-based representation from dataset-local evidence and support references. The upgrade pipeline revises an existing entry when later community review returns evidence-backed amendments, executable inconsistencies, or synchronization conflicts.

\begin{figure}[t]
\centering
\includegraphics[width=\columnwidth]{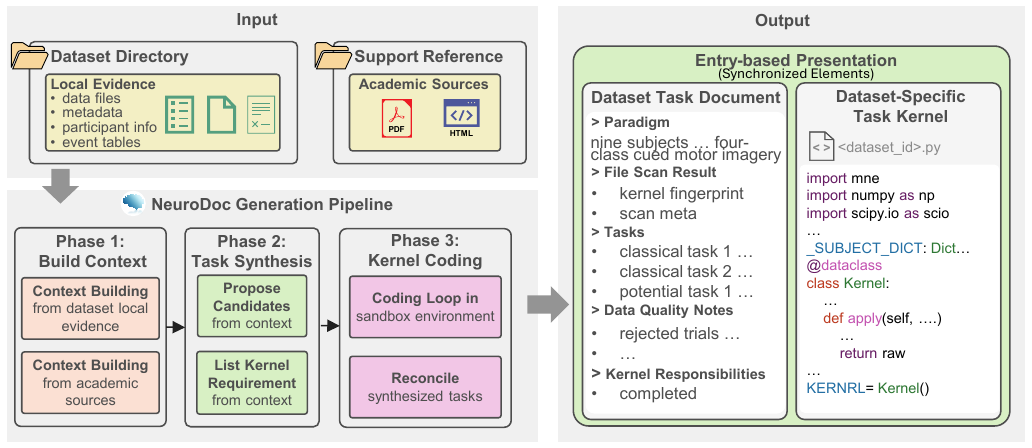}
\caption{NeuroDoc workflow for initial generation and later upgrade of benchmark entries under the shared task specification language and entry rulebook.}
\label{fig:section3_generation}
\end{figure}

The initial generation pipeline has three phases, shown in Figure~\ref{fig:section3_generation} and described below:

\textbf{Phase 1: build context} collects evidence from both the dataset directory and the support reference, combining dataset-local evidence with literature-side explanations.

\textbf{Phase 2: task synthesis} converts that context into a task specification. NeuroDoc first proposes evidence-bounded candidate tasks and then lists the corresponding kernel requirements.

\textbf{Phase 3: kernel coding} turns the synthesized task specification into an executable \emph{Kernel}. This phase combines a coding loop in a sandbox environment with a reconciliation step over synthesized tasks so that the resulting \emph{Kernel} remains aligned with the \emph{Document}.

The upgrading pipeline reuses the same core logic, but starts from an already existing entry. It utilizes reviewer annotations and returned documents to repair the \emph{Document} and the \emph{Kernel}. Besides, NeuroDoc exposes its domain logic through installable skills, enabling external agents (such as Codex or Claude Code) to become interactive tools for maintaining \emph{Document}s and \emph{Kernel}s. In this way, other agents can participate in NeuroDoc-compatible benchmark-construction workflows under the same rulebook.

\subsection{NeuroAudit-assisted Community Review and Amendment Workflow}

NeuroAudit is the reviewer-facing audit web user interface, supporting community review, document-level amendment, and status switch under the shared entry rulebook. Through this interface, community reviewers can audit generated entries, record amendments, and provide the feedback that drives later upgrades. The description of NeuroAudit is provided in Appendix~\ref{sec:appendix_neuroaudit_ui}.

Figure~\ref{fig:section3_community} describes the second stage of the methodology. Once an initial entry-based representation is produced, the \emph{Document} can be reviewed by the community via NeuroAudit. The reviewer-facing interface allows users to make step-by-step, rule-bound modifications within a predefined framework derived from the shared rulebook. This design keeps review focused on correction, adjudication, evidence checking, and document-kernel synchronization, while avoiding unrestricted rewriting of program-derived content.

The NeuroAudit workflow may lead to three document outcomes shown in Figure~\ref{fig:section3_community}: \texttt{wait\_kernel\_update}, \texttt{hallucination}, and \texttt{completed}. The \texttt{completed} state marks entries that pass review under the current rulebook and can be released as valid benchmark entries, but this does not mean they cannot be changed in the future. Completed entries may still receive later corrections, or additional task definitions through the same review process. The \texttt{hallucination} state marks entries whose claims are not adequately supported and should not proceed as valid benchmark entries. The \texttt{wait\_kernel\_update} state captures a failure mode in which the entry remains conceptually useful, but the current \emph{Kernel} requires further revision.

Under the rulebook, entries in \texttt{wait\_kernel\_update} are routed to the NeuroDoc upgrade pipeline. As illustrated in Figure~\ref{fig:section3_community}, \textbf{Phase 1: build context} analyzes the returned document and reviewer notes to identify the source of the conflict. \textbf{Phase 2: kernel coding} then revisits the coding loop and checks whether the revised \emph{Kernel} still matches the synthesized tasks. More generally, this upgrade pipeline turns review feedback into a maintenance process: executable inconsistencies are repaired through kernel-side revision, while broader but evidence-supported additions may be incorporated as reviewed amendments to completed entries. In both cases, the goal is to let entries evolve without breaking the alignment between document semantics and executable task kernels.
\begin{figure}[t]
\centering
\includegraphics[width=\columnwidth]{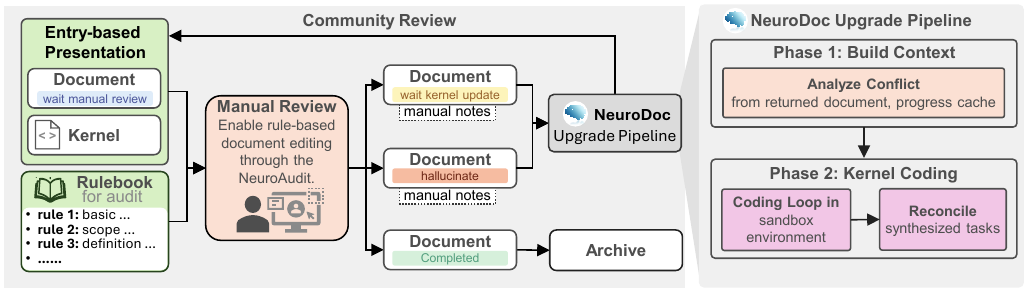}
\caption{NeuroAudit-assisted community review and NeuroDoc upgrade pipeline for auditing generated entries, resolving conflicts, and maintaining synchronized benchmark elements under the shared task specification language and entry rulebook.}
\label{fig:section3_community}
\end{figure}

\subsection{Rulebook Evolution and Extensibility}

The benchmark corpus is intended to grow over time, and the rulebook is therefore treated as extensible governance material. The community may propose extensions to controlled dimensions such as task type, epoch method, label route, target-format conventions, or other task-interface fields. Once accepted by the community, the rulebook can be updated and existing entries can be amended or expanded accordingly.

When a new rule conflicts with a legacy rule, the corpus governance policy may remain temporarily accepted during a deprecation period while entries are progressively upgraded. This preserves backward compatibility for released benchmark entries and allows staged migration as the corpus evolves.

\section{Benchmark Corpus}

Having defined the methodology, we characterize the resulting benchmark corpus as an evaluation resource and demonstrate the type of benchmark infrastructure it produces. We first summarize the scale, coverage, and structural diversity of the completed corpus used in this study, and then use downstream experiments in the next subsection to examine how these released benchmark entries can be consumed under a shared task specification layer.

\subsection{Corpus Statistics}
\begin{figure}[h]
\centering
\includegraphics[width=\textwidth]{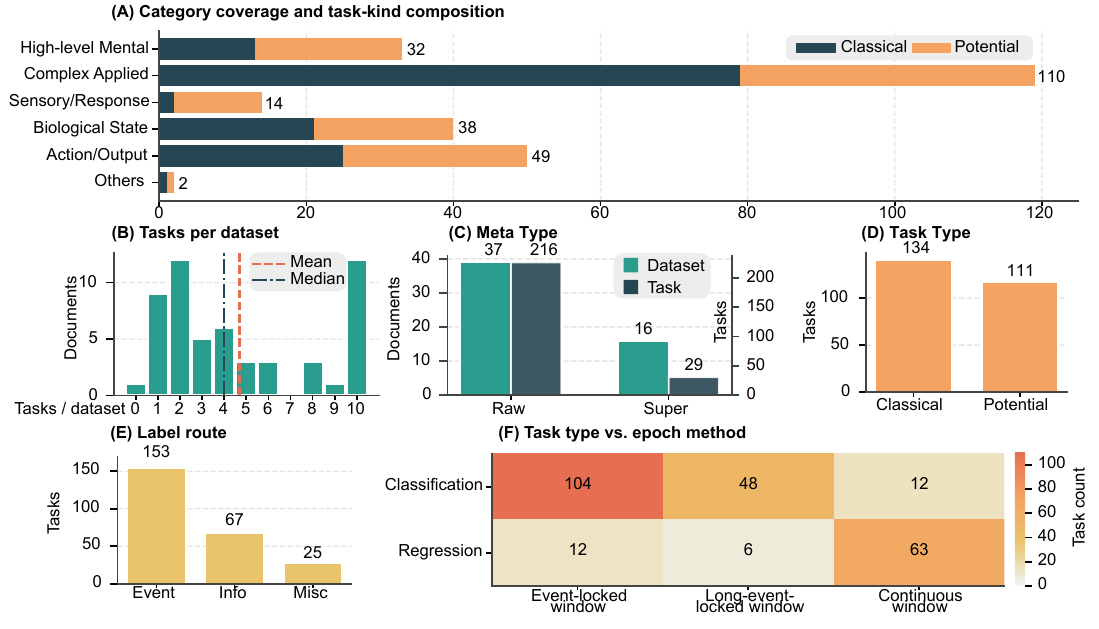}
\caption{Overview of the completed benchmark corpus used in this paper.}
\label{fig:doc_resource_overview}
\end{figure}
Figure~\ref{fig:doc_resource_overview} summarizes the completed benchmark corpus used in this paper. Our quantitative analysis is restricted to \textbf{53 completed and reviewed benchmark entries} that are treated as released benchmark units. These entries contain 245 task definitions, with a mean of 4.62 tasks per dataset document, a median of 4, and a maximum of 10. Most datasets can participate in several benchmark settings through one synchronized document-kernel pair.

The completed corpus also exhibits broad topical coverage across heterogeneous EEG paradigms. At the task-definition level, the largest categories are Complex Applied (110 tasks), Action/Output (49), Biological State (38), and High-level Mental (32). The released corpus therefore spans motor, cognitive, sensory, sleep/clinical, and applied paradigms under a common representation layer.

Among the 245 task definitions, 134 are currently marked as classical tasks and 111 as potential tasks. Classical tasks anchor the corpus to established literature-backed formulations, while potential tasks expand the reusable search space for later evaluation and adaptation studies. Figure~\ref{fig:doc_resource_overview}(B) and (D) therefore show that the corpus is both multi-task at the entry level and diverse in the maturity of its task specifications.

The task-interface statistics also show that the completed corpus is not tied to a single labeling or epoching recipe. Figure~\ref{fig:doc_resource_overview}(E) indicates that most tasks are routed through event-derived targets (153 tasks), but the corpus also contains substantial use of info-based labels (67 tasks) and misc-channel targets (25 tasks). Figure~\ref{fig:doc_resource_overview}(C) shows that 16 completed dataset entries require super-level metadata, whereas 37 do not, and that most task targets remain grounded in raw-level signals or annotations (216 raw-level targets versus 29 super-level targets). Figure~\ref{fig:doc_resource_overview}(F) shows that the completed corpus covers both major supervised task families, with 164 classification tasks and 81 regression tasks. Classification is dominated by event-locked and long-event-locked formulations (104 and 48 tasks, respectively), whereas regression more often relies on continuous-window definitions (63 tasks).

Although this paper focuses on the 53 completed entries above, the broader benchmark database continues to accumulate additional datasets in earlier generation, upgrading, and review stages. Additional project-scale context beyond the released subset is summarized in Appendix~\ref{appendix:project_scale}.

\subsection{Execution-based Validation with Four EEG Model Backbones}
\begin{figure}[h]
\centering
\includegraphics[width=\textwidth]{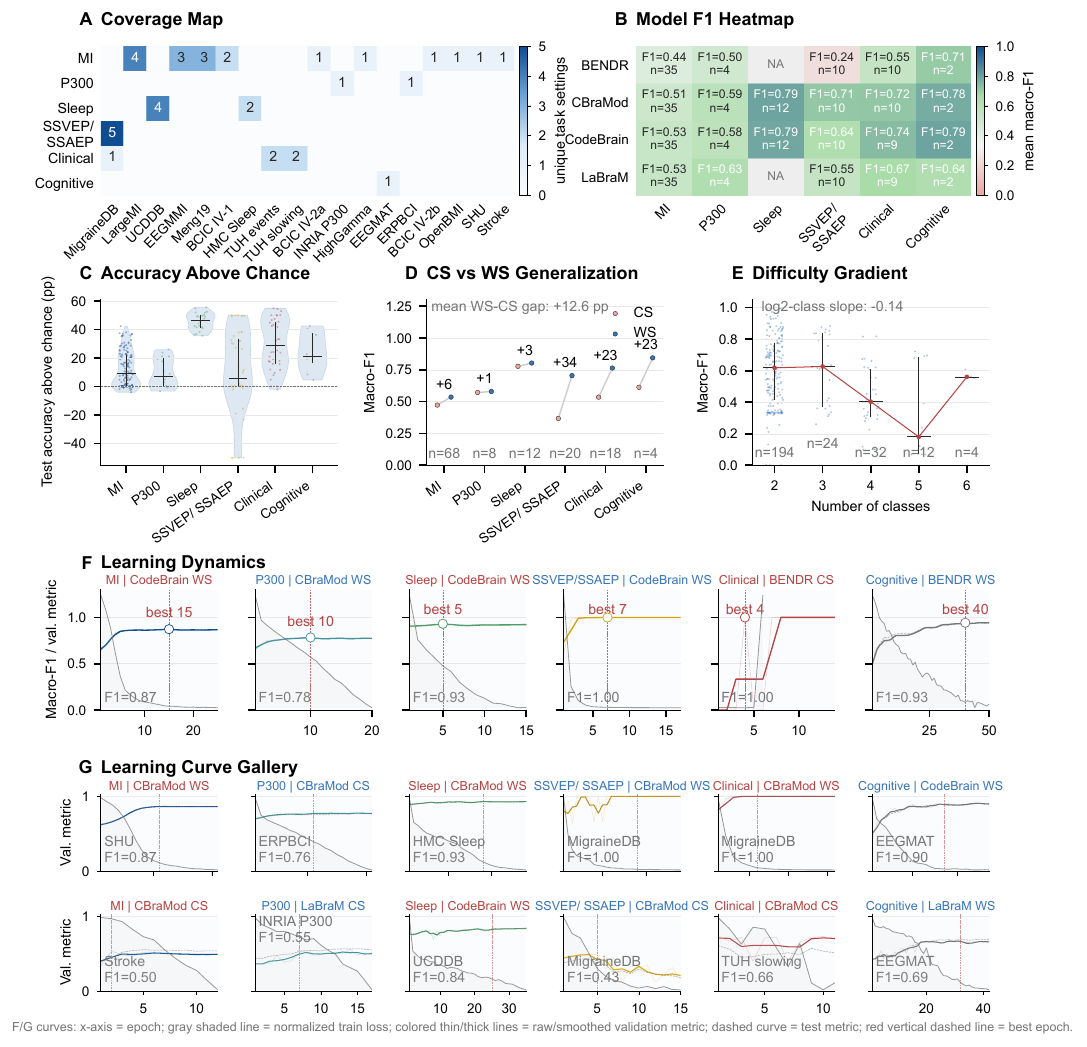}
\caption{Execution-based validation of the released benchmark corpus with four EEG model backbones.}
\label{fig:foundation_model_validation}
\end{figure}
Beyond corpus-level statistics, we further examine whether the released entries function as executable downstream benchmark units once their paired kernels are used to process raw data and instantiate benchmark tasks. To do so, we use four representative EEG foundation model backbones---BENDR \cite{bendr-fnhum-2021}, CBraMod \cite{cbramod-iclr-2025}, CodeBrain \cite{codebrain-iclr-2026}, and LaBraM \cite{labram-iclr-2024}---as downstream consumers under a shared task specification layer. Our aggregate analysis covers 266 runs spanning 18 datasets and 37 benchmark tasks, evaluated under two split modes (cross-subject and within-subject) across six task families: motor imagery (MI), P300, sleep staging, SSVEP/SSAEP, clinical EEG, and cognitive tasks. Macro-F1 is used as the primary cross-task summary metric, complemented by auxiliary views such as test accuracy above a random baseline and representative optimization traces. The experimental results are summarized in Figure~\ref{fig:foundation_model_validation}. The execution-validation subset and its supporting project-scale context are summarized in Appendix~\ref{appendix:execution_validation_datasets}; task definitions follow the corresponding released dataset documents.

Figure~\ref{fig:foundation_model_validation}(A) maps benchmark coverage across datasets and task families, demonstrating that the released corpus encompasses a broad range of paradigms. Figure~\ref{fig:foundation_model_validation}(B) summarizes family-level mean macro-F1 for the four EEG foundation model backbones and shows that all four can be run under the shared task specification layer with interpretable task-level outputs. The remaining sub-figures provide additional evidence that the instantiated benchmark tasks preserve meaningful learning structure. Figure~\ref{fig:foundation_model_validation}(C) shows that test accuracy is generally above the random baseline across task families. Figure~\ref{fig:foundation_model_validation}(D) shows that within-subject (WS) evaluation consistently outperforms cross-subject (CS) evaluation, with a mean WS--CS gap of 12.6 percentage points. Figure~\ref{fig:foundation_model_validation}(E) shows a negative relationship between macro-F1 and task granularity as the number of classes increases (log$_2$-class slope: $-0.14$). Figure~\ref{fig:foundation_model_validation}(F) and (G) provide representative optimization traces, where gray shaded curves denote normalized training loss, colored thin/thick curves denote raw/smoothed validation metrics, dashed curves denote test metrics, and red vertical dashed lines mark the selected best epochs. Overall, these results show that the released corpus can serve as an executable benchmark substrate for downstream evaluation across multiple EEG model backbones.

\section{Conclusion}

As EEG foundation models increasingly rely on large aggregated corpora, the bottleneck is no longer only dataset access, but task specification. This paper studies whether a shared task specification language and rulebook can make heterogeneous public EEG datasets explicit, executable, and auditable enough for reusable benchmarking. Our main methodological contribution is a rulebook-guided EEG task specification language implemented through benchmark entries that pair field-structured task documents with executable kernels. Our main resource contribution is a community-reviewed corpus centered on 53 completed and reviewed benchmark entries. Our main systems contribution is the NeuroDoc--NeuroAudit toolchain for drafting, review, amendment, and release management under shared governance. Together, these contributions elevate EEG task operationalization into a benchmark object that can be inspected, executed, and maintained across datasets without repackaging raw recordings. A current limitation is that the quantitative analysis in this paper is restricted to the 53 entries that have reached the completed-and-reviewed release state under the current rulebook. Beyond this released subset, the broader database already contains roughly 200 additional draft or unaudited documents, but these cannot yet be treated as equally reliable benchmark units until community review, amendment, and document-kernel consistency checks are completed. Future work includes extending the rulebook and corpus to broader EEG task families and datasets, and further characterizing LLM-assisted benchmark drafting and review support, including description precision, evidence grounding, kernel alignment, and the human effort required for review.

\bibliographystyle{plainnat}
\bibliography{references}

\newpage
\appendix
\input{appendix_execution_validation_datasets}
\clearpage
\input{appendix_neurodoc_generation_quality}
\clearpage
\input{appendix_neuroaudit_ui}
\clearpage
\input{appendix_neuroaudit_rulebook}
\clearpage
\input{appendix_semantic_rules}
\clearpage
\input{appendix_document_cases}
\clearpage
\input{appendix_broader_impacts}

\end{document}

%% file: appendix_execution_validation_datasets.tex
\section{Project Scale, Reference Resources, and Execution Validation Context}
\label{appendix:execution_validation_datasets}

This appendix provides project-scale context for the released benchmark. The broader project began from public-source downloads stored on our servers, spanning more than 300 raw datasets and approximately 23~TB of total storage. From this larger collection, we isolated approximately 17.7~TB of scanned EEG/EEG-like signal data for benchmark construction and maintained about 8.46~GB of downloaded reference resources to support document-level task synthesis. NeuroDoc then generated benchmark-document workspaces for 286 datasets, of which 53 reached the completed-and-reviewed release state used in the present paper. Within that released set, a smaller subset was used for execution-based validation across four EEG foundation model backbones. The remainder of this appendix therefore summarizes the scanned data scale, the downloaded reference resources, and the compute and configuration details of the execution-validation experiments.

\subsection{Broader Database and Benchmark-Construction Scale}
\label{appendix:project_scale}

The total project footprint of approximately 23~TB includes both raw signal files and auxiliary materials downloaded during benchmark construction. Table~\ref{tab:database_phase_summary} isolates the scanned EEG/EEG-like subset and summarizes its four-stage database statistics in terms of file count, recording duration, storage size, and metadata completeness. These statistics describe the signal-side resource base from which the current benchmark documents were constructed.

\begin{table}[h]
\centering
\footnotesize
\setlength{\tabcolsep}{4pt}
\begin{tabular}{p{2.5cm}rrrrrr}
\toprule
\textbf{Database Phase} & \textbf{Files} & \textbf{Total Hours} & \textbf{Total Size (GB)} & \textbf{Age Complete} & \textbf{Sex Complete} & \textbf{Task Complete} \\
\midrule
Phase I database & 4,956 & 7,244.22 & 404.4 & 36.86\% & 37.95\% & 85.57\% \\
Phase II database & 83,234 & 30,584.01 & 1,825.5 & 0.00\% & 0.00\% & 16.32\% \\
Phase III database & 67,707 & 30,029.11 & 15,159.7 & 80.42\% & 80.10\% & 87.63\% \\
Phase IV database & 8,444 & 75,466.41 & 348.1 & 100.00\% & 100.00\% & 100.00\% \\
\midrule
Total & 164,341 & 143,323.75 & 17,737.7 & 39.38\% & 39.28\% & 52.09\% \\
\bottomrule
\end{tabular}
\caption{Four-stage summary of the scanned EEG/EEG-like database used in the broader benchmark-construction workflow.}
\label{tab:database_phase_summary}
\end{table}

\subsection{Downloaded Reference Resources for Task Synthesis}
\label{appendix:reference_corpus_scale}

Besides dataset-local evidence, NeuroDoc also builds literature-side context from downloaded reference resources stored alongside the dataset-document workflow. These references were used to support task synthesis, evidence checking, and later review. Across the 286 dataset-document workspaces, the local \texttt{references} directories currently contain 3,556 downloaded reference files occupying approximately 8.46~GB in total. Because many of the downloaded files are copyrighted PDFs or publisher-hosted web pages, the reference files themselves are not redistributed in the repository. We report only aggregate resource statistics here. The local file count reflects all downloaded files present in the workspace \texttt{references} directories, whereas the relevance report below counts only the reference items included in the report snapshot and category-labeling pipeline.

\begin{table}[h]
\centering
\footnotesize
\setlength{\tabcolsep}{6pt}
\begin{tabular}{lr}
\toprule
\textbf{Reference Resource Statistic} & \textbf{Value} \\
\midrule
Dataset documents covered by the report & 285 \\
Total reference items & 3,095 \\
PDF files & 2,948 \\
HTML pages & 147 \\
Relevant & 450 \\
Uncertain & 121 \\
Irrelevant & 2,086 \\
Junk & 438 \\
Average references per dataset & 10.86 \\
Maximum references for one dataset & 64 \\
\bottomrule
\end{tabular}
\caption{Aggregate statistics of the downloaded reference resources used for literature-side context building and relevance filtering. Counts come from the reference relevance report used in the broader NeuroDoc workflow. The report covers 285 dataset documents because one local workspace was outside the report-generation snapshot used for this summary.}
\label{tab:reference_corpus_summary}
\end{table}

\subsection{Execution-validation Dataset Subset and Training Context}
\label{appendix:execution_validation_subset}

The downstream validation experiments in Section~4.2 use 18 released benchmark entries whose dataset documents are already in the \texttt{completed} state. These entries support the 37 benchmark tasks used in the current execution study and are drawn from the broader released subset of 53 completed-and-reviewed entries.

The computational footprint of the project has two distinct parts. First, the larger benchmark-construction workflow is primarily CPU-bound. Its main cost comes from large-scale dataset scanning, metadata extraction, event parsing, table generation, rule-based validation, and reference-side context preparation over the broader repository described above. This corpus-construction stage took roughly two weeks of iterative processing and review.

Second, GPU resources were used mainly for the model-side execution-based validation experiments in Section~4.2. These runs were carried out on internal servers equipped with multiple NVIDIA GPUs, including A800, RTX A6000, RTX 4090, and RTX 3090 devices. Across the reported execution study, we ran 266 evaluations over 18 datasets and 37 benchmark tasks. The exact task definitions used for each dataset follow the corresponding released dataset document.

For the model-side runs, the reported evaluation uses two split modes, cross-subject and within-subject, as described in Section~4.2. Across completed scheduler runs supporting this evaluation campaign, all training jobs used the AdamW optimizer with learning rate $10^{-4}$, weight decay $0.05$, early stopping patience $10$, and up to $50$ epochs, with $8$ data-loading workers. BENDR used raw EEG inputs with batch size $64$, whereas CBraMod, CodeBrain, and LaBraM used patched inputs with patch size $200$ and batch size $128$.

\begin{table}[h]
\centering
\footnotesize
\setlength{\tabcolsep}{4pt}
\begin{tabular}{p{4.0cm}p{1.8cm}p{5.0cm}}
\toprule
\textbf{Dataset ID} & \textbf{Short Name} & \textbf{Primary Validation Family} \\
\midrule
\texttt{bcic\_iv\_1} & BCIC IV-1 & Motor imagery  \\
\texttt{bcic\_iv\_2a} & BCIC IV-2a & Motor imagery  \\
\texttt{bcic\_iv\_2b} & BCIC IV-2b & Motor imagery  \\
\texttt{figshare\_largemi} & LargeMI & Motor imagery  \\
\texttt{figshare\_meng2019} & Meng19 & Motor imagery  \\
\texttt{figshare\_shudb} & SHU & Motor imagery  \\
\texttt{figshare\_stroke} & Stroke & Motor imagery  \\
\texttt{kaggle\_inria} & INRIA P300 & P300  \\
\texttt{other\_highgammadataset} & HighGamma & Motor imagery  \\
\texttt{other\_migrainedb} & MigraineDB & SSVEP/SSAEP and clinical EEG  \\
\texttt{other\_openbmi} & OpenBMI & Motor imagery  \\
\texttt{physionet\_eegmat} & EEGMAT & Cognitive  \\
\texttt{physionet\_eegmmidb} & EEGMMI & Motor imagery  \\
\texttt{physionet\_erpbci} & ERPBCI & P300  \\
\texttt{physionet\_hmcsleepstaging} & HMC Sleep & Sleep staging  \\
\texttt{physionet\_ucddb} & UCDDB & Sleep staging  \\
\texttt{tuh\_eeg\_events} & TUH events & Clinical EEG  \\
\texttt{tuh\_eeg\_slowing} & TUH slowing & Clinical EEG  \\
\bottomrule
\end{tabular}
\caption{Released dataset subset used in the execution-based validation. Task definitions are not restated here; they follow the corresponding released dataset documents, and additional training and compute context is described in the main text of this appendix.}
\label{tab:execution_validation_datasets}
\end{table}

%% file: appendix_neurodoc_generation_quality.tex
\section{NeuroDoc Generation Quality Characterization}
\label{appendix:neurodoc_generation_quality}

This appendix provides experimental results on evaluating NeuroDoc generation quality. The purpose is to evaluate how well NeuroDoc can produce review-ready benchmark-document drafts under the rulebook, and to identify which parts of the drafting workflow are already strong versus which parts still rely more heavily on later review.

For this evaluation, we used Kimi-K2.6 as the LLM interface of NeuroDoc and evaluated NeuroDoc on 43 datasets from several dataset sources, including OpenNeuro/BIDS, custom or competition datasets, PhysioNet, and TUH(Temple University Hospital) EEG. For each dataset, we compared the generated draft document with the corresponding manual-reviewed document. The comparison measures how closely NeuroDoc can approach the manual-reviewed document.

The results are organized as follows. Table~\ref{tab:appendix_neurodoc_headline} reports overall metrics across all evaluated datasets. Table~\ref{tab:appendix_neurodoc_family} reports the results by dataset family.

\begin{table}[h]
\centering
\footnotesize
\setlength{\tabcolsep}{5pt}
\begin{tabularx}{\linewidth}{Xr@{\hspace{1.2em}}Xr}
\toprule
\multicolumn{2}{c}{\textbf{Pipeline \& Coverage}} & \multicolumn{2}{c}{\textbf{Quality \& Tasks}} \\
\cmidrule(r){1-2}\cmidrule(l){3-4}
\textbf{Metric} & \textbf{Result} & \textbf{Metric} & \textbf{Result} \\
\midrule
Datasets tested & 43 & Placeholder DOI rate & 0.0\% \\
Generated documents & 39/43 (90.7\%) & Missing-evidence rate & 0.0\% \\
Feasible datasets & 42/43 (97.7\%) & Not-dispatchable task rate & 0.0\% \\
Scan completed among feasible datasets & 38/42 (90.5\%) & Invalid-category rate & 0.0\% \\
Kernel produced & 42/42 (100.0\%) & BCIC/MI hallucination warning rate & 3/39 docs (7.7\%) \\
Kernel validation success & 20/42 (47.6\%) & Task count exact match & 18.6\% \\
Structural pass & 35/39 (89.7\%) & Generated-task redundancy rate & 36.1\% \\
Event Meaning coverage & 80.0\% & Reviewed-task missing rate & 36.9\% \\
Info Notes coverage for non-missing fields & 100.0\% & Median LLM budget & 29 calls \\
\bottomrule
\end{tabularx}
\caption{NeuroDoc generation-quality results.}
\label{tab:appendix_neurodoc_headline}
\end{table}

Table~\ref{tab:appendix_neurodoc_headline} shows that NeuroDoc performs well as a structured drafting system. Almost all feasible datasets receive a generated document, most generated drafts pass rule-level structural checks, and classic low-level hallucination are rare. The table show that the structure, evidence formatting, and basic completeness are already relatively stable, whereas kernel validation, sidecar-heavy event extraction, and task over- or under-splitting remain the main sources of downstream mismatch against the reviewed documents.

\begin{table}[h]
\centering
\footnotesize
\setlength{\tabcolsep}{4pt}
\begin{tabular}{lrrrrrr}
\toprule
\textbf{Dataset Family} & \textbf{N} & \textbf{Kernel OK} & \textbf{Scan Done} & \textbf{Structural Pass} & \textbf{Task Macro F1} & \textbf{Missing Tasks} \\
\midrule
OpenNeuro/BIDS & 17 & 5/16 (31.3\%) & 16/16 (100.0\%) & 17/17 (100.0\%) & 45.8\% & 14.0\% \\
Custom/Competition & 12 & 10/12 (83.3\%) & 12/12 (100.0\%) & 9/12 (75.0\%) & 36.9\% & 45.5\% \\
PhysioNet & 8 & 4/8 (50.0\%) & 6/8 (75.0\%) & 6/6 (100.0\%) & 7.7\% & 81.3\% \\
TUH EEG & 6 & 1/6 (16.7\%) & 4/6 (66.7\%) & 3/4 (75.0\%) & 0.0\% & 77.8\% \\
\bottomrule
\end{tabular}
\caption{Family-level summary.}
\label{tab:appendix_neurodoc_family}
\end{table}

Table~\ref{tab:appendix_neurodoc_family} shows that NeuroDoc quality is strongly family-dependent. The strongest behavior appears in settings where local evidence is explicit and structurally standardized. For OpenNeuro/BIDS datasets, event extraction and document structure are often strong enough that later review can focus more on task pruning, task merging, and evidence wording than on wholesale reconstruction. For custom and competition datasets, kernel generation can already work well in practice, but these datasets still expose a recurring semantic risk: raw event identifiers may be preserved correctly while task interpretation remains too broad or too eager. By contrast, PhysioNet and TUH remain the weakest families in this evaluation, indicating that interval-sidecar handling, external annotation discovery, and more clinically loaded event semantics are still the hardest parts of the current drafting workflow.

In short, these experiments show that NeuroDoc already provides useful rulebook-constrained drafting support at a meaningful scale, enabling heterogeneous raw inputs to be transformed into structured, review-ready draft document with relatively high completion rates.

%% file: appendix_neuroaudit_ui.tex
\section{NeuroAudit UI}
\label{sec:appendix_neuroaudit_ui}

This appendix summarizes the reviewer-facing interface of NeuroAudit, which is the audit web user interface used to support community review, amendment, and release-time status control. NeuroAudit emphasizes three goals: keeping the \emph{Document} and \emph{Kernel} synchronized during review, exposing task definitions in a structured and editable form, and making rulebook-guided checking visible to reviewers while they audit an entry.

\begin{figure}[h]
\centering
\includegraphics[width=\textwidth]{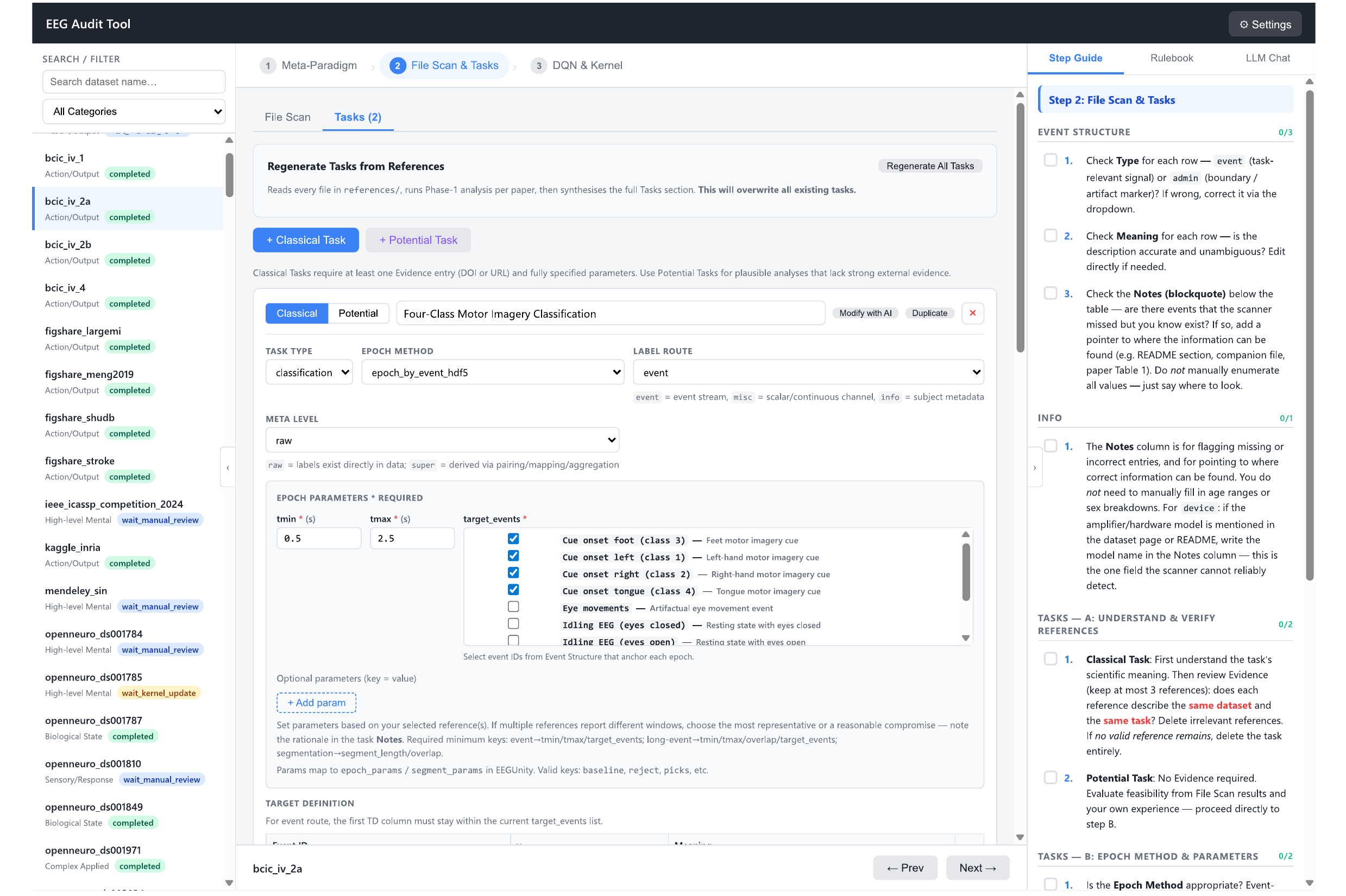}
\caption{Representative NeuroAudit task-review page.}
\label{fig:appendix_neuroaudit_ui}
\end{figure}

Figure~\ref{fig:appendix_neuroaudit_ui} shows a representative task-review page. The interface is organized into three coordinated regions. The left column acts as the corpus navigation panel: reviewers can search datasets, filter by paradigm category, and inspect status values such as \texttt{completed}, \texttt{wait\_manual\_review}, and \texttt{wait\_kernel\_update}.

The center panel is the main structured review workspace. NeuroAudit exposes the benchmark entry through staged tabs, including meta-paradigm information, file-scan results, tasks, data-quality notes, and kernel-related content. In the task view illustrated here, reviewers can inspect and correct each task through explicit fields for task type, epoch method, label route, meta level, epoch parameters, and target definition. Classical and potential tasks are edited through the same structured template, which helps preserve \emph{Document} consistency while still allowing reviewers to add evidence-backed refinements or propose additional task formulations.

The same task workspace also supports assisted regeneration through the \texttt{regenerate all tasks} control. This action is designed for cases where the reviewer has already completed file scanning and event inspection, but wants to rebuild the task region more efficiently after collecting additional support references. When triggered, NeuroAudit looks up the configured local reference directory on the user's disk, retrieves the relevant source materials, combines them with the scan-derived annotation and event information already stored in the current \emph{Document}, and then calls an attached LLM interface to regenerate the entire \texttt{Tasks} section. In effect, the interface lets users convert newly collected references into updated task cards without restarting the full audit process from scratch.

This regeneration step is especially useful for community maintenance. Reviewers often agree on the observed event structure of a dataset before they agree on the exact task decomposition, epoch design, or target construction supported by the literature. The \texttt{regenerate all tasks} pathway gives them a practical way to incorporate better references, refresh the task proposals, and then continue manual inspection in the same structured editor. Because the regenerated tasks re-enter the same rulebook-constrained UI, the feature accelerates task-card construction while still keeping later review and amendment decisions explicit and comparable across datasets.

The right column provides reviewer guidance and rulebook support. NeuroAudit presents checklist-style validation hints tied to the current review stage, including event-structure checks, information-field checks, evidence verification, and task-parameter inspection. This design helps reviewers distinguish between missing evidence, ambiguous semantics, and executable inconsistencies. In practice, it also improves reviewer consistency across datasets by turning the rulebook into a visible part of the interface.

\begin{figure}[h]
\centering
\begin{minipage}[t]{0.49\textwidth}
\centering
\includegraphics[width=\linewidth]{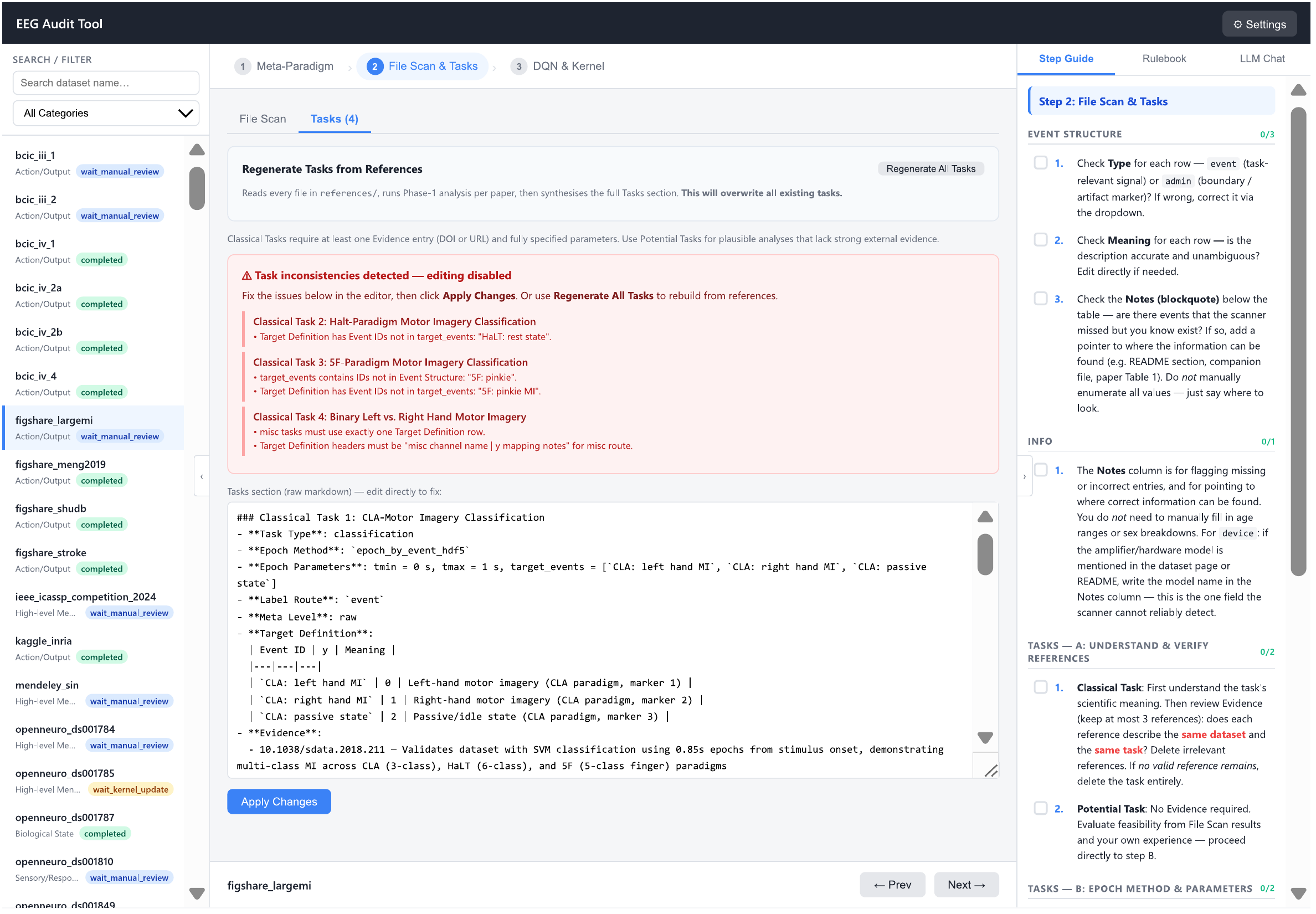}

\small \textbf{(a)}
\end{minipage}\hfill
\begin{minipage}[t]{0.49\textwidth}
\centering
\includegraphics[width=\linewidth]{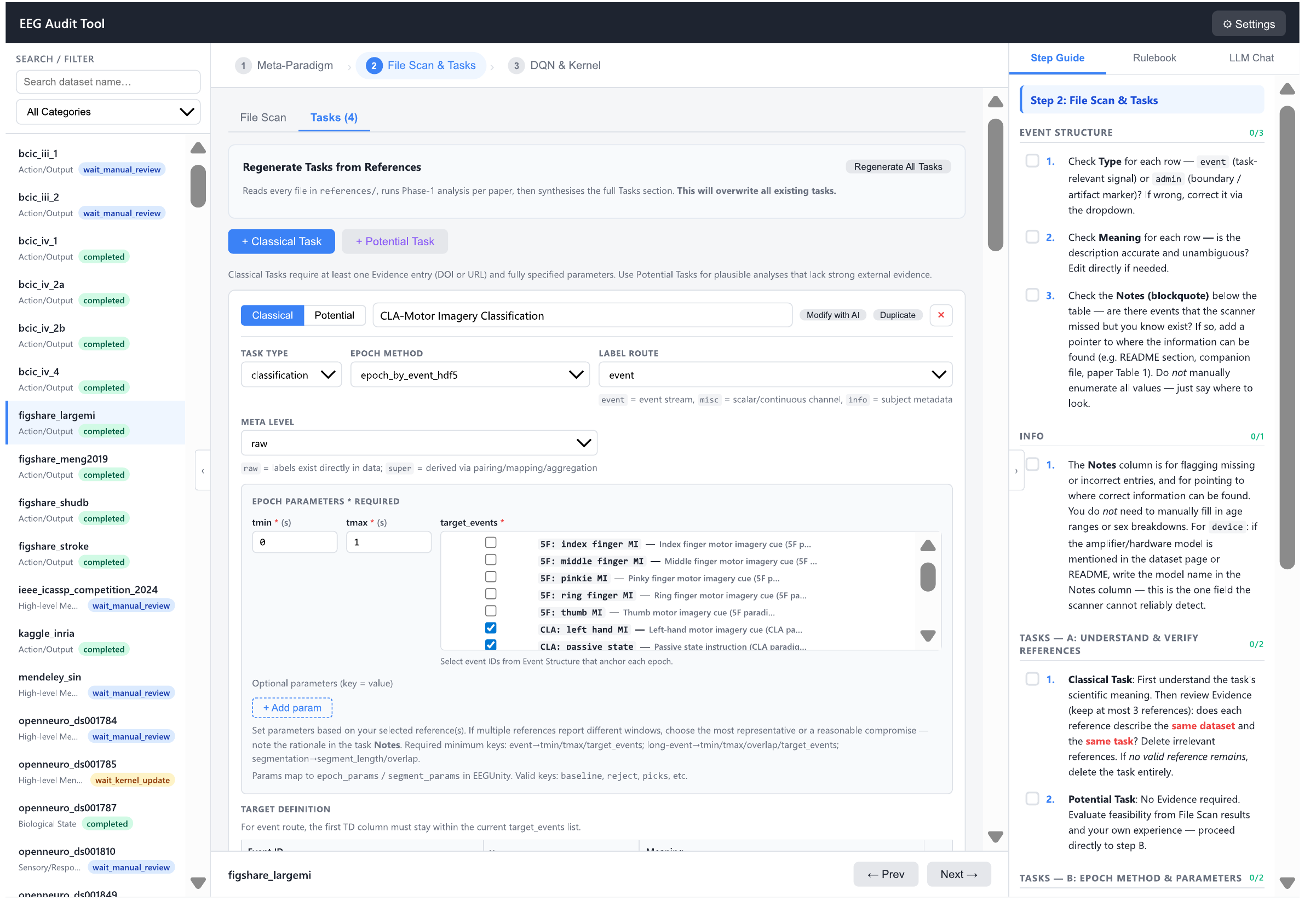}

\small \textbf{(b)}
\end{minipage}
\caption{Rulebook-guided task validation in NeuroAudit: (a) the editor surfaces detailed errors when task fields violate logical constraints, and (b) the corrected task returns to compact card mode after successful validation.}
\label{fig:appendix_neuroaudit_validation}
\end{figure}

Figure~\ref{fig:appendix_neuroaudit_validation} illustrates a more focused interaction in the task editor. NeuroAudit does not rely on reviewers to remember rulebook constraints from memory while editing task fields. Instead, it performs rulebook-guided validation over both field-level admissibility and cross-field logical consistency. When a reviewer introduces a contradiction in the task region, the interface surfaces localized error messages that explain which constraints have been violated. After the conflicting fields are corrected and the update passes validation, the task is rendered again in compact card form. This interaction pattern reduces reviewer memory load, makes document audit easier to complete, and helps different reviewers apply the same rules when forming community judgments.

Overall, NeuroAudit is where benchmark legitimacy is established: reviewers verify event meaning, confirm whether tasks are classical or potential, record data-quality issues, and decide whether an entry should be marked as \texttt{completed}, returned for upgrading, or rejected as unsupported. The web user interface is the main mechanism through which community review becomes structured, auditable, and scalable.

%% file: appendix_neuroaudit_rulebook.tex
\section{Entry Rulebook}
\label{appendix:entry_rulebook}

\begin{tcolorbox}[breakable,enhanced,colback=white,colframe=gray!50,boxrule=0.8pt,arc=4pt,left=8pt,right=8pt,top=6pt,bottom=6pt,title={\normalsize\bfseries NeuroAudit Rulebook v1.0.0},coltitle=black,colbacktitle=gray!15,titlerule=0.5pt]
\small
\par\bigskip
\noindent\rule{\linewidth}{0.8pt}\par\nopagebreak
\noindent{\bfseries\normalsize Part I. General Provisions}\par\nopagebreak
\noindent\rule{\linewidth}{0.8pt}\par\smallskip

\medskip\noindent{\bfseries\normalsize Chapter 1. Purpose and Scope}
\par\nopagebreak\noindent\rule{\linewidth}{0.4pt}\par\smallskip

\smallskip\noindent{\bfseries\normalsize Article 1. Purpose}\par\smallskip

\noindent This rulebook standardizes generation, review, revision, and re-review of EEG dataset \texttt{analysis.md} documents so that they can:\par\smallskip

\begin{enumerate}[leftmargin=1.5em,noitemsep,topsep=2pt]
\item Provide human reviewers with consistent, explicit, actionable review standards.
\item Provide LLM review scripts with stable, programmable, traceable decision rules.
\item Provide structured implementation guidance for later kernels, EEGUnity exports, and HDF5 production.
\end{enumerate}

\smallskip\noindent{\bfseries\normalsize Article 2. Scope}\par\smallskip

\noindent This rulebook applies to:\par\smallskip

\begin{enumerate}[leftmargin=1.5em,noitemsep,topsep=2pt]
\item Creating new dataset documents.
\item Revising existing dataset documents.
\item Reviewing core areas such as \texttt{Classical Task}, \texttt{Potential Task}, \texttt{Event Structure}, and metadata.
\item Writing later LLM review rules, prompt rules, and fallback rules.
\end{enumerate}

\smallskip\noindent{\bfseries\normalsize Article 3. Core Principles}\par\smallskip

\noindent Document generation and review must follow these principles:\par\smallskip

\begin{enumerate}[leftmargin=1.5em,noitemsep,topsep=2pt]
\item Truthfulness first.
\item Actual data has priority over web descriptions.
\item Conservatism has priority over guessing.
\item Traceability has priority over brevity.
\item Implementable task definitions have priority over theoretically imaginable ones.
\end{enumerate}

\medskip\noindent{\bfseries\normalsize Chapter 2. Terms}
\par\nopagebreak\noindent\rule{\linewidth}{0.4pt}\par\smallskip

\smallskip\noindent{\bfseries\normalsize Article 4. Dataset Document}\par\smallskip

\noindent A dataset document means the \texttt{analysis.md} file under a single dataset directory, or a same-structure successor version.\par\smallskip

\smallskip\noindent{\bfseries\normalsize Article 5. Raw Meta}\par\smallskip

\noindent \texttt{raw meta} means labels, events, codes, annotations, fields, or markers that already exist in the raw data and can be read without additional derivation. Examples include:\par\smallskip

\begin{enumerate}[leftmargin=1.5em,noitemsep,topsep=2pt]
\item \texttt{trial\_type}, event codes, and event names in \texttt{events.tsv}.
\item Raw codes in trigger or stim channels.
\item Discrete events in raw annotation files.
\item Directly available behavioral fields in raw tables.
\end{enumerate}

\smallskip\noindent{\bfseries\normalsize Article 6. Super Meta}\par\smallskip

\noindent \texttt{super meta} means labels or targets that do not directly exist in the raw data and must be generated, mapped, folded, paired, aggregated, joined, or otherwise derived by code, rules, or human logic. Examples include:\par\smallskip

\begin{enumerate}[leftmargin=1.5em,noitemsep,topsep=2pt]
\item Response events.
\item Binary labels folded from multiple raw classes.
\item Clinical groups derived from filenames, participant tables, or external tables.
\item Correct or incorrect, target or non-target, fast or slow, and similar labels derived by pairing multiple events.
\end{enumerate}

\smallskip\noindent{\bfseries\normalsize Article 7. Kernel}\par\smallskip

\noindent \texttt{kernel} means the later data-processing code or data-loading logic used to parse raw data, construct labels, route labels, and generate EEGUnity-compatible output.\par\smallskip

\smallskip\noindent{\bfseries\normalsize Article 8. Label Route}\par\smallskip

\noindent Historically, \texttt{Label Route} meant the final EEGUnity or MNE-compatible path where labels landed, not necessarily the raw storage form. Earlier values were:\par\smallskip

\begin{enumerate}[leftmargin=1.5em,noitemsep,topsep=2pt]
\item \texttt{Annotation}: time-based annotations with \texttt{onset}, \texttt{duration}, and \texttt{description}.
\item \texttt{STIM}: trigger or stim channel route for code values or trigger pulses.
\item \texttt{MISC}: misc channel route for numeric auxiliary labels, continuous quantities, or attached labels unsuitable as event names.
\end{enumerate}

\noindent Deprecated: \texttt{Annotation}, \texttt{STIM}, and \texttt{MISC} as route values are not allowed in new documents. Current legal values are \texttt{event}, \texttt{misc}, and \texttt{info}; see Article 77.\par\smallskip

\smallskip\noindent{\bfseries\normalsize Article 9. Task}\par\smallskip

\noindent A \texttt{Task} is a definition that can directly guide EEG epoching, label construction, and supervised-learning configuration.\par\smallskip

\smallskip\noindent{\bfseries\normalsize Article 10. Classical Task}\par\smallskip

\noindent A \texttt{Classical Task} is a core task that is supported by external evidence and can be implemented from the currently public data.\par\smallskip

\smallskip\noindent{\bfseries\normalsize Article 11. Potential Task}\par\smallskip

\noindent A \texttt{Potential Task} is supported by the data itself but lacks enough external evidence to confirm it as a core benchmark task, or is only a recommended, extended, or derived task.\par\smallskip

\smallskip\noindent{\bfseries\normalsize Article 12. Evidence}\par\smallskip

\noindent \texttt{Evidence} means external material supporting a task, parameter choice, or document judgment. It includes:\par\smallskip

\begin{enumerate}[leftmargin=1.5em,noitemsep,topsep=2pt]
\item The official dataset page.
\item Official documentation.
\item Papers that clearly use the dataset.
\item Other authoritative sources.
\end{enumerate}

\medskip\noindent{\bfseries\normalsize Chapter 3. Evidence Priority and Conflict Handling}
\par\nopagebreak\noindent\rule{\linewidth}{0.4pt}\par\smallskip

\smallskip\noindent{\bfseries\normalsize Article 13. Evidence Priority}\par\smallskip

\noindent When sources conflict, use this priority order:\par\smallskip

\begin{enumerate}[leftmargin=1.5em,noitemsep,topsep=2pt]
\item Verifiable data files and accompanying annotation files.
\item Official dataset page, official README, or official documentation.
\item Official papers or papers that clearly use the dataset.
\item Third-party pages, blogs, or unofficial descriptions.
\end{enumerate}

\smallskip\noindent{\bfseries\normalsize Article 14. General Conflict Rule}\par\smallskip

\noindent When sources conflict, use the higher-priority evidence and record the conflict and decision in \texttt{Data Quality Notes}.\par\smallskip

\smallskip\noindent{\bfseries\normalsize Article 15. Data Priority}\par\smallskip

\noindent When online material or paper descriptions conflict with actual data files, the actual data files prevail. The inconsistency must still be stated explicitly.\par\smallskip

\smallskip\noindent{\bfseries\normalsize Article 16. Conservative Handling}\par\smallskip

\noindent If existing evidence is insufficient to confirm a field, task, or label:\par\smallskip

\begin{enumerate}[leftmargin=1.5em,noitemsep,topsep=2pt]
\item Prefer downgrading it to \texttt{Potential Task}.
\item Use placeholders such as \texttt{Needs verification}, \texttt{Variable}, \texttt{N/A}, or \texttt{null} when appropriate.
\item Do not fabricate missing information.
\end{enumerate}

\smallskip\noindent{\bfseries\normalsize Article 17. Implementability}\par\smallskip

\noindent Even if a paper used the dataset, a task must not be treated as a valid \texttt{Classical Task} unless the public data contains labels, events, or a verifiable derivation path sufficient to implement it.\par\smallskip

\medskip\noindent{\bfseries\normalsize Chapter 4. Roles}
\par\nopagebreak\noindent\rule{\linewidth}{0.4pt}\par\smallskip

\smallskip\noindent{\bfseries\normalsize Article 18. Administrator Responsibilities}\par\smallskip

\noindent Administrators are responsible for:\par\smallskip

\begin{enumerate}[leftmargin=1.5em,noitemsep,topsep=2pt]
\item Deciding dataset entry names.
\item Deciding whether to accept human revisions or rule updates in pull requests.
\item Deciding whether a kernel must be added or rewritten.
\end{enumerate}

\smallskip\noindent{\bfseries\normalsize Article 19. Human Reviewer Responsibilities}\par\smallskip

\noindent Human reviewers are responsible for:\par\smallskip

\begin{enumerate}[leftmargin=1.5em,noitemsep,topsep=2pt]
\item Reviewing whether a \texttt{Classical Task} is valid.
\item Reviewing whether fields match actual data and official material.
\item Deciding how to resolve conflicting information.
\item Filling \texttt{Manual Review} when appropriate.
\end{enumerate}

\smallskip\noindent{\bfseries\normalsize Article 20. LLM Review Script Responsibilities}\par\smallskip

\noindent LLM review scripts are responsible for:\par\smallskip

\begin{enumerate}[leftmargin=1.5em,noitemsep,topsep=2pt]
\item Pre-generating or pre-correcting documents under the rulebook.
\item Detecting likely conflicts, omissions, and invalid tasks.
\item Producing conservative, traceable suggestions without inventing raw labels.
\end{enumerate}

\par\bigskip
\noindent\rule{\linewidth}{0.8pt}\par\nopagebreak
\noindent{\bfseries\normalsize Part II. Specific Rules}\par\nopagebreak
\noindent\rule{\linewidth}{0.8pt}\par\smallskip

\medskip\noindent{\bfseries\normalsize Chapter 5. Meta Area}
\par\nopagebreak\noindent\rule{\linewidth}{0.4pt}\par\smallskip

\smallskip\noindent{\bfseries\normalsize Article 21. General Field Rule}\par\smallskip

\noindent \texttt{Meta Area} records identifying dataset metadata. Unless otherwise specified, fields should be concise, stable, and easy for programs to read.\par\smallskip

\smallskip\noindent{\bfseries\normalsize Article 22. \texttt{Dataset}}\par\smallskip

\begin{enumerate}[leftmargin=1.5em,noitemsep,topsep=2pt]
\item \texttt{Dataset} is the dataset identifier.
\item It should follow \texttt{<source\_name\_id>}.
\item Administrators decide this name.
\item Reviewers and LLMs must not proactively modify it.
\end{enumerate}

\smallskip\noindent{\bfseries\normalsize Article 23. \texttt{Year}}\par\smallskip

\begin{enumerate}[leftmargin=1.5em,noitemsep,topsep=2pt]
\item \texttt{Year} records key years related to collection, processing, or release.
\item It does not need to distinguish collection year from release year.
\item Multiple key years may be joined with \texttt{/}, for example \texttt{2004/2021}.
\item Continuous ranges may use \texttt{-}, for example \texttt{2008-2015/2017}.
\item The field need not cover every historical year, but should be updated as well as trusted sources allow.
\item Do not ignore key years clearly present in the data because a web page title shows a different year.
\end{enumerate}

\smallskip\noindent{\bfseries\normalsize Article 24. \texttt{URL}}\par\smallskip

\begin{enumerate}[leftmargin=1.5em,noitemsep,topsep=2pt]
\item \texttt{URL} should preferably point to an information page that gives access to downloads or documentation.
\item If both a direct file link and an information page exist, keep the information page.
\item If a link is dead and cannot be replaced with an authoritative information page, delete it or replace it with an accessible authoritative page.
\item The field does not need to be permanently valid, but must have practical review value at the time of review.
\end{enumerate}

\smallskip\noindent{\bfseries\normalsize Article 25. \texttt{Category}}\par\smallskip

\begin{enumerate}[leftmargin=1.5em,noitemsep,topsep=2pt]
\item \texttt{Category} must be exactly one of:
\begin{itemize}[leftmargin=1.2em,noitemsep,topsep=1pt]
\item \texttt{Complex Applied}
\item \texttt{High-level Mental}
\item \texttt{Action/Output}
\item \texttt{Sensory/Response}
\item \texttt{Biological State}
\item \texttt{Others}
\end{itemize}
\item The field is normally proposed by a program or preprocessing LLM.
\item Human reviewers should correct it only when the classification is clearly wrong.
\item Do not create category values outside the six listed values.
\end{enumerate}

\smallskip\noindent{\bfseries\normalsize Article 26. \texttt{Subjects}}\par\smallskip

\begin{enumerate}[leftmargin=1.5em,noitemsep,topsep=2pt]
\item \texttt{Subjects} is the integer number of valid subjects in the dataset.
\item If the official count differs from the actually usable count, prefer the actually usable count.
\item Record this conflict in \texttt{Data Quality Notes}.
\end{enumerate}

\smallskip\noindent{\bfseries\normalsize Article 27. \texttt{Files (Completed)}}\par\smallskip

\begin{enumerate}[leftmargin=1.5em,noitemsep,topsep=2pt]
\item \texttt{Files (Completed)} is the number of files parsed by EEGUnity as \texttt{Completed}.
\item LLMs and human reviewers normally must not edit this field manually.
\item If the parser result is clearly abnormal, report that the dataset needs a rescan or kernel correction instead of manually changing the count.
\end{enumerate}

\smallskip\noindent{\bfseries\normalsize Article 28. \texttt{Channels}}\par\smallskip

\begin{enumerate}[leftmargin=1.5em,noitemsep,topsep=2pt]
\item \texttt{Channels} is the number of channels used by files, including non-EEG channels.
\item If only a few channel counts exist and can be listed clearly, use forms such as \texttt{32/64} or \texttt{22-128}.
\item If the situation is complex and should not be fully enumerated, use \texttt{Variable}.
\item LLMs or reviewers may use the clearest representation, but must avoid ambiguity.
\end{enumerate}

\smallskip\noindent{\bfseries\normalsize Article 29. \texttt{Sampling Rate}}\par\smallskip

\begin{enumerate}[leftmargin=1.5em,noitemsep,topsep=2pt]
\item \texttt{Sampling Rate} records the main sampling-rate information.
\item If the whole dataset has one sampling rate, write a single value.
\item If files, modalities, or phases differ, use multiple values or explanatory wording.
\item If differences are complex, use \texttt{Variable} and explain in \texttt{Data Quality Notes}.
\end{enumerate}

\smallskip\noindent{\bfseries\normalsize Article 30. \texttt{Manual Review}}\par\smallskip

\begin{enumerate}[leftmargin=1.5em,noitemsep,topsep=2pt]
\item Only human reviewers may fill \texttt{Manual Review}.
\item The value may be a real name, alias, or reviewer-approved identifier.
\item Multiple reviewers should be separated by \texttt{/} in review order.
\item LLMs must not forge, add, or impersonate a \texttt{Manual Review} mark.
\item \texttt{Manual Review} must be inline at the end of the second meta line, after \texttt{Sampling Rate}, using \texttt{| **Manual Review**: ...}.
\item A separate third-line \texttt{Manual Review} field is no longer allowed.
\end{enumerate}

\smallskip\noindent{\bfseries\normalsize Article 30.1. Fixed Structure}\par\smallskip

\begin{enumerate}[leftmargin=1.5em,noitemsep,topsep=2pt]
\item The current template normally keeps only these top-level sections in this order:
\begin{itemize}[leftmargin=1.2em,noitemsep,topsep=1pt]
\item \texttt{\#\# Paradigm}
\item \texttt{\#\# File Scan Result}
\item \texttt{\#\# Tasks}
\item \texttt{\#\# Data Quality Notes}
\item \texttt{\#\# Kernel Responsibilities}
\end{itemize}
\item \texttt{\#\# File Scan Result} contains these third-level subsections in fixed order:
\begin{itemize}[leftmargin=1.2em,noitemsep,topsep=1pt]
\item \texttt{\#\#\# Event Structure} (required)
\item \texttt{\#\#\# MISC Label Summary} (optional)
\item \texttt{\#\#\# Info} (required)
\end{itemize}
\item Do not rename fixed section headings except for explicitly optional sections listed in this rulebook.
\item \texttt{STIM Label Summary} is deprecated and must not be added to new documents. Discrete events from trigger or stim channels must be registered in \texttt{\#\#\# Event Structure}.
\end{enumerate}

\smallskip\noindent{\bfseries\normalsize Article 30.2. No Extra Sections}\par\smallskip

\begin{enumerate}[leftmargin=1.5em,noitemsep,topsep=2pt]
\item Unless maintainers explicitly request it, do not add non-template top-level sections such as \texttt{\#\# Meta}, standalone \texttt{\#\# References}, or custom extension sections.
\item If extra explanation is necessary, prefer task \texttt{Notes}, \texttt{Data Quality Notes}, \texttt{\#\#\# Kernel Notes} under \texttt{\#\# Kernel Responsibilities}, or review records.
\item When substantially revising historical documents, remove extra sections that violate the current structure unless maintainers explicitly request otherwise.
\end{enumerate}

\smallskip\noindent{\bfseries\normalsize Article 30.3. Automated Behavior for \texttt{Manual Review}}\par\smallskip

\begin{enumerate}[leftmargin=1.5em,noitemsep,topsep=2pt]
\item LLM stages in generation pipelines must not fill \texttt{Manual Review}, including placeholders such as \texttt{Pending}, \texttt{TBD}, or \texttt{N/A}.
\item LLMs should output \texttt{| **Manual Review**:} at the end of the second meta line with nothing after the colon. If template reuse accidentally inserts a non-empty value, post-processing must clear it.
\item This field may be non-empty only when a human reviewer manually edits \texttt{analysis.md}; use Article 30 for formatting.
\end{enumerate}

\medskip\noindent{\bfseries\normalsize Chapter 6. Paradigm Area}
\par\nopagebreak\noindent\rule{\linewidth}{0.4pt}\par\smallskip

\smallskip\noindent{\bfseries\normalsize Article 31. General Paradigm Rule}\par\smallskip

\noindent \texttt{Paradigm Area} should describe the dataset`s experimental paradigm, task form, subject behavior, and recording method, and should match the actual data.\par\smallskip

\smallskip\noindent{\bfseries\normalsize Article 32. Consistency Requirements}\par\smallskip

\noindent Paradigm descriptions must align with:\par\smallskip

\begin{enumerate}[leftmargin=1.5em,noitemsep,topsep=2pt]
\item Data directories and file structure.
\item Event files.
\item Official pages and official documentation.
\item Authoritative papers that clearly use the dataset.
\end{enumerate}

\smallskip\noindent{\bfseries\normalsize Article 33. Data Priority}\par\smallskip

\noindent When online material conflicts with data material, the data prevails.\par\smallskip

\smallskip\noindent{\bfseries\normalsize Article 34. No Over-Inference}\par\smallskip

\noindent Paradigm descriptions must not turn a paper`s experimental goal, theoretical hypothesis, or common analysis style into a label task supposedly supported by the public data unless the data provides implementable events or labels.\par\smallskip

\smallskip\noindent{\bfseries\normalsize Article 35. Description Boundary}\par\smallskip

\noindent Paradigm descriptions may state the original experimental intent, but must distinguish:\par\smallskip

\begin{enumerate}[leftmargin=1.5em,noitemsep,topsep=2pt]
\item The goal of the experimental design.
\item The labels and tasks actually available in the public data.
\end{enumerate}

\medskip\noindent{\bfseries\normalsize Chapter 7. Kernel Responsibilities Area}
\par\nopagebreak\noindent\rule{\linewidth}{0.4pt}\par\smallskip

\smallskip\noindent{\bfseries\normalsize Article 36. Purpose and Structure}\par\smallskip

\begin{enumerate}[leftmargin=1.5em,noitemsep,topsep=2pt]
\item \texttt{Kernel Responsibilities Area} records the core semantic judgment for the dataset at the kernel level.
\item The section uses blockquote metadata fields. The current version keeps only \texttt{Super Meta Required} and \texttt{Status}, in that order.
\item Do not add other fields, custom blocks, or explanatory text in this section.
\item Historical fields such as Type 1/2/3 template text, \texttt{Selected Mode}, \texttt{Why}, \texttt{Primary Label Routes}, and \texttt{Web Reference Used for Decision} are deprecated and should be removed during substantial revisions.
\end{enumerate}

\smallskip\noindent{\bfseries\normalsize Article 37. \texttt{Super Meta Required}}\par\smallskip

\begin{enumerate}[leftmargin=1.5em,noitemsep,topsep=2pt]
\item \texttt{Super Meta Required} states whether the dataset has target labels that require extra derivation, pairing, folding, mapping, or cross-file joining.
\item Legal values are exactly:
\begin{itemize}[leftmargin=1.2em,noitemsep,topsep=1pt]
\item \texttt{Yes}: at least one task in \texttt{Tasks Area} has \texttt{Meta Level = super}.
\item \texttt{No}: all tasks have \texttt{Meta Level = raw}.
\end{itemize}
\item Do not use synonyms or case variants.
\end{enumerate}

\smallskip\noindent{\bfseries\normalsize Article 38. Deciding \texttt{Super Meta Required}}\par\smallskip

\begin{enumerate}[leftmargin=1.5em,noitemsep,topsep=2pt]
\item If any task, including \texttt{Classical Task} or \texttt{Potential Task}, has \texttt{Meta Level = super}, \texttt{Super Meta Required} must be \texttt{Yes}.
\item If all tasks have \texttt{Meta Level = raw}, \texttt{Super Meta Required} should be \texttt{No}.
\item If this field conflicts with \texttt{Tasks Area}, \texttt{Tasks Area} prevails and this field must be corrected.
\item LLM review scripts may infer and fill this field from \texttt{Tasks Area}.
\end{enumerate}

\smallskip\noindent{\bfseries\normalsize Article 42.1. \texttt{Status}}\par\smallskip

\begin{enumerate}[leftmargin=1.5em,noitemsep,topsep=2pt]
\item \texttt{Status} marks whether the current document text is sufficient to support later kernel refactoring, EEGUnity export, or event and label reconstruction.
\item \texttt{Status} is a dataset-level field, not a per-task field.
\item Although \texttt{Status} is metadata, it must currently be written in the \texttt{Kernel Responsibilities Area} blockquote, not in a new top-level section.
\item LLM review scripts, human reviewers, or maintainers may update \texttt{Status} according to the latest judgment.
\end{enumerate}

\smallskip\noindent{\bfseries\normalsize Article 42.2. Legal \texttt{Status} Values}\par\smallskip

\noindent \texttt{Status} may use only these four machine-readable values:\par\smallskip

\begin{enumerate}[leftmargin=1.5em,noitemsep,topsep=2pt]
\item \texttt{wait\_kernel\_update}
\item \texttt{wait\_manual\_review}
\item \texttt{completed}
\item \texttt{hallucination}
\end{enumerate}

\noindent Do not replace them with synonyms, natural-language phrases, or case variants.\par\smallskip

\smallskip\noindent{\bfseries\normalsize Article 42.2.1. Authority Boundary for \texttt{completed} and \texttt{hallucination}}\par\smallskip

\begin{enumerate}[leftmargin=1.5em,noitemsep,topsep=2pt]
\item Only human reviewers may assign \texttt{completed} or \texttt{hallucination}. LLM review scripts must not proactively assign either value.
\item At the current stage, LLM review scripts should choose only between \texttt{wait\_kernel\_update} and \texttt{wait\_manual\_review}, and should default conservatively to \texttt{wait\_manual\_review}.
\item If the current text is already sufficient to support existing tasks and does not expose a clear structural kernel gap, LLM review scripts should output \texttt{wait\_manual\_review} and wait for final human confirmation.
\item After human review is complete, the final status must be \texttt{completed} or \texttt{hallucination}; it must not remain \texttt{wait\_manual\_review} or \texttt{wait\_kernel\_update}.
\end{enumerate}

\smallskip\noindent{\bfseries\normalsize Article 42.2.2. Boundary for \texttt{hallucination}}\par\smallskip

\begin{enumerate}[leftmargin=1.5em,noitemsep,topsep=2pt]
\item \texttt{hallucination} is a severe document-level defect flag for cases where the entire \texttt{analysis.md} cannot be trusted.
\item Triggering conditions include, but are not limited to:
\begin{itemize}[leftmargin=1.2em,noitemsep,topsep=1pt]
\item The dataset URL is invalid and no key field can be verified.
\item The document describes a dataset that does not match the actual dataset.
\item Task definitions, \texttt{Event Structure}, and \texttt{Paradigm} have no credible basis and cannot be repaired by reasonable supplementation.
\item Dataset information is so incomplete that even experts cannot make a reliable judgment.
\end{itemize}
\item \texttt{hallucination} does not apply to local issues such as one incorrect task citation or partially inaccurate channel description. Local issues should be recorded in \texttt{Data Quality Notes}, and status should remain \texttt{completed} after human review.
\item For datasets marked \texttt{hallucination}, administrators will consider deleting the document or arranging a rescan. Reviewers are not required to keep repairing it.
\end{enumerate}

\smallskip\noindent{\bfseries\normalsize Article 42.3. Deciding \texttt{Status}}\par\smallskip

\begin{enumerate}[leftmargin=1.5em,noitemsep,topsep=2pt]
\item \texttt{wait\_kernel\_update} means the document clearly exposes a structural semantic kernel gap: \texttt{Event Structure}, \texttt{MISC Label Summary}, \texttt{Label Route}, companion-file ingestion, or \texttt{super meta} integration is not yet sufficient to support existing tasks stably. This is an intermediate state for LLM scripts or automatic scanning.
\item \texttt{wait\_manual\_review} means no clear structural kernel gap is exposed, or remaining doubts mainly require human confirmation of task validity, task retention, or final release. This is the default LLM output status.
\item \texttt{completed} means a human reviewer confirms that document review is finished. Known quality issues may still exist if they are recorded in \texttt{Data Quality Notes} and the overall document is trustworthy and ready to proceed. \texttt{completed} means the reviewer has done due diligence, not that the document is flawless.
\item \texttt{hallucination} means a human reviewer judges the whole document to have unrecoverable severe defects and that it should be handled by administrators.
\item The following must not, by themselves, justify \texttt{wait\_kernel\_update}:
\begin{itemize}[leftmargin=1.2em,noitemsep,topsep=1pt]
\item Routine preprocessing, filtering, rereferencing, or downsampling.
\item Artifact rejection, bad-segment removal, or bad-file exclusion.
\item General data quality control.
\item Implementation details such as training parameters, epoch length, or model hyperparameters.
\item Guessing from outside the document that the kernel might be unfinished or unverified.
\end{itemize}
\item If the document semantically explains relevant events, labels, companion files, or metadata access paths and presents tasks as currently valid, do not mark \texttt{wait\_kernel\_update} only because of implementation caution.
\item When a dataset has multiple issue types, judgment priority is: \texttt{hallucination} above all other values; \texttt{wait\_manual\_review} above \texttt{wait\_kernel\_update}; \texttt{wait\_kernel\_update} above \texttt{completed}.
\end{enumerate}

\smallskip\noindent{\bfseries\normalsize Article 42.4. Blockquote Position and Order}\par\smallskip

\begin{enumerate}[leftmargin=1.5em,noitemsep,topsep=2pt]
\item In the current version, blockquote metadata under \texttt{Kernel Responsibilities Area} must appear in this order:
\begin{itemize}[leftmargin=1.2em,noitemsep,topsep=1pt]
\item \texttt{Super Meta Required}
\item \texttt{Status}
\end{itemize}
\item Do not create separate top-level sections or custom blocks for these fields.
\item If a maintenance script batch-rewrites \texttt{Status}, it should modify only the fixed line for that field and should not rewrite unrelated body text.
\end{enumerate}

\medskip\noindent{\bfseries\normalsize Chapter 8. File Scan Result Area}
\par\nopagebreak\noindent\rule{\linewidth}{0.4pt}\par\smallskip

\noindent \texttt{\#\# File Scan Result} summarizes the kernel scan over raw data files, including kernel fingerprint, event structure, MISC labels, and subject information. It always contains \texttt{\#\#\# Event Structure} and \texttt{\#\#\# Info}; \texttt{\#\#\# MISC Label Summary} appears only when the dataset exposes misc-route labels or scanned native misc channels. The area begins with a kernel fingerprint blockquote.\par\smallskip

\smallskip\noindent{\bfseries\normalsize Article 42.5. Kernel Fingerprint}\par\smallskip

\begin{enumerate}[leftmargin=1.5em,noitemsep,topsep=2pt]
\item The start of \texttt{\#\# File Scan Result} should record the kernel fingerprint used to generate the scan result as a blockquote.
\item The reference format is:
\begin{quote}\ttfamily\footnotesize
> **Kernel SHA256**: `<full\_hash>`\\
> **EEGUnity Version**: `X.Y.Z`\\
> **Scan Type**: Full\\
\end{quote}
\item \texttt{Kernel SHA256} is the hash of the kernel file content and is calculated by maintenance scripts; it should not be filled manually. \texttt{EEGUnity Version} is the version used during scanning.
\item \texttt{Scan Type} is \texttt{Full} or \texttt{Sampled (N/M files)}.
\item If the dataset has no dataset-specific kernel, or scanning can be completed through the default/generic parser path without a dedicated kernel file, \texttt{Kernel SHA256} may be \texttt{N/A}.
\item Neurodoc maintains this blockquote automatically. Human reviewers and LLMs must not manually edit the hash value.
\item The frontend display layer treats all consecutive \texttt{>} lines at the beginning of \texttt{\#\# File Scan Result} as Kernel Fingerprint and renders them as one blockquote without parsing field names. This keeps the frontend independent of field changes.
\item Do not insert HTML comments such as \texttt{<!-- AUTO-KERNEL-... -->} inside \texttt{\#\#\# Event Structure} or other subsections. All kernel-derived information must be written directly in standard Markdown.
\end{enumerate}

\smallskip\noindent{\bfseries\normalsize Article 43. Table Form}\par\smallskip

\noindent \texttt{\#\#\# Event Structure} must be presented as a table. The table may be empty, but emptiness has rule consequences.\par\smallskip

\smallskip\noindent{\bfseries\normalsize Article 44. Required Columns}\par\smallskip

\noindent The table must contain at least:\par\smallskip

\begin{enumerate}[leftmargin=1.5em,noitemsep,topsep=2pt]
\item \texttt{Event ID}
\item \texttt{Count (total)}
\item \texttt{Type}
\item \texttt{Meta Level}
\item \texttt{Meaning}
\end{enumerate}

\smallskip\noindent{\bfseries\normalsize Article 45. Consequences of an Empty Table}\par\smallskip

\noindent When \texttt{Event Structure} is empty or explicitly states that no valid event labels exist:\par\smallskip

\begin{enumerate}[leftmargin=1.5em,noitemsep,topsep=2pt]
\item Do not keep \texttt{epoch\_by\_event\_hdf5} tasks.
\item Do not keep \texttt{epoch\_by\_long\_event\_hdf5} tasks.
\item Normally only \texttt{epoch\_by\_segmentation\_hdf5} may be considered.
\end{enumerate}

\smallskip\noindent{\bfseries\normalsize Article 46. Legal \texttt{Type} Values}\par\smallskip

\noindent \texttt{Type} may use only:\par\smallskip

\begin{enumerate}[leftmargin=1.5em,noitemsep,topsep=2pt]
\item \texttt{event}: discrete events or time spans that can be used for tasks, including stimuli, responses, state transitions, sleep stages, and other domain events.
\item \texttt{admin}: system, boundary, or excluded events, such as device markers, file boundaries, or motion-artifact markers.
\end{enumerate}

\noindent The historical \texttt{short}, \texttt{long}, and \texttt{rest} subdivisions are deprecated. Event duration behavior is determined by \texttt{Epoch Method} in \texttt{Tasks Area}, not repeated in \texttt{Event Structure}.\par\smallskip

\smallskip\noindent{\bfseries\normalsize Article 47. \texttt{admin} Events}\par\smallskip

\begin{enumerate}[leftmargin=1.5em,noitemsep,topsep=2pt]
\item \texttt{admin} events should remain explicitly listed in \texttt{Event Structure}; do not delete them by default.
\item \texttt{admin} events must not normally be used as \texttt{target\_events} for supervised-learning tasks.
\item Any exception requires strong evidence and a reason in task \texttt{Notes}.
\end{enumerate}

\smallskip\noindent{\bfseries\normalsize Article 48. Deciding \texttt{Meta Level}}\par\smallskip

\begin{enumerate}[leftmargin=1.5em,noitemsep,topsep=2pt]
\item Events that can be read directly from raw data are \texttt{raw}.
\item Events that require pairing, folding, derivation, mapping, joining, or reconstruction are \texttt{super}.
\end{enumerate}

\smallskip\noindent{\bfseries\normalsize Article 49. First-Pass Documentation}\par\smallskip

\noindent On first document creation, \texttt{Event Structure} should normally record only \texttt{raw meta}. If \texttt{super meta} is not yet supported by a stable kernel, do not pretend it is a raw event.\par\smallskip

\smallskip\noindent{\bfseries\normalsize Article 50. Handling Mismatches with Source Material}\par\smallskip

\noindent When \texttt{Event Structure} seriously disagrees with dataset documentation:\par\smallskip

\begin{enumerate}[leftmargin=1.5em,noitemsep,topsep=2pt]
\item If task validity is not affected, warn in \texttt{Data Quality Notes}.
\item If task validity is affected, correct event interpretation, downgrade tasks, delete tasks, or report that a rescan is needed.
\item Do not keep incorrect tasks without explaining the conflict.
\end{enumerate}

\smallskip\noindent{\bfseries\normalsize Article 51. Parser Differences}\par\smallskip

\noindent When mismatch is likely caused by kernel parser differences:\par\smallskip

\begin{enumerate}[leftmargin=1.5em,noitemsep,topsep=2pt]
\item If task validity is not affected, correction may be postponed.
\item If task validity is affected, clearly mark that a rescan or kernel correction is needed.
\end{enumerate}

\smallskip\noindent{\bfseries\normalsize Article 51.1. Linkage with \texttt{Status}}\par\smallskip

\begin{enumerate}[leftmargin=1.5em,noitemsep,topsep=2pt]
\item If \texttt{Event Structure}, \texttt{MISC Label Summary} when present, and \texttt{Tasks Area} already form a sufficiently consistent evidence chain, do not mark \texttt{wait\_kernel\_update} only because of preprocessing, artifact rejection, quality control, or implementation details.
\item If the document is already sufficient as textual support for existing \texttt{Classical Task} or \texttt{Potential Task} entries, but final release still requires administrator or human-reviewer confirmation, LLM review scripts should temporarily mark \texttt{wait\_manual\_review}.
\item If \texttt{Event Structure} cannot support existing tasks and the issue mainly concerns human judgment about task validity, event semantics, evidence conflict, label mapping, deletion, or downgrading, \texttt{Status} should normally be \texttt{wait\_manual\_review}.
\item \texttt{Status} may be \texttt{completed} only when \texttt{Event Structure}, relevant tasks, route explanations, and \texttt{MISC Label Summary} when present form a sufficiently consistent and traceable implementation basis, and a human reviewer or administrator has confirmed it.
\end{enumerate}

\smallskip\noindent{\bfseries\normalsize Article 52. Context-Dependent Events}\par\smallskip

\noindent If the same event code has different meanings across files, tasks, or experimental phases, it is a context-dependent event. For such events:\par\smallskip

\begin{enumerate}[leftmargin=1.5em,noitemsep,topsep=2pt]
\item Do not interpret their meaning without file-level context.
\item Warn explicitly in \texttt{Event Structure} or \texttt{Data Quality Notes}.
\item Without stable context rules, do not use them directly in high-confidence \texttt{Classical Task} entries.
\end{enumerate}

\smallskip\noindent{\bfseries\normalsize Article 53. Later Updates for \texttt{super meta}}\par\smallskip

\noindent When a \texttt{Classical Task} depends on \texttt{super meta}, human reviewers are normally not required to manually rewrite \texttt{Event Structure} first. Such updates should be handled by later kernel and LLM automation.\par\smallskip

\smallskip\noindent{\bfseries\normalsize Article 53.1. Optional Notes under \texttt{Event Structure}}\par\smallskip

\begin{enumerate}[leftmargin=1.5em,noitemsep,topsep=2pt]
\item An optional explanatory \texttt{Note} may appear after the \texttt{Event Structure} table, but is not required.
\item The note must not repeat the meanings of the \texttt{Type} enum values; \texttt{event} and \texttt{admin} are defined by this rulebook.
\item Use the note only when:
\begin{itemize}[leftmargin=1.2em,noitemsep,topsep=1pt]
\item Raw data has parsing ambiguity, such as the same event code meaning different things in different files.
\item Events come from multiple sources, such as companion files and trigger channels, and their merge logic matters.
\item Other explanations materially affect kernel implementation.
\end{itemize}
\item Do not use the note to introduce new events that are absent from the table.
\item If a later kernel update changes event representation, first update this note and related task routes, then decide whether to rewrite the main \texttt{Event Structure} table.
\item When neurodoc writes new scan results, any change to the \texttt{Event Structure} table, including count changes or event additions or removals, clears this note automatically. If the table is unchanged, the note remains unchanged.
\end{enumerate}

\smallskip\noindent{\bfseries\normalsize Article 53.2. Companion Files and External Tables}\par\smallskip

\begin{enumerate}[leftmargin=1.5em,noitemsep,topsep=2pt]
\item Labels, event intervals, or auxiliary fields directly present in official companion files, \texttt{participants.tsv}, behavioral tables, sleep-stage files, or other formal sidecars may normally be used as valid information sources.
\item If such information is a discrete event or time span and will be used for \texttt{target\_events} or event-route tasks, it must first be registered in \texttt{Event Structure}.
\item If such information is an epoch-level class, scalar, count, score, or continuous quantity, do not force it into \texttt{Event Structure}; route it through \texttt{MISC Label Summary} or task \texttt{Notes} as appropriate.
\item These fields are \texttt{raw} only when directly readable without extra derivation. If they require pairing, mapping, normalization, cross-file joining, or rule-based reconstruction, they are \texttt{super}.
\end{enumerate}

\smallskip\noindent{\bfseries\normalsize Article 54. General Rule}\par\smallskip

\noindent \texttt{\#\#\# MISC Label Summary} describes the source, meaning, and implementation of numeric or continuous auxiliary labels.\par\smallskip

\noindent \texttt{STIM Label Summary} is deprecated. Discrete events originally stored in trigger or stim channels must be registered in \texttt{\#\#\# Event Structure} instead.\par\smallskip

\smallskip\noindent{\bfseries\normalsize Article 55. When It Appears}\par\smallskip

\begin{enumerate}[leftmargin=1.5em,noitemsep,topsep=2pt]
\item If any task uses \texttt{misc} as \texttt{Label Route}, \texttt{\#\#\# MISC Label Summary} must exist.
\item If no task uses the \texttt{misc} route, this subsection may be omitted.
\item Native misc channels found during scanning may be registered here for future tasks even when no current task uses them.
\end{enumerate}

\smallskip\noindent{\bfseries\normalsize Article 56. Content Requirements}\par\smallskip

\noindent \texttt{\#\#\# MISC Label Summary} is a table with fixed columns \texttt{Channel | Source | Meta | Notes}:\par\smallskip

\begin{enumerate}[leftmargin=1.5em,noitemsep,topsep=2pt]
\item \texttt{Channel}: misc channel name, referenced by \texttt{Label Route = misc} tasks in the first column of \texttt{Target Definition}.
\item \texttt{Source}: field source, such as \texttt{native (raw misc in EEG file)}, \texttt{native (events.tsv column: response\_time)}, or \texttt{derived from participants.tsv: p\_factor}.
\item \texttt{Meta}: \texttt{raw} or \texttt{super}. Derived misc must be \texttt{super}; see Article 77.8.
\item \texttt{Notes}: units, alignment, derivation logic, fill rules, and other implementation notes.
\end{enumerate}

\smallskip\noindent{\bfseries\normalsize Article 57. Automation Principle}\par\smallskip

\noindent This section is normally generated by the kernel and LLM. Human reviewers should edit it only when it is clearly wrong, incomplete, or in conflict with task dependencies.\par\smallskip

\smallskip\noindent{\bfseries\normalsize Article 57.1. General Rule}\par\smallskip

\begin{enumerate}[leftmargin=1.5em,noitemsep,topsep=2pt]
\item \texttt{\#\#\# Info} records availability of file-level metadata exported by the kernel for \texttt{Label Route = info} tasks, typically from \texttt{raw.info["description"]["eegunity\_description"]}.
\item This subsection is required and must not be omitted. If information is entirely missing, still list the relevant fields and mark them \texttt{missing}.
\end{enumerate}

\smallskip\noindent{\bfseries\normalsize Article 57.2. Fields and Completeness}\par\smallskip

\begin{enumerate}[leftmargin=1.5em,noitemsep,topsep=2pt]
\item \texttt{\#\#\# Info} is a table with fixed columns \texttt{Field}, \texttt{Completeness}, and \texttt{Notes}.
\item \texttt{Field} names come from the kernel-exported \texttt{eegunity\_description} dictionary. The rulebook does not define a closed mandatory field list.
\item Common examples include \texttt{age}, \texttt{sex}, \texttt{device}, \texttt{handedness}, \texttt{group}, or other dataset-specific metadata keys. Any key exported by the kernel should be registered as-is and should appear only once.
\item If the kernel exports no usable info fields, the table may still contain placeholder rows such as \texttt{age}, \texttt{sex}, or \texttt{device}, marked \texttt{missing}, to make missingness explicit.
\item Any \texttt{Info Field} referenced by a \texttt{Label Route = info} task must be one of the fields registered in this table.
\end{enumerate}

\smallskip\noindent{\bfseries\normalsize Article 57.3. Legal \texttt{Completeness} Values}\par\smallskip

\noindent \texttt{Completeness} may use only:\par\smallskip

\begin{enumerate}[leftmargin=1.5em,noitemsep,topsep=2pt]
\item \texttt{complete}: every EEG record in the dataset has this field.
\item \texttt{partial (x/n)}: the field exists for \texttt{x} out of \texttt{n} scanned records.
\item \texttt{partial}: the field is known to be incomplete, but the stable scanned ratio is not recorded in the table.
\item \texttt{incomplete}: many missing values are known, but a stable ratio is not recorded.
\item \texttt{missing}: the field is entirely absent.
\end{enumerate}

\smallskip\noindent{\bfseries\normalsize Article 57.4. \texttt{Notes} Column}\par\smallskip

\begin{enumerate}[leftmargin=1.5em,noitemsep,topsep=2pt]
\item \texttt{Notes} is optional supplementary explanation, such as numeric range, enum values, or missingness reasons.
\item If no supplement is needed, write \texttt{N/A}.
\end{enumerate}

\smallskip\noindent{\bfseries\normalsize Article 57.5. Automation Principle}\par\smallskip

\begin{enumerate}[leftmargin=1.5em,noitemsep,topsep=2pt]
\item Field discovery and \texttt{Completeness} should normally be inferred automatically from kernel scan results, based on the metadata the kernel actually writes into \texttt{eegunity\_description}.
\item \texttt{Notes} may be assisted by the LLM during document generation, but the existence of fields and their completeness status should remain scan-derived.
\item Human reviewers should edit this subsection only when automatic filling is clearly wrong, stale, or semantically misleading.
\end{enumerate}

\medskip\noindent{\bfseries\normalsize Chapter 9. Retired}
\par\nopagebreak\noindent\rule{\linewidth}{0.4pt}\par\smallskip

\noindent The former Chapter 9, \texttt{MISC Label Summary}, has been merged into the \texttt{MISC Label Summary} articles under Chapter 8. This chapter number is retained as a placeholder and has no active content.\par\smallskip

\medskip\noindent{\bfseries\normalsize Chapter 10. Tasks Area}
\par\nopagebreak\noindent\rule{\linewidth}{0.4pt}\par\smallskip

\smallskip\noindent{\bfseries\normalsize Article 58. Role of the Tasks Area}\par\smallskip

\noindent \texttt{Tasks Area} is the most important executable area in the document. It should directly guide EEG epoching, label generation, and later implementation.\par\smallskip

\smallskip\noindent{\bfseries\normalsize Article 59. Task Headings}\par\smallskip

\noindent Task headings must use one of:\par\smallskip

\begin{enumerate}[leftmargin=1.5em,noitemsep,topsep=2pt]
\item \texttt{\#\#\# Classical Task N: ...}
\item \texttt{\#\#\# Potential Task N: ...}
\end{enumerate}

\smallskip\noindent{\bfseries\normalsize Article 60. Task Field Order}\par\smallskip

\noindent Each task should normally contain these fields in stable order:\par\smallskip

\begin{enumerate}[leftmargin=1.5em,noitemsep,topsep=2pt]
\item \texttt{Task Type}
\item \texttt{Epoch Method}
\item \texttt{Epoch Parameters}
\item \texttt{Label Route}
\item \texttt{Meta Level}
\item \texttt{Target Definition}
\item \texttt{Super Meta Notes}, if applicable
\item \texttt{Evidence}, required for \texttt{Classical Task}
\item \texttt{Notes}, as free supplementary text with no forced format. It may be used for super-meta derivation steps, parameter inference, review notes, and similar content. The old field name \texttt{Kernel Notes} is unified as \texttt{Notes} from v3 onward.
\end{enumerate}

\smallskip\noindent{\bfseries\normalsize Article 61. Historical Legacy Syntax}\par\smallskip

\noindent Some historical documents may write \texttt{segmentation} as \texttt{Task Type}. From this rulebook onward:\par\smallskip

\begin{enumerate}[leftmargin=1.5em,noitemsep,topsep=2pt]
\item \texttt{Task Type} should use only \texttt{classification} or \texttt{regression}.
\item \texttt{epoch\_by\_segmentation\_hdf5} is only an \texttt{Epoch Method}.
\item Historical documents need not be batch-cleaned immediately, but new documents and new revisions should follow this rule.
\end{enumerate}

\smallskip\noindent{\bfseries\normalsize Article 62. Conditions for \texttt{Classical Task}}\par\smallskip

\noindent A \texttt{Classical Task} is valid only when both conditions hold:\par\smallskip

\begin{enumerate}[leftmargin=1.5em,noitemsep,topsep=2pt]
\item Evidence shows that external literature or an authoritative source clearly used this dataset or an identifiable subset of it.
\item The evidence clearly supports the current task objective, label definition, or supervision target.
\end{enumerate}

\smallskip\noindent{\bfseries\normalsize Article 62.1. Explicitness Required for \texttt{Classical Task}}\par\smallskip

\begin{enumerate}[leftmargin=1.5em,noitemsep,topsep=2pt]
\item Key fields in a \texttt{Classical Task} must not use vague placeholders.
\item Key fields include at least:
\begin{itemize}[leftmargin=1.2em,noitemsep,topsep=1pt]
\item \texttt{Task Type}
\item \texttt{Epoch Method}
\item \texttt{Epoch Parameters}
\item \texttt{Label Route}
\item \texttt{Meta Level}
\item \texttt{Target Definition}
\item \texttt{Evidence}
\end{itemize}
\item Vague placeholders or vague styles include:
\begin{itemize}[leftmargin=1.2em,noitemsep,topsep=1pt]
\item \texttt{Needs verification}
\item \texttt{TBD}
\item \texttt{task-dependent}
\item \texttt{see Target Definition}
\item \texttt{Variable}
\item \texttt{N/A}
\item \texttt{unknown}
\item parameters marked with \texttt{*} or another undefined uncertainty marker
\item any broad description that cannot directly guide implementation
\end{itemize}
\item If there is not enough basis for explicit values, the task must not remain a \texttt{Classical Task}; downgrade it to \texttt{Potential Task}, mark it for review, or delete it.
\item LLMs and human reviewers may use judgment to choose the best explicit values for \texttt{Classical Task}.
\item The best explicit value should be chosen in this priority order:
\begin{itemize}[leftmargin=1.2em,noitemsep,topsep=1pt]
\item official page or official documentation recommendation
\item actual values used in papers that clearly use the dataset
\item reasonable values chosen by a reviewer or LLM based on domain knowledge
\end{itemize}
\item If the chosen value is not quoted directly from the official page or paper, briefly explain the judgment in task \texttt{Notes}.
\end{enumerate}

\smallskip\noindent{\bfseries\normalsize Article 63. When to Use \texttt{Potential Task}}\par\smallskip

\noindent Use \texttt{Potential Task} when:\par\smallskip

\begin{enumerate}[leftmargin=1.5em,noitemsep,topsep=2pt]
\item The data itself supports the task, but external evidence is insufficient.
\item Literature proves the dataset was used, but does not prove this exact task.
\item The task is a reasonable derived, recommended, or engineering task rather than the dataset`s recognized main task.
\end{enumerate}

\smallskip\noindent{\bfseries\normalsize Article 63.1. Minimum Explicitness for \texttt{Potential Task}}\par\smallskip

\begin{enumerate}[leftmargin=1.5em,noitemsep,topsep=2pt]
\item A \texttt{Potential Task} may be more conservative than a \texttt{Classical Task}, but must remain readable, reviewable, and implementable.
\item At minimum, \texttt{Task Type}, \texttt{Epoch Method}, \texttt{Label Route}, and \texttt{Meta Level} must have explicit legal values.
\item If required parameters for an \texttt{Epoch Method} are unclear, do not merge alternatives into one task using \texttt{or}, \texttt{task-dependent}, or \texttt{see Target Definition}. Split into multiple candidates, move the uncertainty to review notes, or delete the task.
\item If a task is only a possible research direction without a stable label source or epoch definition, do not keep it in \texttt{Tasks Area}; move it to review records or \texttt{Data Quality Notes}.
\end{enumerate}

\smallskip\noindent{\bfseries\normalsize Article 64. Downgrading}\par\smallskip

\noindent When an existing \texttt{Classical Task} no longer satisfies Article 62, downgrade it to \texttt{Potential Task} rather than keeping it as \texttt{Classical Task}.\par\smallskip

\smallskip\noindent{\bfseries\normalsize Article 65. Deletion}\par\smallskip

\noindent When a task depends on labels, events, or derivation paths that do not exist in the data and cannot be reconstructed by trusted stable rules, delete it.\par\smallskip

\smallskip\noindent{\bfseries\normalsize Article 66. Coexistence}\par\smallskip

\noindent One dataset may contain multiple \texttt{Classical Task} and multiple \texttt{Potential Task} entries, but each task must have independent and self-consistent support.\par\smallskip

\smallskip\noindent{\bfseries\normalsize Article 67. \texttt{Task Type}}\par\smallskip

\noindent \texttt{Task Type} may use only:\par\smallskip

\begin{enumerate}[leftmargin=1.5em,noitemsep,topsep=2pt]
\item \texttt{classification}
\item \texttt{regression}
\end{enumerate}

\smallskip\noindent{\bfseries\normalsize Article 68. \texttt{Epoch Method}}\par\smallskip

\noindent \texttt{Epoch Method} may use only:\par\smallskip

\begin{enumerate}[leftmargin=1.5em,noitemsep,topsep=2pt]
\item \texttt{epoch\_by\_event\_hdf5}
\item \texttt{epoch\_by\_long\_event\_hdf5}
\item \texttt{epoch\_by\_segmentation\_hdf5}
\end{enumerate}

\smallskip\noindent{\bfseries\normalsize Article 69. General Rule for \texttt{Epoch Parameters}}\par\smallskip

\noindent \texttt{Epoch Parameters} must include at least the parameters required by the selected \texttt{Epoch Method}; see Articles 70-72. The semantic rule \texttt{epoch\_parameters\_complete} in \texttt{detect.py} enforces this minimum set with severity \texttt{error} for both \texttt{Classical Task} and \texttt{Potential Task}.\par\smallskip

\noindent Additional optional parameters accepted by \texttt{MNE.Epoch}, such as \texttt{baseline}, \texttt{reject}, or \texttt{picks}, may be added as \texttt{key = value} pairs and are not constrained by \texttt{epoch\_parameters\_complete}.\par\smallskip

\smallskip\noindent{\bfseries\normalsize Article 70. Required Parameters for \texttt{epoch\_by\_event\_hdf5}}\par\smallskip

\noindent When using \texttt{epoch\_by\_event\_hdf5}, provide at least:\par\smallskip

\begin{enumerate}[leftmargin=1.5em,noitemsep,topsep=2pt]
\item \texttt{tmin}
\item \texttt{tmax}
\item \texttt{target\_events}
\end{enumerate}

\smallskip\noindent{\bfseries\normalsize Article 71. Required Parameters for \texttt{epoch\_by\_long\_event\_hdf5}}\par\smallskip

\noindent When using \texttt{epoch\_by\_long\_event\_hdf5}, provide at least:\par\smallskip

\begin{enumerate}[leftmargin=1.5em,noitemsep,topsep=2pt]
\item \texttt{tmin}
\item \texttt{tmax}
\item \texttt{overlap}
\item \texttt{target\_events}
\end{enumerate}

\smallskip\noindent{\bfseries\normalsize Article 72. Required Parameters for \texttt{epoch\_by\_segmentation\_hdf5}}\par\smallskip

\noindent When using \texttt{epoch\_by\_segmentation\_hdf5}, provide at least:\par\smallskip

\begin{enumerate}[leftmargin=1.5em,noitemsep,topsep=2pt]
\item \texttt{segment\_length}
\item \texttt{overlap}
\end{enumerate}

\smallskip\noindent{\bfseries\normalsize Article 73. Parameter Sources}\par\smallskip

\noindent \texttt{Epoch Parameters} should be supported by \texttt{Evidence} when possible. If evidence does not provide all parameters, human reviewers and LLMs may choose the best explicit values conservatively, using this priority:\par\smallskip

\begin{enumerate}[leftmargin=1.5em,noitemsep,topsep=2pt]
\item Recommended values from official pages or official documentation.
\item Actual values used in references that clearly use the dataset.
\item Reasonable values chosen by a reviewer or LLM based on domain knowledge.
\end{enumerate}

\noindent Filled parameters must be explicit and directly implementable; do not replace them with vague placeholders.\par\smallskip

\smallskip\noindent{\bfseries\normalsize Article 74. Valid Sources for \texttt{target\_events}}\par\smallskip

\noindent \texttt{target\_events} in \texttt{Epoch Parameters} may only come from entries in \texttt{Event Structure} that explicitly exist and have \texttt{Type = event}.\par\smallskip

\smallskip\noindent{\bfseries\normalsize Article 75. No Invented \texttt{target\_events}}\par\smallskip

\noindent If a task`s \texttt{target\_events} do not exist in \texttt{Event Structure}, or the corresponding entries have \texttt{Type = admin}, the task must not be treated as valid.\par\smallskip

\smallskip\noindent{\bfseries\normalsize Article 76. Human Review Handling}\par\smallskip

\noindent When \texttt{target\_events} and \texttt{Event Structure} are inconsistent, human reviewers should do one of:\par\smallskip

\begin{enumerate}[leftmargin=1.5em,noitemsep,topsep=2pt]
\item Adjust them to real existing labels.
\item Mark in the submission notes that a rescan is needed.
\item Delete the task.
\end{enumerate}

\smallskip\noindent{\bfseries\normalsize Article 77. \texttt{Label Route}}\par\smallskip

\begin{enumerate}[leftmargin=1.5em,noitemsep,topsep=2pt]
\item \texttt{Label Route} describes the source and storage form of the task`s supervision signal. Legal values are:
\begin{itemize}[leftmargin=1.2em,noitemsep,topsep=1pt]
\item \texttt{event}: supervision comes from the event stream, namely discrete events in \texttt{\#\#\# Event Structure}, regardless of whether raw storage was annotation or stim channel.
\item \texttt{misc}: supervision is numeric or continuous and stored in a misc channel registered in \texttt{\#\#\# MISC Label Summary}.
\item \texttt{info}: supervision comes from file-level subject or device properties registered in \texttt{\#\#\# Info}, such as sex, age, or group.
\end{itemize}
\item \texttt{Label Route}, \texttt{target\_events}, and \texttt{Target Definition} must be consistent:
\begin{itemize}[leftmargin=1.2em,noitemsep,topsep=1pt]
\item \texttt{Label Route = event}: \texttt{Target Definition} uses an \texttt{Event ID -> y} mapping table.
\item \texttt{Label Route = misc}: \texttt{Target Definition} uses two columns, \texttt{MISC Channel Name} and \texttt{Y Mapping Notes}.
\item \texttt{Label Route = info}: \texttt{Target Definition} uses two columns, \texttt{Info Field} and \texttt{Y Mapping Notes}; see Article 77.2.
\end{itemize}
\item For \texttt{misc} and \texttt{info} routes, the label source is uniquely specified by the first column of \texttt{Target Definition}; do not additionally declare \texttt{Label Field}.
\item If a task is an exception, explain the reason in task \texttt{Notes}.
\end{enumerate}

\smallskip\noindent{\bfseries\normalsize Article 77.1. \texttt{misc} Route Usage: Epoch-Level Scalars and Continuous Values}\par\smallskip

\begin{enumerate}[leftmargin=1.5em,noitemsep,topsep=2pt]
\item The \texttt{misc} route stores epoch-level numeric labels, continuous quantities, or categorical labels mapped from channel values, such as \texttt{response\_time}, ratings, density, duration, physiological indicators, or derived factor scores such as \texttt{p\_factor}.
\item Epoch anchor events, which define epoch boundaries, are independent from the supervision-signal route. \texttt{Label Route = misc} does not prevent \texttt{Epoch Method} from using event anchoring or segmentation.
\item \texttt{misc} sources are of two kinds:
\begin{itemize}[leftmargin=1.2em,noitemsep,topsep=1pt]
\item Native misc: a misc channel already present in raw files, or a numeric \texttt{events.tsv} column such as \texttt{response\_time} that can be directly read and filled per segmentation. \texttt{Meta Level = raw}.
\item Derived misc: a misc channel newly created by the kernel from external tables, such as \texttt{participants.tsv} or companion files, or from derivation logic. Values are filled per segmentation, with subject-level values constant across all segmentations for that subject. \texttt{Meta Level = super}; Article 77.8 applies.
\end{itemize}
\item One dataset may contain multiple misc channels, such as \texttt{misc:p\_factor}, \texttt{misc:externalizing}, or \texttt{misc:response\_time}. Each task may reference only one; that channel name must appear in the first column of \texttt{Target Definition}.
\item Tasks using this route require \texttt{\#\#\# MISC Label Summary} to register source, unit, alignment, and necessary derivation logic.
\end{enumerate}

\smallskip\noindent{\bfseries\normalsize Article 77.2. \texttt{Target Definition} Format for \texttt{Label Route = info}}\par\smallskip

\begin{enumerate}[leftmargin=1.5em,noitemsep,topsep=2pt]
\item When \texttt{Label Route = info}, \texttt{Target Definition} uses this format:
\begingroup\footnotesize
\begin{tabular}{@{}p{\dimexpr 0.88\linewidth/2\relax}p{\dimexpr 0.88\linewidth/2\relax}@{}}
\toprule
\textbf{Info Field} & \textbf{Y Mapping Notes} \\
\midrule
\texttt{sex} & Male -> 0; Female -> 1 \\
\bottomrule
\end{tabular}
\endgroup
\item Values in \texttt{Info Field} must match field names registered in \texttt{\#\#\# Info}.
\item If that field has \texttt{Completeness = missing}, \texttt{partial (x/n)}, or \texttt{incomplete}, the task cannot be a \texttt{Classical Task}; it may be at most a \texttt{Potential Task}, and task \texttt{Notes} should describe missingness.
\item The \texttt{info} route applies only to raw file-level fields directly exported by the kernel into \texttt{\#\#\# Info}, such as sex, age, device, or native group. If median split, normalization, pairing, cross-file joining, or derivation is needed, use derived misc instead; see Articles 77.1 and 77.8.
\item \texttt{Y Mapping Notes} may freely describe how original values map to \texttt{y}. If \texttt{y} directly follows the field value, write \texttt{no mapping}.
\item Each \texttt{info} route task may use only one source field; therefore \texttt{Target Definition} may contain only one row.
\end{enumerate}

\smallskip\noindent{\bfseries\normalsize Article 77.3. \texttt{Label Field} and the Single-Source Rule}\par\smallskip

\begin{enumerate}[leftmargin=1.5em,noitemsep,topsep=2pt]
\item \texttt{Label Route = event} tasks do not use \texttt{Label Field}.
\item \texttt{Label Route} in \texttt{\{misc, info\}} tasks also no longer declare \texttt{Label Field}; the label source is specified by the first column of \texttt{Target Definition}.
\item The first column of \texttt{Target Definition} must be locatable inside the document:
\begin{itemize}[leftmargin=1.2em,noitemsep,topsep=1pt]
\item \texttt{Label Route = misc}: it appears in the \texttt{Channel} column of \texttt{\#\#\# MISC Label Summary}.
\item \texttt{Label Route = info}: it appears in the \texttt{Field} column of \texttt{\#\#\# Info}.
\end{itemize}
\item For \texttt{misc} and \texttt{info}, the first column of \texttt{Target Definition} may contain only one unique value, and the table may contain only one row. This restriction does not apply to \texttt{event} tasks.
\end{enumerate}

\smallskip\noindent{\bfseries\normalsize Article 77.4. \texttt{Target Definition} Format Matrix}\par\smallskip

\noindent \texttt{Target Definition} table columns are determined by \texttt{Label Route x Task Type}:\par\smallskip

\begingroup\footnotesize
\begin{tabular}{@{}p{\dimexpr 0.88\linewidth/3\relax}p{\dimexpr 0.88\linewidth/3\relax}p{\dimexpr 0.88\linewidth/3\relax}@{}}
\toprule
\textbf{Label Route} & \textbf{Task Type} & \textbf{Header columns} \\
\midrule
\texttt{event} & \texttt{classification} & \texttt{\textbar{} Event ID \textbar{} y \textbar{} Meaning \textbar{}} \\
\texttt{event} & \texttt{regression} & \texttt{\textbar{} Event ID \textbar{} Value \textbar{} Meaning \textbar{}} (rare) \\
\texttt{misc} & \texttt{classification} & \texttt{\textbar{} MISC Channel Name \textbar{} Y Mapping Notes \textbar{}} \\
\texttt{misc} & \texttt{regression} & \texttt{\textbar{} MISC Channel Name \textbar{} Y Mapping Notes \textbar{}} \\
\texttt{info} & \texttt{classification} & \texttt{\textbar{} Info Field \textbar{} Y Mapping Notes \textbar{}} \\
\texttt{info} & \texttt{regression} & \texttt{\textbar{} Info Field \textbar{} Y Mapping Notes \textbar{}} \\
\bottomrule
\end{tabular}
\endgroup

\begin{enumerate}[leftmargin=1.5em,noitemsep,topsep=2pt]
\item Headers and column order must exactly match this matrix. LLMs and human reviewers must not add or remove columns.
\item For event routes, every \texttt{Event ID} cell must come from \texttt{\#\#\# Event Structure}; see Article 74.
\item For \texttt{misc} and \texttt{info} routes, \texttt{Y Mapping Notes} may be \texttt{no mapping} when \texttt{y} equals the raw field or channel value, or may freely describe a custom conversion.
\end{enumerate}

\smallskip\noindent{\bfseries\normalsize Article 77.5. Single-Target Task Rule}\par\smallskip

\begin{enumerate}[leftmargin=1.5em,noitemsep,topsep=2pt]
\item Each task encodes one learning target: one task equals one supervision route and one target schema.
\item Multidimensional labels, such as psychological four-factor scores \texttt{internalizing/externalizing/attention/p\_factor}, must be split into separate tasks. Each \texttt{misc} or \texttt{info} task may use only one \texttt{MISC Channel Name} or one \texttt{Info Field}.
\item \texttt{event} tasks may contain multiple \texttt{Event ID} rows when those rows jointly define a valid classification or regression mapping.
\item If one external evidence item covers multiple dimensions, reuse the same evidence in each split task instead of merging tasks.
\item For \texttt{misc} and \texttt{info}, the first column of \texttt{Target Definition}, namely \texttt{MISC Channel Name} or \texttt{Info Field}, may contain only one unique value within the task.
\item More complex composite tasks, such as a joint label built from multiple sources, may conceptually be expressed through combinations of simpler tasks. For consistency and brevity, this document currently defines no dedicated schema for complex tasks.
\end{enumerate}

\smallskip\noindent{\bfseries\normalsize Article 77.6. Mapping Requirements for the \texttt{misc} Route}\par\smallskip

\begin{enumerate}[leftmargin=1.5em,noitemsep,topsep=2pt]
\item \texttt{Label Route = misc} tasks may be \texttt{classification} or \texttt{regression}.
\item When \texttt{Task Type = classification} and the label comes from a misc channel, \texttt{Y Mapping Notes} must clearly explain how channel values map to class labels.
\item Thresholding, binning, median split, taking the first value, aggregation, or any other conversion on \texttt{misc} must be described in \texttt{Y Mapping Notes} or \texttt{Notes}.
\item If the misc channel itself is kernel-derived, Article 77.8 still applies.
\end{enumerate}

\smallskip\noindent{\bfseries\normalsize Article 77.7. Multiple-Class Requirement for \texttt{classification}}\par\smallskip

\begin{enumerate}[leftmargin=1.5em,noitemsep,topsep=2pt]
\item For \texttt{Task Type = classification}, the \texttt{y} column in \texttt{Target Definition} must contain at least two distinct values.
\item For \texttt{Label Route = event}, a \texttt{y} cell may be either one class id or a comma-separated list of class ids when one event is intentionally mapped to multiple categories.
\item If evidence or data supports only a single class, for example only a cue-onset marker with no contrast class, change the task to \texttt{regression} when applicable or downgrade/delete it. Do not keep a one-class classification task.
\end{enumerate}

\smallskip\noindent{\bfseries\normalsize Article 77.8. Derived Misc Implies Super Meta}\par\smallskip

\begin{enumerate}[leftmargin=1.5em,noitemsep,topsep=2pt]
\item Any misc channel newly created by a kernel is derived data. Any task carried by it must have \texttt{Meta Level = super}, and \texttt{Super Meta Required = Yes}; see Article 37.
\item Derived misc must be explicitly declared as derived in \texttt{\#\#\# MISC Label Summary}. It is recommended to describe derivation steps in task \texttt{Notes}, including source column, fill method per segmentation, units, and outlier handling. This recommendation is not enforced by \texttt{detect.py}; it is review and implementation guidance.
\item If the source field for derivation does not exist in the dataset, including \texttt{events.tsv}, \texttt{participants.tsv}, or native misc channels, the task is invalid and must be deleted.
\end{enumerate}

\smallskip\noindent{\bfseries\normalsize Article 78. \texttt{Meta Level}}\par\smallskip

\begin{enumerate}[leftmargin=1.5em,noitemsep,topsep=2pt]
\item \texttt{Meta Level = raw} means the task can be implemented from raw labels.
\item \texttt{Meta Level = super} means the task depends on additional derived, joined, or transformed labels.
\item If the label requires event pairing, class folding, subject-info mapping, standard-system mapping, or continuous-variable calculation, mark it \texttt{super}.
\end{enumerate}

\smallskip\noindent{\bfseries\normalsize Article 79. \texttt{Evidence}}\par\smallskip

\begin{enumerate}[leftmargin=1.5em,noitemsep,topsep=2pt]
\item \texttt{Evidence} presents external support for task validity or parameter choice.
\item \texttt{Classical Task} must provide \texttt{Evidence}.
\item \texttt{Potential Task} may provide \texttt{Evidence}, but it is not mandatory.
\item \texttt{Evidence} for \texttt{Classical Task} must be a specific locatable and identifiable source, not a vague placeholder.
\item If a DOI is unknown but an official page, competition page, or formal literature page is locatable, write that specific source instead of a vague phrase such as \texttt{DOI unknown}.
\end{enumerate}

\smallskip\noindent{\bfseries\normalsize Article 80. Evidence Review Focus}\par\smallskip

\noindent Human reviewers and LLMs should focus on:\par\smallskip

\begin{enumerate}[leftmargin=1.5em,noitemsep,topsep=2pt]
\item Whether the evidence clearly used this dataset.
\item Whether the evidence clearly supports this task rather than only a similar task.
\item Whether label definitions in the evidence match the public data.
\end{enumerate}

\smallskip\noindent{\bfseries\normalsize Article 81. Hallucination Prevention}\par\smallskip

\noindent Because LLMs may hallucinate paper interpretations, any case that cannot be confirmed as both using this dataset and supporting the task should be downgraded or deleted rather than forced to remain a \texttt{Classical Task}.\par\smallskip

\smallskip\noindent{\bfseries\normalsize Article 82. Task \texttt{Notes}}\par\smallskip

\noindent From v3 onward, \texttt{- **Notes**:} inside a task block, formerly \texttt{Kernel Notes}, is free supplementary explanation with no forced structure and no \texttt{detect.py} enforcement. Common uses include:\par\smallskip

\begin{enumerate}[leftmargin=1.5em,noitemsep,topsep=2pt]
\item Label mapping details, file-level context dependencies, or external table parsing requirements.
\item Event pairing, window-generation rules, or admin and pseudo-events to exclude.
\item Super-meta derivation steps; see Article 77.8.
\item Rationale for inferred \texttt{Epoch Parameters}; see Article 73.
\item Human judgment when \texttt{Classical Task} parameters differ from official or reference values.
\end{enumerate}

\noindent Task \texttt{Notes} and \texttt{\#\#\# Kernel Notes} under \texttt{\#\# Kernel Responsibilities} are different: the former is free annotation for a single task; the latter is implementation-level, cross-task kernel memo. They may both exist, but should not be conflated.\par\smallskip

\medskip\noindent{\bfseries\normalsize Chapter 11. Data Quality Notes Area}
\par\nopagebreak\noindent\rule{\linewidth}{0.4pt}\par\smallskip

\smallskip\noindent{\bfseries\normalsize Article 83. General Requirement}\par\smallskip

\noindent \texttt{Data Quality Notes Area} should use bullet points to record data quality issues, source conflicts, parsing risks, and use limitations.\par\smallskip
\noindent The \texttt{\#\# Data Quality Notes} heading must always be present. If there are no material notes to record, the area may be left empty; placeholder bullets are not required.\par\smallskip

\smallskip\noindent{\bfseries\normalsize Article 84. Cases to Record}\par\smallskip

\noindent Record these cases when possible:\par\smallskip

\begin{enumerate}[leftmargin=1.5em,noitemsep,topsep=2pt]
\item Official material conflicts with actual data.
\item Data are missing, excluded, empty, or lack events or fields.
\item Subject count, file count, channel count, or sampling rate differs from official description.
\item Event semantics depend on file-level context.
\item Some tasks are valid only for a subset of files.
\end{enumerate}

\smallskip\noindent{\bfseries\normalsize Article 85. Stance Requirement}\par\smallskip

\noindent When official material conflicts with real data, explicitly state that this document follows the real data.\par\smallskip

\smallskip\noindent{\bfseries\normalsize Article 86. No Exhaustiveness Requirement}\par\smallskip

\noindent This area does not need to list every issue, but human reviewers and LLMs should try to record quality issues that materially affect task validity.\par\smallskip

\medskip\noindent{\bfseries\normalsize Chapter 12. References Area, Deprecated}
\par\nopagebreak\noindent\rule{\linewidth}{0.4pt}\par\smallskip

\smallskip\noindent{\bfseries\normalsize Article 87. Deprecation}\par\smallskip

\begin{enumerate}[leftmargin=1.5em,noitemsep,topsep=2pt]
\item \texttt{References Area} is a legacy section from early document versions.
\item Current \texttt{analysis.md} documents should normally not keep a standalone \texttt{References Area}.
\item Sources directly related to a task should be written under \texttt{Evidence} in the corresponding \texttt{Classical Task} or \texttt{Potential Task}.
\item Sources related to data conflicts, quality issues, or review decisions should be written in the relevant note or review record, not separately summarized as \texttt{References Area}.
\end{enumerate}

\smallskip\noindent{\bfseries\normalsize Article 88. Transition Rule}\par\smallskip

\begin{enumerate}[leftmargin=1.5em,noitemsep,topsep=2pt]
\item Historical documents may temporarily keep \texttt{References Area}.
\item Once a document is substantially revised, remove that section unless maintainers explicitly request otherwise.
\item LLMs should not add \texttt{References Area} when generating new versions.
\end{enumerate}

\medskip\noindent{\bfseries\normalsize Chapter 12.1. STIM Label Summary Transition Rule}
\par\nopagebreak\noindent\rule{\linewidth}{0.4pt}\par\smallskip

\smallskip\noindent{\bfseries\normalsize Article 88.1. Deprecation}\par\smallskip

\begin{enumerate}[leftmargin=1.5em,noitemsep,topsep=2pt]
\item \texttt{STIM Label Summary} is a v0.1 legacy section and is deprecated from v0.2 onward.
\item All discrete events from trigger or stim channels should be registered in \texttt{Event Structure}, with \texttt{Type} set to \texttt{event} or \texttt{admin}.
\item Deprecated \texttt{Event Structure.Type} values \texttt{short}, \texttt{long}, and \texttt{rest} are also no longer used from v0.2 onward. Their information is now represented by \texttt{Epoch Method} in \texttt{Tasks Area}.
\end{enumerate}

\smallskip\noindent{\bfseries\normalsize Article 88.2. Existing-Document Transition}\par\smallskip

\begin{enumerate}[leftmargin=1.5em,noitemsep,topsep=2pt]
\item Existing historical \texttt{STIM Label Summary} sections do not have to be migrated immediately.
\item Once a document is substantially revised, migrate discrete events from \texttt{STIM Label Summary} into \texttt{Event Structure} and delete \texttt{STIM Label Summary}.
\item Existing \texttt{Event Structure} rows with \texttt{Type} equal to \texttt{short}, \texttt{long}, or \texttt{rest} should be rewritten to \texttt{event} during substantial revision; \texttt{admin} remains unchanged.
\item LLM review scripts must not add \texttt{STIM Label Summary} or use deprecated \texttt{Event Structure.Type} enum values when generating or revising documents.
\end{enumerate}

\par\bigskip
\noindent\rule{\linewidth}{0.8pt}\par\nopagebreak
\noindent{\bfseries\normalsize Part III. Supplementary Rules}\par\nopagebreak
\noindent\rule{\linewidth}{0.8pt}\par\smallskip

\medskip\noindent{\bfseries\normalsize Chapter 13. Placeholders, Format, and Style}
\par\nopagebreak\noindent\rule{\linewidth}{0.4pt}\par\smallskip

\smallskip\noindent{\bfseries\normalsize Article 90. Allowed Placeholders}\par\smallskip

\noindent When specific information cannot be determined, use:\par\smallskip

\begin{enumerate}[leftmargin=1.5em,noitemsep,topsep=2pt]
\item \texttt{Needs verification}
\item \texttt{Variable}
\item \texttt{N/A}
\item \texttt{null}
\end{enumerate}

\noindent These placeholders should normally not be used in key fields of \texttt{Classical Task}.\par\smallskip

\smallskip\noindent{\bfseries\normalsize Article 91. Preserve Raw Names}\par\smallskip

\noindent When raw event names, raw code values, or raw labels can be used directly, preserve them as-is. Do not rename them merely for language polish.\par\smallskip

\smallskip\noindent{\bfseries\normalsize Article 92. Style}\par\smallskip

\noindent Documents should be technical, factual, and executable. Avoid promotional, exaggerated, or unverifiable conclusions.\par\smallskip

\medskip\noindent{\bfseries\normalsize Chapter 14. Review Action Types}
\par\nopagebreak\noindent\rule{\linewidth}{0.4pt}\par\smallskip

\smallskip\noindent{\bfseries\normalsize Article 93. Review Actions}\par\smallskip

\noindent Later human and LLM reviews may use these action categories:\par\smallskip

\begin{enumerate}[leftmargin=1.5em,noitemsep,topsep=2pt]
\item \texttt{keep}
\item \texttt{correct}
\item \texttt{downgrade}
\item \texttt{delete}
\item \texttt{mark Needs verification}
\item \texttt{mark rescan required}
\item \texttt{mark kernel correction required}
\end{enumerate}

\smallskip\noindent{\bfseries\normalsize Article 94. Action Use}\par\smallskip

\begin{enumerate}[leftmargin=1.5em,noitemsep,topsep=2pt]
\item If a field adjustment can solve the problem, prefer \texttt{correct}.
\item If task evidence is insufficient but the data support the task, prefer \texttt{downgrade}.
\item If the label source does not exist or cannot be reconstructed stably, use \texttt{delete}.
\item If the parsing chain has an obvious programmatic issue, mark \texttt{rescan required} or \texttt{kernel correction required}.
\end{enumerate}
\normalsize
\end{tcolorbox}

%% file: appendix_semantic_rules.tex
\section{Machine-checkable Subset of the Entry Rulebook}
\label{appendix:machine_rule_subset}

\noindent This appendix summarizes a programmatic encoding of machine-checkable provisions from the shared entry rulebook, as used by \texttt{detect.py} to check \texttt{analysis.md} and kernel files. Check types include \texttt{section\_exists}, \texttt{pattern}, \texttt{field\_value}, and \texttt{cross\_check}.\par\smallskip
\begin{tcolorbox}[breakable,enhanced,colback=white,colframe=gray!50,boxrule=0.8pt,arc=4pt,left=8pt,right=8pt,top=6pt,bottom=6pt,title={\normalsize\bfseries NeuroDoc Semantic Rulebook},coltitle=black,colbacktitle=gray!15,titlerule=0.5pt]
\noindent\rule{\linewidth}{0.4pt}\par\smallskip
\noindent{\bfseries\small Hard Constraints (HC)}
{\footnotesize\ ---\ enforced in generation pipeline; not detectable by \texttt{detect.py}}\par\smallskip
\noindent{\footnotesize\textbf{HC-1.}~Data fields in File Scan Result (Event ID, Count, Info table values, Kernel SHA256 fingerprint) must come from kernel scan program output; LLM must not invent or alter these fields. Semantic fields (Event Type, Meaning) are generated by LLM in an isolated inference step but must be based on actual evidence from dataset files; fabrication is not allowed.}\par\smallskip
\noindent{\footnotesize\textbf{HC-2.}~Event Meaning and Type inference must be based on actual evidence in the dataset (events.tsv, sidecar JSON, README, etc.). LLM may consult any context information, but must not use unrelated reference papers, domain conventions, or BCI competition code systems as inference basis (e.g., must not infer four motor-imagery classes solely because event IDs are 1-4). Once an inference result is fixed it becomes a locked value; downstream pipeline must not modify it.}\par\smallskip
\noindent{\footnotesize\textbf{HC-3.}~Meta Area fields (Channels, Sampling Rate) should be cross-referenced against the locator cache CSV. LLM may fill narrative gaps but must not override program-derived values.}\par\smallskip
\noindent{\footnotesize\textbf{HC-4.}~LLM or automation may only set Status to: wait\_manual\_review -- default output when no explicit structural kernel gap is exposed wait\_kernel\_update -- explicit structural kernel gap or scan failure "completed" and "hallucination" are reserved for human reviewers only.}\par\smallskip
\noindent{\footnotesize\textbf{HC-5.}~Kernel must NOT remap original event IDs. Prefix-based enrichment is allowed (e.g., subject\_attr-event\_id, session\_attr-event\_id). Original event IDs must pass through unchanged.}\par\smallskip
\noindent{\footnotesize\textbf{HC-6.}~Kernel channel remapping is only allowed when explicit evidence exists in the dataset directory (companion file, README, sidecar) mapping ambiguous/numeric channels to named channels. Add a comment in kernel citing the evidence file.}\par\smallskip
\noindent{\footnotesize\textbf{HC-7.}~Full scan count should be minimized to avoid re-scanning the same dataset. The pipeline should run a consistency check before final output to ensure document content matches scan results.}\par\smallskip
\noindent\rule{\linewidth}{0.4pt}\par\smallskip
\noindent{\bfseries\small Semantic Check Rules}\par\smallskip
\smallskip\noindent{\footnotesize\bfseries Group~A:~Document Structure}\par\smallskip
\begingroup\footnotesize
\begin{tabular}{@{}p{0.22\linewidth}ll p{0.46\linewidth}@{}}
\toprule
\textbf{Rule ID} & \textbf{Sev.} & \textbf{Target} & \textbf{Description} \\
\midrule
\texttt{has\_\allowbreak{}paradigm} & \textbf{[E]} & \texttt{analysis\_md} & analysis.md must contain \#\# Paradigm section. \\
\addlinespace[1pt]
\texttt{has\_\allowbreak{}file\_\allowbreak{}scan\_\allowbreak{}result} & \textbf{[E]} & \texttt{analysis\_md} & analysis.md must contain \#\# File Scan Result section. \\
\addlinespace[1pt]
\texttt{has\_\allowbreak{}tasks} & \textbf{[E]} & \texttt{analysis\_md} & analysis.md must contain \#\# Tasks section. \\
\addlinespace[1pt]
\texttt{has\_\allowbreak{}data\_\allowbreak{}quality\_\allowbreak{}notes} & \textbf{[E]} & \texttt{analysis\_md} & analysis.md must contain \#\# Data Quality Notes section. \\
\addlinespace[1pt]
\texttt{has\_\allowbreak{}kernel\_\allowbreak{}responsibilities} & \textbf{[E]} & \texttt{analysis\_md} & analysis.md must contain \#\# Kernel Responsibilities section. \\
\addlinespace[1pt]
\texttt{has\_\allowbreak{}event\_\allowbreak{}structure} & \textbf{[E]} & \texttt{analysis\_md} & \#\# File Scan Result must contain \#\#\# Event Structure subsection. \\
\addlinespace[1pt]
\texttt{has\_\allowbreak{}info\_\allowbreak{}subsection} & \textbf{[E]} & \texttt{analysis\_md} & \#\# File Scan Result must contain \#\#\# Info subsection (rulebook Article 57.1). \\
\addlinespace[1pt]
\texttt{no\_\allowbreak{}stim\_\allowbreak{}label\_\allowbreak{}summary} & \textbf{[E]} & \texttt{analysis\_md} & STIM Label Summary is a deprecated legacy section and must not appear. \\
\addlinespace[1pt]
\texttt{section\_\allowbreak{}order\_\allowbreak{}correct} & \textbf{[E]} & \texttt{analysis\_md} & Top-level sections must follow order: Paradigm -> File Scan Result -> Tasks -> Data Quality Notes -> Kernel Responsibilities. \\
\addlinespace[1pt]
\texttt{no\_\allowbreak{}extra\_\allowbreak{}top\_\allowbreak{}level\_\allowbreak{}sections} & \textbf{[E]} & \texttt{analysis\_md} & Only the 5 required \#\# sections are allowed. No extra top-level headings (rulebook Article 30.2). \\
\addlinespace[1pt]
\bottomrule
\end{tabular}
\endgroup
\smallskip\noindent{\footnotesize\bfseries Group~B:~Field Legal Values}\par\smallskip
\begingroup\footnotesize
\begin{tabular}{@{}p{0.22\linewidth}ll p{0.46\linewidth}@{}}
\toprule
\textbf{Rule ID} & \textbf{Sev.} & \textbf{Target} & \textbf{Description} \\
\midrule
\texttt{status\_\allowbreak{}valid\_\allowbreak{}value} & \textbf{[E]} & \texttt{analysis\_md} & Status must be one of: wait\_kernel\_update, wait\_manual\_review, completed, hallucination (rulebook Article 42.2). \\
\addlinespace[1pt]
\texttt{super\_\allowbreak{}meta\_\allowbreak{}required\_\allowbreak{}valid\_\allowbreak{}value} & \textbf{[E]} & \texttt{analysis\_md} & Super Meta Required must be exactly `Yes` or `No` (rulebook Article 37). \\
\addlinespace[1pt]
\texttt{category\_\allowbreak{}valid\_\allowbreak{}value} & \textbf{[E]} & \texttt{analysis\_md} & Category must be one of 6 valid values (rulebook Article 25): Complex Applied / High-level Mental / Action/Output / Sensory/Response / Biological State / Others. \\
\addlinespace[1pt]
\texttt{label\_\allowbreak{}route\_\allowbreak{}no\_\allowbreak{}deprecated} & \textbf{[E]} & \texttt{analysis\_md} & Label Route must use current values event, misc, or info. Historical values Annotation/STIM/MISC are not allowed (rulebook Article 77). \\
\addlinespace[1pt]
\texttt{event\_\allowbreak{}type\_\allowbreak{}no\_\allowbreak{}deprecated} & \textbf{[E]} & \texttt{analysis\_md} & Event Structure Type must use event or admin only. Historical short/long/rest values are not allowed (rulebook Article 46). \\
\addlinespace[1pt]
\texttt{kernel\_\allowbreak{}fingerprint\_\allowbreak{}exists} & \textbf{[E]} & \texttt{analysis\_md} & \#\# File Scan Result must begin with a kernel fingerprint blockquote containing Kernel SHA256 (rulebook Article 42.5). \\
\addlinespace[1pt]
\texttt{no\_\allowbreak{}placeholder\_\allowbreak{}doi} & \textbf{[W]} & \texttt{analysis\_md} & Evidence DOIs must not be placeholder strings (TODO / N/A / none / unknown / placeholder). \\
\addlinespace[1pt]
\bottomrule
\end{tabular}
\endgroup
\smallskip\noindent{\footnotesize\bfseries Group~C:~Semantic Consistency}\par\smallskip
\begingroup\footnotesize
\begin{tabular}{@{}p{0.22\linewidth}ll p{0.46\linewidth}@{}}
\toprule
\textbf{Rule ID} & \textbf{Sev.} & \textbf{Target} & \textbf{Description} \\
\midrule
\texttt{super\_\allowbreak{}meta\_\allowbreak{}matches\_\allowbreak{}tasks} & \textbf{[E]} & \texttt{analysis\_md} & Super Meta Required must match actual Meta Level in Tasks: Yes if any task has Meta Level=super, No if all raw (rulebook Article 37-38). \\
\addlinespace[1pt]
\texttt{target\_\allowbreak{}events\_\allowbreak{}in\_\allowbreak{}event\_\allowbreak{}structure} & \textbf{[E]} & \texttt{analysis\_md} & All event IDs referenced in Tasks Target Definition tables must appear in \#\#\# Event Structure. \\
\addlinespace[1pt]
\texttt{classical\_\allowbreak{}task\_\allowbreak{}has\_\allowbreak{}evidence} & \textbf{[E]} & \texttt{analysis\_md} & Every Classical Task must have a non-empty Evidence field (rulebook Articles 62 and 79). \\
\addlinespace[1pt]
\texttt{info\_\allowbreak{}completeness\_\allowbreak{}valid} & \textbf{[E]} & \texttt{analysis\_md} & \#\#\# Info table Completeness column must use: complete, partial (x/n), incomplete, or missing (rulebook Article 57.3). \\
\addlinespace[1pt]
\texttt{kernel\_\allowbreak{}sha256\_\allowbreak{}matches\_\allowbreak{}file} & \textbf{[E]} & \texttt{analysis\_md} & If Kernel SHA256 is a concrete hash, it must match the effective kernel file used for scanning. \texttt{N/A} is allowed when scanning does not require a dataset-specific kernel file (for example generic/default parser cases). \\
\addlinespace[1pt]
\texttt{no\_\allowbreak{}auto\_\allowbreak{}removed\_\allowbreak{}tasks} & \textbf{[W]} & \texttt{analysis\_md} & Tasks section must not contain [AUTO-REMOVED] blocks. These indicate guard-removed hallucinations; dataset needs regeneration. \\
\addlinespace[1pt]
\texttt{no\_\allowbreak{}mi\_\allowbreak{}label\_\allowbreak{}on\_\allowbreak{}phase\_\allowbreak{}marker} & \textbf{[E]} & \texttt{analysis\_md} & Event Structure must not assign MI class labels to events whose Meaning is a trial phase marker (rest/ready/task/interval). See HC-2. \\
\addlinespace[1pt]
\texttt{no\_\allowbreak{}hallucinated\_\allowbreak{}bcic\_\allowbreak{}classes} & \textbf{[W]} & \texttt{analysis\_md} & Must not assign BCIC IV 2a class codes (769/770/771/772) as event labels without explicit sidecar evidence. \\
\addlinespace[1pt]
\texttt{kr\_\allowbreak{}has\_\allowbreak{}super\_\allowbreak{}meta} & \textbf{[E]} & \texttt{analysis\_md} & \#\# Kernel Responsibilities must contain > \textbf{Super Meta Required}: blockquote field (rulebook Articles 36-38). \\
\addlinespace[1pt]
\texttt{kr\_\allowbreak{}has\_\allowbreak{}status} & \textbf{[E]} & \texttt{analysis\_md} & \#\# Kernel Responsibilities must contain > \textbf{Status}: blockquote field (rulebook Articles 36 and 42.1). \\
\addlinespace[1pt]
\texttt{kr\_\allowbreak{}only\_\allowbreak{}blockquote\_\allowbreak{}fields} & \textbf{[W]} & \texttt{analysis\_md} & \#\# Kernel Responsibilities must contain only the two blockquote fields (Super Meta Required, Status), no extra prose (rulebook Articles 36 and 42.4). \\
\addlinespace[1pt]
\texttt{dqn\_\allowbreak{}bullet\_\allowbreak{}format} & \textbf{[W]} & \texttt{analysis\_md} & \#\# Data Quality Notes content must use `- ` bullet format (rulebook Article 83). \\
\addlinespace[1pt]
\texttt{task\_\allowbreak{}type\_\allowbreak{}no\_\allowbreak{}normal} & \textbf{[E]} & \texttt{analysis\_md} & Task headings must be `Classical Task` or `Potential Task` only. `Normal Task` is a legacy label and is not allowed. \\
\addlinespace[1pt]
\texttt{task\_\allowbreak{}order\_\allowbreak{}classical\_\allowbreak{}first} & \textbf{[E]} & \texttt{analysis\_md} & All Classical Tasks must appear before all Potential Tasks in \#\# Tasks section. \\
\addlinespace[1pt]
\texttt{task\_\allowbreak{}numbering\_\allowbreak{}from\_\allowbreak{}one} & \textbf{[E]} & \texttt{analysis\_md} & Task numbering must start from 1 within each type and increase consecutively. \\
\addlinespace[1pt]
\texttt{task\_\allowbreak{}dataset\_\allowbreak{}dispatchable} & \textbf{[E]} & \texttt{analysis\_md} & Every Task block must be accepted by dataset\_api.build\_dataset\_from\_task(dry\_run=True). \\
\addlinespace[1pt]
\texttt{target\_\allowbreak{}def\_\allowbreak{}schema\_\allowbreak{}matches\_\allowbreak{}route} & \textbf{[E]} & \texttt{analysis\_md} & Target Definition table header must match the Label Route x Task Type matrix (rulebook Article 77.4). \\
\addlinespace[1pt]
\texttt{single\_\allowbreak{}target\_\allowbreak{}per\_\allowbreak{}task} & \textbf{[E]} & \texttt{analysis\_md} & For misc/info tasks, the first Target Definition column must contain exactly one unique source value. This rule does not restrict event tasks with multiple Event ID rows (rulebook Articles 77.3 and 77.5). \\
\addlinespace[1pt]
\texttt{classification\_\allowbreak{}task\_\allowbreak{}has\_\allowbreak{}multiple\_\allowbreak{}classes} & \textbf{[E]} & \texttt{analysis\_md} & Classification tasks must have >= 2 distinct y values in Target Definition (rulebook Article 77.7). \\
\addlinespace[1pt]
\texttt{classification\_\allowbreak{}y\_\allowbreak{}zero\_\allowbreak{}indexed\_\allowbreak{}sequential} & \textbf{[E]} & \texttt{analysis\_md} & Event-route classification tasks must use 0-based sequential integer y values: the distinct y set must equal set(range(N)). Many-to-one mapping is allowed (e.g. A->0, B->1, C->1 is valid; \{0,1,3\} or \{1,2\} are not). Misc/info routes use Y Mapping Notes instead of a y column. \\
\addlinespace[1pt]
\texttt{derived\_\allowbreak{}misc\_\allowbreak{}requires\_\allowbreak{}super\_\allowbreak{}meta} & \textbf{[E]} & \texttt{analysis\_md} & Tasks with Label Route = misc whose Target Definition source is not listed as native in MISC Label Summary must have Meta Level = super (rulebook Article 77.8). \\
\addlinespace[1pt]
\texttt{segmentation\_\allowbreak{}requires\_\allowbreak{}empty\_\allowbreak{}target\_\allowbreak{}events} & \textbf{[E]} & \texttt{analysis\_md} & Epoch Method = epoch\_by\_segmentation\_hdf5 requires target\_events = [] (rulebook Articles 45 and 72). \\
\addlinespace[1pt]
\texttt{epoch\_\allowbreak{}parameters\_\allowbreak{}complete} & \textbf{[E]} & \texttt{analysis\_md} & Epoch Parameters must list the minimum required keys for the declared Epoch Method (rulebook Articles 69-72). \\
\addlinespace[1pt]
\texttt{manual\_\allowbreak{}review\_\allowbreak{}blank} & \textbf{[W]} & \texttt{analysis\_md} & Completed documents must record at least one reviewer name. \texttt{wait\_manual\_review} documents may be blank or may retain reviewer names from later manual rework. \\
\addlinespace[1pt]
\bottomrule
\end{tabular}
\endgroup
\smallskip\noindent{\footnotesize\bfseries Group~D:~Kernel Code Constraints}\par\smallskip
\begingroup\footnotesize
\begin{tabular}{@{}p{0.22\linewidth}ll p{0.46\linewidth}@{}}
\toprule
\textbf{Rule ID} & \textbf{Sev.} & \textbf{Target} & \textbf{Description} \\
\midrule
\texttt{kernel\_\allowbreak{}has\_\allowbreak{}kernel\_\allowbreak{}id} & \textbf{[E]} & \texttt{kernel} & Every kernel file must define KERNEL\_ID. \\
\addlinespace[1pt]
\texttt{kernel\_\allowbreak{}has\_\allowbreak{}apply} & \textbf{[E]} & \texttt{kernel} & Every kernel file must implement apply(). \\
\addlinespace[1pt]
\texttt{kernel\_\allowbreak{}no\_\allowbreak{}shell\_\allowbreak{}commands} & \textbf{[E]} & \texttt{kernel} & Kernel must not use os.system, subprocess.call, or shell=True. \\
\addlinespace[1pt]
\texttt{kernel\_\allowbreak{}no\_\allowbreak{}file\_\allowbreak{}writes} & \textbf{[E]} & \texttt{kernel} & Kernel must not write or delete files. \\
\addlinespace[1pt]
\texttt{kernel\_\allowbreak{}no\_\allowbreak{}event\_\allowbreak{}remapping} & \textbf{[W]} & \texttt{kernel} & Kernel must not hard-remap original event IDs. Prefix enrichment (attr-event\_id) is allowed; replacing/renaming event IDs is not (HC-5). \\
\addlinespace[1pt]
\bottomrule
\end{tabular}
\endgroup
\end{tcolorbox}

%% file: appendix_document_cases.tex
\section{\emph{Document} Cases}
\label{appendix:document_cases}

{\setlength{\parskip}{2pt}\setlength{\parindent}{0pt}
\noindent\begin{tcolorbox}[breakable,enhanced,colback=white,colframe=gray!55,boxrule=0.9pt,arc=4pt,left=6pt,right=6pt,top=4pt,bottom=4pt,title={\small\bfseries BCI Competition IV Dataset 2a (4-Class Motor Imagery)\hfill \colorbox{gray!20}{\scriptsize\sffamily\bfseries Completed}},coltitle=black,colbacktitle=gray!12,titlerule=0.5pt]
{\footnotesize \textbf{ID:}~\texttt{bcic\_iv\_2a} $\mid$ \textbf{Year:}~2008 $\mid$ \textbf{Category:}~Action/Output $\mid$ \textbf{Subjects:}~9 $\mid$ \textbf{Channels:}~25 $\mid$ \textbf{SR:}~250 Hz $\mid$ \textbf{Files (completed):}~18 $\mid$ \textbf{Super Meta:}~No $\mid$ \textbf{Reviewer:}~Anonymous}
\par{\footnotesize\textbf{URL:}~\url{https://www.bbci.de/competition/iv/desc_2a.pdf}}
\tcbline
\noindent{\bfseries\footnotesize Paradigm}\par\smallskip
{\footnotesize Nine subjects (A01-A09) performed four-class cued motor imagery (BCI Competition IV, 2008): left hand, right hand, feet, and tongue. Each session contained 6 runs of 48 trials each (288 trials per session), balanced across 4 classes (72 trials per class). A fixation cross preceded each trial, followed by a visual arrow cue indicating the imagery class, and a 4-second imagery period. Short resting intervals separated trials. Each subject was recorded in two sessions: a training session (suffix \texttt{T.gdf}) with event labels, and an evaluation session (suffix \texttt{E.gdf}) with initially unknown labels replaced by kernel-loaded \texttt{.mat} companion files. EEG was recorded from 22 channels (standard 10-20, sensorimotor focus) plus 3 bipolar EOG channels (filtered 0.5-100 Hz, notch at 50 Hz).}
\noindent\rule{\linewidth}{0.4pt}\par\nopagebreak
\noindent{\bfseries\footnotesize File Scan Result}\par\smallskip
{\scriptsize
\noindent\textbf{Kernel SHA256:}~\texttt{489182843c6974d9f77d2a7ccd1f1bad91a5f3eb1d01dfc7e15b3321e4eb8803}\par\smallskip
\noindent\textbf{EEGUnity Version:}~\texttt{0.8.1}\par\smallskip
\noindent\textbf{Scan Type:}~Full\par\smallskip
}
\smallskip\noindent{\footnotesize\bfseries Event Structure}\par\smallskip
\begingroup\scriptsize\setlength{\tabcolsep}{2pt}
\begin{tabular}{@{}p{0.1567\linewidth}p{0.1567\linewidth}p{0.0783\linewidth}p{0.0783\linewidth}p{0.4700\linewidth}@{}}
\toprule
\textbf{Event ID} & \textbf{Count (total)} & \textbf{Type} & \textbf{Meta Level} & \textbf{Meaning} \\
\midrule
\texttt{Cue onset foot (class 3)} & 1375 & event & raw & Feet motor imagery cue \\
\texttt{Cue onset left (class 1)} & 1224 & event & raw & Left-hand motor imagery cue \\
\texttt{Cue onset right (class 2)} & 1224 & event & raw & Right-hand motor imagery cue \\
\texttt{Cue onset tongue (class 4)} & 1656 & event & raw & Tongue motor imagery cue \\
\texttt{Eye movements} & 18 & admin & raw & Artifactual eye movement event \\
\texttt{Idling EEG (eyes closed)} & 17 & admin & raw & Resting state with eyes closed \\
\texttt{Idling EEG (eyes open)} & 17 & admin & raw & Resting state with eyes open \\
\texttt{Rejected trial} & 488 & admin & raw & Trial rejected by experimenter \\
\texttt{Start of a new run} & 153 & admin & raw & Run boundary marker \\
\texttt{Start of a trial} & 4896 & admin & raw & Trial-onset fixation cross (pre-cue marker) \\
\bottomrule
\end{tabular}
\endgroup
\smallskip\noindent{\footnotesize\bfseries Info}\par\smallskip
\begingroup\scriptsize\setlength{\tabcolsep}{2pt}
\begin{tabular}{@{}p{0.2350\linewidth}p{0.2350\linewidth}p{0.4700\linewidth}@{}}
\toprule
\textbf{Field} & \textbf{Completeness} & \textbf{Notes} \\
\midrule
\texttt{age} & complete & N/A \\
\texttt{sex} & complete & N/A \\
\texttt{device} & missing & N/A \\
\bottomrule
\end{tabular}
\endgroup
\noindent\rule{\linewidth}{0.7pt}\par\nopagebreak
\noindent{\bfseries\footnotesize Tasks}
\noindent{\footnotesize\bfseries Classical~Task~1:} {\footnotesize Four-Class Motor Imagery Classification}\par\smallskip
{\footnotesize\begin{tabular}{@{}lp{0.62\linewidth}@{}}
\textbf{Type} & classification \\
\textbf{Epoch Method} & \texttt{epoch\_by\_event\_hdf5} \\
\textbf{Epoch Params} & tmin = 0.5 s, tmax = 2.5 s, target\_events = [\texttt{Cue onset left (class 1)}, \texttt{Cue onset right (class 2)}, \texttt{Cue onset foot (class 3)}, \texttt{Cue onset tongue (class 4)}] \\
\textbf{Label Route} & \texttt{event} \\
\textbf{Meta Level} & raw \\
\end{tabular}}
\par\noindent{\footnotesize\bfseries Target Definition:}\par\smallskip
\begingroup\footnotesize
\begin{tabular}{@{}p{0.3911\linewidth}p{0.0978\linewidth}p{0.3911\linewidth}@{}}
\toprule
\textbf{Event ID} & \textbf{y} & \textbf{Meaning} \\
\midrule
\texttt{Cue onset left (class 1)} & 0 & Left-hand motor imagery \\
\texttt{Cue onset right (class 2)} & 1 & Right-hand motor imagery \\
\texttt{Cue onset foot (class 3)} & 2 & Feet motor imagery \\
\texttt{Cue onset tongue (class 4)} & 3 & Tongue motor imagery \\
\bottomrule
\end{tabular}
\endgroup
\par\noindent{\footnotesize\itshape Evidence:}\par
\begin{itemize}[leftmargin=1.3em,noitemsep,topsep=1pt]
\item[\scriptsize$\cdot$] {\scriptsize DOI: official website --- Official BCI Competition IV Dataset 2a description defines four-class MI (left hand, right hand, feet, tongue) as benchmark task with 4s imagery window from cue onset.}
\item[\scriptsize$\cdot$] {\scriptsize DOI: 10.1109/tnnls.2018.2789927 --- CNN trained on 0.5-2.5s MI period after cue onset.}
\item[\scriptsize$\cdot$] {\scriptsize DOI: 10.1109/tnsre.2022.3156076 --- SincNet hybrid achieves 74.26\% accuracy.}
\end{itemize}
\noindent\rule{\linewidth}{0.4pt}\par\nopagebreak
\noindent{\footnotesize\bfseries Potential~Task~1:} {\footnotesize Binary Motor Imagery - Left vs. Right Hand}\par\smallskip
{\footnotesize\begin{tabular}{@{}lp{0.62\linewidth}@{}}
\textbf{Type} & classification \\
\textbf{Epoch Method} & \texttt{epoch\_by\_event\_hdf5} \\
\textbf{Epoch Params} & tmin = 0.5 s, tmax = 2.5 s, target\_events = [\texttt{Cue onset left (class 1)}, \texttt{Cue onset right (class 2)}] \\
\textbf{Label Route} & \texttt{event} \\
\textbf{Meta Level} & raw \\
\end{tabular}}
\par\noindent{\footnotesize\bfseries Target Definition:}\par\smallskip
\begingroup\footnotesize
\begin{tabular}{@{}p{0.3911\linewidth}p{0.0978\linewidth}p{0.3911\linewidth}@{}}
\toprule
\textbf{Event ID} & \textbf{y} & \textbf{Meaning} \\
\midrule
\texttt{Cue onset left (class 1)} & 0 & Left-hand motor imagery \\
\texttt{Cue onset right (class 2)} & 1 & Right-hand motor imagery \\
\bottomrule
\end{tabular}
\endgroup
\par\smallskip
\noindent\rule{\linewidth}{0.4pt}\par\nopagebreak
\noindent{\bfseries\footnotesize Data Quality Notes}\par\smallskip
\begin{itemize}[leftmargin=1.4em,noitemsep,topsep=2pt]
\item{\footnotesize Rejected trials (variable per subject/session) should be excluded from analysis.}
\item{\footnotesize EOG channels present; apply artifact rejection or regression before motor imagery classification.}
\item{\footnotesize Evaluation files initially contain \texttt{Cue unknown (class 783)} entries that the kernel replaces with proper class labels from \texttt{.mat} companion files; verify companion file presence.}
\item{\footnotesize Class balance is enforced by design (72 trials per class per session); minor imbalance may arise from rejected trials.}
\item{\footnotesize Subject age range: 17-26 years; 7 female, 2 male subjects.}
\end{itemize}
\end{tcolorbox}
\vspace{3pt}
\noindent\begin{tcolorbox}[breakable,enhanced,colback=white,colframe=gray!55,boxrule=0.9pt,arc=4pt,left=6pt,right=6pt,top=4pt,bottom=4pt,title={\small\bfseries EEG Motor Movement/Imagery Database\hfill \colorbox{gray!20}{\scriptsize\sffamily\bfseries Completed}},coltitle=black,colbacktitle=gray!12,titlerule=0.5pt]
{\footnotesize \textbf{ID:}~\texttt{physionet\_eegmmidb} $\mid$ \textbf{Year:}~2009 $\mid$ \textbf{Category:}~Action/Output $\mid$ \textbf{Subjects:}~109 $\mid$ \textbf{Channels:}~64 $\mid$ \textbf{SR:}~160 Hz $\mid$ \textbf{Files (completed):}~1526 $\mid$ \textbf{Super Meta:}~No $\mid$ \textbf{Reviewer:}~Anonymous}
\par{\footnotesize\textbf{URL:}~\url{https://physionet.org/content/eegmmidb/1.0.0/}}
\tcbline
\noindent{\bfseries\footnotesize Paradigm}\par\smallskip
{\footnotesize One hundred and nine volunteers performed a series of motor movement and motor imagery tasks, recorded with 64 EEG electrodes (international 10-10 system) at 160 Hz using the BCI2000 system (Schalk et al., 2004). Each subject completed 14 runs (R01-R14): two baseline runs (eyes open and eyes closed), and twelve task runs organized in three repeating cycles of four run types. Task runs alternate between real movements and mental imagery of the same movements. Two movement classes are used: left/right fist, and both fists/both feet. Events T0 (rest), T1, and T2 are annotated within each run; T1/T2 meanings differ by run number.}
\noindent\rule{\linewidth}{0.4pt}\par\nopagebreak
\noindent{\bfseries\footnotesize File Scan Result}\par\smallskip
{\scriptsize
\noindent\textbf{Kernel SHA256:}~\texttt{bd1ed7ea5b310949f1bb2ce29057964e4e25db94ffb6a393129b948b562b88b8}\par\smallskip
\noindent\textbf{EEGUnity Version:}~\texttt{0.8.1}\par\smallskip
\noindent\textbf{Scan Type:}~Full\par\smallskip
}
\smallskip\noindent{\footnotesize\bfseries Event Structure}\par\smallskip
\begingroup\scriptsize\setlength{\tabcolsep}{2pt}
\begin{tabular}{@{}p{0.1567\linewidth}p{0.1567\linewidth}p{0.0783\linewidth}p{0.0783\linewidth}p{0.4700\linewidth}@{}}
\toprule
\textbf{Event ID} & \textbf{Count (total)} & \textbf{Type} & \textbf{Meta Level} & \textbf{Meaning} \\
\midrule
\texttt{Baseline, eyes closed} & 109 & event & raw & Eyes closed resting state baseline recording \\
\texttt{Baseline, eyes open} & 109 & event & raw & Eyes open resting state baseline recording \\
\texttt{Imagine Opening Both Fists Movement} & 2465 & event & raw & Motor imagery cue: open both fists \\
\texttt{Imagine Opening Feet Movement} & 2455 & event & raw & Motor imagery cue: open/move both feet \\
\texttt{Imagine Opening Left Fist Movement} & 2480 & event & raw & Motor imagery cue: open left fist \\
\texttt{Imagine Opening Right Fist Movement} & 2438 & event & raw & Motor imagery cue: open right fist \\
\texttt{Open Both Feet Movement} & 2469 & event & raw & Real movement cue: open both feet \\
\texttt{Open Both Fists Movement} & 2440 & event & raw & Real movement cue: open both fists \\
\texttt{Open Left Fist Movement} & 2471 & event & raw & Real movement cue: open left fist \\
\texttt{Open Right Fist Movement} & 2456 & event & raw & Real movement cue: open right fist \\
\texttt{Rest} & 19675 & event & raw & Rest period / trial baseline marker \\
\texttt{T1} & 2 & admin & raw & Task cue type 1 (meaning varies by run) \\
\bottomrule
\end{tabular}
\endgroup
\smallskip\noindent{\footnotesize\bfseries Info}\par\smallskip
\begingroup\scriptsize\setlength{\tabcolsep}{2pt}
\begin{tabular}{@{}p{0.2350\linewidth}p{0.2350\linewidth}p{0.4700\linewidth}@{}}
\toprule
\textbf{Field} & \textbf{Completeness} & \textbf{Notes} \\
\midrule
\texttt{age} & missing & N/A \\
\texttt{sex} & missing & N/A \\
\texttt{device} & missing & N/A \\
\bottomrule
\end{tabular}
\endgroup
\noindent\rule{\linewidth}{0.7pt}\par\nopagebreak
\noindent{\bfseries\footnotesize Tasks}
\noindent{\footnotesize\bfseries Classical~Task~1:} {\footnotesize 4-Class Motor Imagery Classification}\par\smallskip
{\footnotesize\begin{tabular}{@{}lp{0.62\linewidth}@{}}
\textbf{Type} & classification \\
\textbf{Epoch Method} & \texttt{epoch\_by\_event\_hdf5} \\
\textbf{Epoch Params} & tmin = 0 s, tmax = 3 s, target\_events = [\texttt{Imagine Opening Left Fist Movement}, \texttt{Imagine Opening Right Fist Movement}, \texttt{Imagine Opening Both Fists Movement}, \texttt{Imagine Opening Feet Movement}] \\
\textbf{Label Route} & \texttt{event} \\
\textbf{Meta Level} & raw \\
\end{tabular}}
\par\noindent{\footnotesize\bfseries Target Definition:}\par\smallskip
\begingroup\footnotesize
\begin{tabular}{@{}p{0.3911\linewidth}p{0.0978\linewidth}p{0.3911\linewidth}@{}}
\toprule
\textbf{Event ID} & \textbf{y} & \textbf{Meaning} \\
\midrule
\texttt{Imagine Opening Left Fist Movement} & 0 & Left-fist motor imagery \\
\texttt{Imagine Opening Right Fist Movement} & 1 & Right-fist motor imagery \\
\texttt{Imagine Opening Both Fists Movement} & 2 & Both-fists motor imagery \\
\texttt{Imagine Opening Feet Movement} & 3 & Both-feet motor imagery \\
\bottomrule
\end{tabular}
\endgroup
\par\noindent{\footnotesize\itshape Evidence:}\par
\begin{itemize}[leftmargin=1.3em,noitemsep,topsep=1pt]
\item[\scriptsize$\cdot$] {\scriptsize DOI: 10.1109/TNSRE.2022.3186442 --- 4-class motor imagery classification (left/right fist, both fists, both feet) using the PhysioNet EEG Motor Imagery Database with tmin=0.0s and tmax=3.0s}
\end{itemize}
\noindent\rule{\linewidth}{0.4pt}\par\nopagebreak
\noindent{\footnotesize\bfseries Potential~Task~1:} {\footnotesize 4-Class Motor Movement Classification}\par\smallskip
{\footnotesize\begin{tabular}{@{}lp{0.62\linewidth}@{}}
\textbf{Type} & classification \\
\textbf{Epoch Method} & \texttt{epoch\_by\_event\_hdf5} \\
\textbf{Epoch Params} & tmin = 0 s, tmax = 3 s, target\_events = [\texttt{Open Left Fist Movement}, \texttt{Open Right Fist Movement}, \texttt{Open Both Fists Movement}, \texttt{Open Both Feet Movement}] \\
\textbf{Label Route} & \texttt{event} \\
\textbf{Meta Level} & raw \\
\end{tabular}}
\par\noindent{\footnotesize\bfseries Target Definition:}\par\smallskip
\begingroup\footnotesize
\begin{tabular}{@{}p{0.3911\linewidth}p{0.0978\linewidth}p{0.3911\linewidth}@{}}
\toprule
\textbf{Event ID} & \textbf{y} & \textbf{Meaning} \\
\midrule
\texttt{Open Left Fist Movement} & 0 & Left-fist movement \\
\texttt{Open Right Fist Movement} & 1 & Right-fist movement \\
\texttt{Open Both Fists Movement} & 2 & Both-fists movement \\
\texttt{Open Both Feet Movement} & 3 & Both-feet movement \\
\bottomrule
\end{tabular}
\endgroup
\noindent\rule{\linewidth}{0.4pt}\par\nopagebreak
\noindent{\footnotesize\bfseries Potential~Task~2:} {\footnotesize 4-Class Motor Movement Classification}\par\smallskip
{\footnotesize\begin{tabular}{@{}lp{0.62\linewidth}@{}}
\textbf{Type} & classification \\
\textbf{Epoch Method} & \texttt{epoch\_by\_event\_hdf5} \\
\textbf{Epoch Params} & tmin = 0.0 s, tmax = 3 s, target\_events = [\texttt{Imagine Opening Left Fist Movement}, \texttt{Open Left Fist Movement}, \texttt{Imagine Opening Right Fist Movement}, \texttt{Open Right Fist Movement}, \texttt{Imagine Opening Both Fists Movement}, \texttt{Open Both Fists Movement}, \texttt{Imagine Opening Feet Movement}, \texttt{Open Both Feet Movement}] \\
\textbf{Label Route} & \texttt{event} \\
\textbf{Meta Level} & raw \\
\end{tabular}}
\par\noindent{\footnotesize\bfseries Target Definition:}\par\smallskip
\begingroup\footnotesize
\begin{tabular}{@{}p{0.3911\linewidth}p{0.0978\linewidth}p{0.3911\linewidth}@{}}
\toprule
\textbf{Event ID} & \textbf{y} & \textbf{Meaning} \\
\midrule
\texttt{Imagine Opening Left Fist Movement} & 0 & Left-fist motor imagery (R04/08/12) \\
\texttt{Open Left Fist Movement} & 0 & Left-fist real movement (R03/07/11) \\
\texttt{Imagine Opening Right Fist Movement} & 1 & Right-fist motor imagery (R04/08/12) \\
\texttt{Open Right Fist Movement} & 1 & Right-fist real movement (R03/07/11) \\
\texttt{Imagine Opening Both Fists Movement} & 2 & Both-fists motor imagery (R06/10/14) \\
\texttt{Open Both Fists Movement} & 2 & Both-fists real movement (R05/09/13) \\
\texttt{Imagine Opening Feet Movement} & 3 & Both-feet motor imagery (R06/10/14) \\
\texttt{Open Both Feet Movement} & 3 & Both-feet real movement (R05/09/13) \\
\bottomrule
\end{tabular}
\endgroup
\noindent\rule{\linewidth}{0.4pt}\par\nopagebreak
\noindent{\footnotesize\bfseries Potential~Task~3:} {\footnotesize Binary Left vs. Right Fist Motor Imagery}\par\smallskip
{\footnotesize\begin{tabular}{@{}lp{0.62\linewidth}@{}}
\textbf{Type} & classification \\
\textbf{Epoch Method} & \texttt{epoch\_by\_event\_hdf5} \\
\textbf{Epoch Params} & tmin = 0 s, tmax = 3 s, target\_events = [\texttt{Imagine Opening Left Fist Movement}, \texttt{Imagine Opening Right Fist Movement}] \\
\textbf{Label Route} & \texttt{event} \\
\textbf{Meta Level} & raw \\
\end{tabular}}
\par\noindent{\footnotesize\bfseries Target Definition:}\par\smallskip
\begingroup\footnotesize
\begin{tabular}{@{}p{0.3911\linewidth}p{0.0978\linewidth}p{0.3911\linewidth}@{}}
\toprule
\textbf{Event ID} & \textbf{y} & \textbf{Meaning} \\
\midrule
\texttt{Imagine Opening Left Fist Movement} & 0 & Left-fist motor imagery \\
\texttt{Imagine Opening Right Fist Movement} & 1 & Right-fist motor imagery \\
\bottomrule
\end{tabular}
\endgroup
\noindent\rule{\linewidth}{0.4pt}\par\nopagebreak
\noindent{\footnotesize\bfseries Potential~Task~4:} {\footnotesize Both Fists vs. Both Feet Motor Imagery}\par\smallskip
{\footnotesize\begin{tabular}{@{}lp{0.62\linewidth}@{}}
\textbf{Type} & classification \\
\textbf{Epoch Method} & \texttt{epoch\_by\_event\_hdf5} \\
\textbf{Epoch Params} & tmin = 0 s, tmax = 3 s, target\_events = [\texttt{Imagine Opening Both Fists Movement}, \texttt{Imagine Opening Feet Movement}] \\
\textbf{Label Route} & \texttt{event} \\
\textbf{Meta Level} & raw \\
\end{tabular}}
\par\noindent{\footnotesize\bfseries Target Definition:}\par\smallskip
\begingroup\footnotesize
\begin{tabular}{@{}p{0.3911\linewidth}p{0.0978\linewidth}p{0.3911\linewidth}@{}}
\toprule
\textbf{Event ID} & \textbf{y} & \textbf{Meaning} \\
\midrule
\texttt{Imagine Opening Both Fists Movement} & 0 & Both-fists motor imagery (R06/10/14) \\
\texttt{Imagine Opening Feet Movement} & 1 & Both-feet motor imagery (R06/10/14) \\
\bottomrule
\end{tabular}
\endgroup
\par\smallskip
\noindent\rule{\linewidth}{0.4pt}\par\nopagebreak
\noindent{\bfseries\footnotesize Data Quality Notes}\par\smallskip
\begin{itemize}[leftmargin=1.4em,noitemsep,topsep=2pt]
\item{\footnotesize Age and sex information are not available in the public metadata.}
\item{\footnotesize Known data quality issues: several subjects have noisy recordings; EEG.physionet.org community notes suggest subjects S088, S089, S092, S100, S104 may have unusual recordings.}
\item{\footnotesize T1/T2 event meaning \textbf{changes by run number}; always filter by run before assigning class labels.}
\item{\footnotesize \texttt{Baseline} (T0) events in task runs are rest periods and should be excluded from MI classification targets.}
\end{itemize}
\end{tcolorbox}
\vspace{3pt}
\noindent\begin{tcolorbox}[breakable,enhanced,colback=white,colframe=gray!55,boxrule=0.9pt,arc=4pt,left=6pt,right=6pt,top=4pt,bottom=4pt,title={\small\bfseries Sleep-EDF Database Expanded\hfill \colorbox{gray!20}{\scriptsize\sffamily\bfseries Completed}},coltitle=black,colbacktitle=gray!12,titlerule=0.5pt]
{\footnotesize \textbf{ID:}~\texttt{physionet\_sleepedfx} $\mid$ \textbf{Year:}~1987-1991/1994 $\mid$ \textbf{Category:}~Biological State $\mid$ \textbf{Subjects:}~106 $\mid$ \textbf{Channels:}~7 $\mid$ \textbf{SR:}~100 Hz $\mid$ \textbf{Files (completed):}~197 $\mid$ \textbf{Super Meta:}~No $\mid$ \textbf{Reviewer:}~Anonymous}
\par{\footnotesize\textbf{URL:}~\url{https://physionet.org/content/sleep-edfx/1.0.0/}}
\tcbline
\noindent{\bfseries\footnotesize Paradigm}\par\smallskip
{\footnotesize The sleep-edf database contains 197 whole-night PolySomnoGraphic sleep recordings, containing EEG, EOG, chin EMG, and event markers. Some records also contain respiration and body temperature. Corresponding hypnograms (sleep patterns) were manually scored by well-trained technicians according to the Rechtschaffen and Kales manual, and are also available.}
\noindent\rule{\linewidth}{0.4pt}\par\nopagebreak
\noindent{\bfseries\footnotesize File Scan Result}\par\smallskip
{\scriptsize
\noindent\textbf{Kernel SHA256:}~\texttt{db96b3badc0af51588705980412f7ed0016ef5345e218e37bd3d1fd192911aad}\par\smallskip
\noindent\textbf{EEGUnity Version:}~\texttt{0.8.1}\par\smallskip
\noindent\textbf{Scan Type:}~Full\par\smallskip
}
\smallskip\noindent{\footnotesize\bfseries Event Structure}\par\smallskip
\begingroup\scriptsize\setlength{\tabcolsep}{2pt}
\begin{tabular}{@{}p{0.1567\linewidth}p{0.1567\linewidth}p{0.0783\linewidth}p{0.0783\linewidth}p{0.4700\linewidth}@{}}
\toprule
\textbf{Event ID} & \textbf{Count (total)} & \textbf{Type} & \textbf{Meta Level} & \textbf{Meaning} \\
\midrule
\texttt{Movement time} & 196 & event & raw & Subject movement marker; out-of-bed period during recording \\
\texttt{Sleep stage 1} & 7292 & event & raw & Sleep stage 1 (light sleep) epoch annotation \\
\texttt{Sleep stage 2} & 8171 & event & raw & Sleep stage 2 (light sleep) epoch annotation \\
\texttt{Sleep stage 3} & 4820 & event & raw & Sleep stage 3 (slow wave sleep) epoch annotation \\
\texttt{Sleep stage 4} & 1462 & event & raw & Sleep stage 4 (deep slow wave sleep) epoch annotation \\
\texttt{Sleep stage ?} & 173 & event & raw & Unscored or ambiguous epoch; annotation uncertain \\
\texttt{Sleep stage R} & 1953 & event & raw & Sleep stage R (REM sleep) epoch annotation \\
\texttt{Sleep stage W} & 4461 & event & raw & Sleep stage W (wakefulness) epoch annotation \\
\bottomrule
\end{tabular}
\endgroup
\smallskip\noindent{\footnotesize\bfseries Info}\par\smallskip
\begingroup\scriptsize\setlength{\tabcolsep}{2pt}
\begin{tabular}{@{}p{0.2350\linewidth}p{0.2350\linewidth}p{0.4700\linewidth}@{}}
\toprule
\textbf{Field} & \textbf{Completeness} & \textbf{Notes} \\
\midrule
\texttt{age} & complete & N/A \\
\texttt{sex} & complete & N/A \\
\texttt{device} & missing & N/A \\
\bottomrule
\end{tabular}
\endgroup
\noindent\rule{\linewidth}{0.7pt}\par\nopagebreak
\noindent{\bfseries\footnotesize Tasks}
\noindent{\footnotesize\bfseries Classical~Task~1:} {\footnotesize Sleep stage classification (5-class)}\par\smallskip
{\footnotesize\begin{tabular}{@{}lp{0.62\linewidth}@{}}
\textbf{Type} & classification \\
\textbf{Epoch Method} & \texttt{epoch\_by\_long\_event\_hdf5} \\
\textbf{Epoch Params} & tmin = 0.0 s, tmax = 30.0 s, overlap = 0.0, target\_events = [\texttt{Sleep stage W}, \texttt{Sleep stage 1}, \texttt{Sleep stage 2}, \texttt{Sleep stage 3}, \texttt{Sleep stage 4}, \texttt{Sleep stage R}] \\
\textbf{Label Route} & \texttt{event} \\
\textbf{Meta Level} & raw \\
\end{tabular}}
\par\noindent{\footnotesize\bfseries Target Definition:}\par\smallskip
\begingroup\footnotesize
\begin{tabular}{@{}p{0.3911\linewidth}p{0.0978\linewidth}p{0.3911\linewidth}@{}}
\toprule
\textbf{Event ID} & \textbf{y} & \textbf{Meaning} \\
\midrule
\texttt{Sleep stage W} & 0 & Wake \\
\texttt{Sleep stage 1} & 1 & NREM Stage 1 \\
\texttt{Sleep stage 2} & 2 & NREM Stage 2 \\
\texttt{Sleep stage 3} & 3 & NREM Stage 3 (includes S3+S4 merged per AASM) \\
\texttt{Sleep stage 4} & 3 & NREM Stage 3 (mapped to y=3; S3 and S4 merged) \\
\texttt{Sleep stage R} & 4 & REM \\
\bottomrule
\end{tabular}
\endgroup
\par\noindent{\footnotesize\itshape Evidence:}\par
\begin{itemize}[leftmargin=1.3em,noitemsep,topsep=1pt]
\item[\scriptsize$\cdot$] {\scriptsize 10.7717/peerjcs.1561\_table --- 2 --- Five-class sleep staging (W, N1, N2, N3, REM) on Sleep-EDF Expanded, 30-second epochs, AASM standard merging S3+S4 into N3}
\end{itemize}
\noindent\rule{\linewidth}{0.4pt}\par\nopagebreak
\noindent{\footnotesize\bfseries Potential~Task~1:} {\footnotesize Wake vs. Sleep Binary Classification (epoch\_by\_long\_event)}\par\smallskip
{\footnotesize\begin{tabular}{@{}lp{0.62\linewidth}@{}}
\textbf{Type} & classification \\
\textbf{Epoch Method} & \texttt{epoch\_by\_long\_event\_hdf5} \\
\textbf{Epoch Params} & tmin = 0.0 s, tmax = 30.0 s, overlap = 0.0, target\_events = [\texttt{Sleep stage W}, \texttt{Sleep stage 1}, \texttt{Sleep stage 2}, \texttt{Sleep stage 3}, \texttt{Sleep stage 4}, \texttt{Sleep stage R}] \\
\textbf{Label Route} & \texttt{event} \\
\textbf{Meta Level} & raw \\
\end{tabular}}
\par\noindent{\footnotesize\bfseries Target Definition:}\par\smallskip
\begingroup\footnotesize
\begin{tabular}{@{}p{0.3911\linewidth}p{0.0978\linewidth}p{0.3911\linewidth}@{}}
\toprule
\textbf{Event ID} & \textbf{y} & \textbf{Meaning} \\
\midrule
\texttt{Sleep stage W} & 0 & Wake \\
\texttt{Sleep stage 1} & 1 & Any sleep (stages 1-4 and R) \\
\texttt{Sleep stage 2} & 1 & Any sleep \\
\texttt{Sleep stage 3} & 1 & Any sleep \\
\texttt{Sleep stage 4} & 1 & Any sleep \\
\texttt{Sleep stage R} & 1 & Any sleep \\
\bottomrule
\end{tabular}
\endgroup
\par\noindent{\footnotesize\itshape Evidence:}\par
\begin{itemize}[leftmargin=1.3em,noitemsep,topsep=1pt]
\item[\scriptsize$\cdot$] {\scriptsize \textbf{Notes}:}
\end{itemize}
\par\noindent{\footnotesize\bfseries Notes:} {\footnotesize \textbf{Kernel Notes}:}
\par\smallskip
\noindent\rule{\linewidth}{0.4pt}\par\nopagebreak
\noindent{\bfseries\footnotesize Data Quality Notes}\par\smallskip
\begin{itemize}[leftmargin=1.4em,noitemsep,topsep=2pt]
\item{\footnotesize Only PSG files should be processed; Hypnogram files (matched companion files) are used for annotation only and are returned unchanged by the kernel.}
\item{\footnotesize Class imbalance: Stage 2 is the most frequent stage; Wake also contributes large counts from lights-on periods at start/end of recording.}
\item{\footnotesize Two EEG channels only (Fpz-Cz, Pz-Oz); limited spatial resolution.}
\end{itemize}
\end{tcolorbox}
\vspace{3pt}
\noindent\begin{tcolorbox}[breakable,enhanced,colback=white,colframe=gray!55,boxrule=0.9pt,arc=4pt,left=6pt,right=6pt,top=4pt,bottom=4pt,title={\small\bfseries TUH EEG Seizure Corpus\hfill \colorbox{gray!20}{\scriptsize\sffamily\bfseries Completed}},coltitle=black,colbacktitle=gray!12,titlerule=0.5pt]
{\footnotesize \textbf{ID:}~\texttt{tuh\_eeg\_seizure} $\mid$ \textbf{Year:}~2018 $\mid$ \textbf{Category:}~Biological State $\mid$ \textbf{Subjects:}~675 $\mid$ \textbf{Channels:}~17-128 $\mid$ \textbf{SR:}~250/256/400/1000 Hz $\mid$ \textbf{Files (completed):}~7361 $\mid$ \textbf{Super Meta:}~No $\mid$ \textbf{Reviewer:}~Anonymous}
\par{\footnotesize\textbf{URL:}~\url{https://isip.piconepress.com/projects/nedc/html/tuh_eeg}}
\tcbline
\noindent{\bfseries\footnotesize Paradigm}\par\smallskip
{\footnotesize The TUH EEG Seizure Corpus (TUSZ) is a specialized dataset developed to advance seizure detection algorithms through machine learning. A subset of the broader TUH EEG Corpus, this dataset includes sessions specifically featuring seizure events, alongside sessions without seizures for testing false alarm rates. It consists of data from three primary directories: training (train), development (dev), and evaluation (eval), with 7361 files in total, across various seizure and non-seizure events. The dataset features EEG data in European Data Format (edf), with event-based annotations captured in both multi-class and bi-class formats. Multi-class annotations provide detailed seizure types, while bi-class annotations simplify the dataset into seizure (seiz) and background (bckg) labels. These annotations are recorded per channel, capturing the spread of seizures across multiple electrodes. EEG recordings are captured using two common reference configurations: averaged reference (AR) and linked ears reference (LE), with the data adhering to the standard 10/20 channel montage for consistency across files. The dataset includes a vast number of seizure events, with specific counts for training, development, and evaluation sets. The TUSZ corpus plays a vital role in the development and testing of advanced seizure detection systems, supporting both the detection of seizures and the analysis of seizure-free segments}
\noindent\rule{\linewidth}{0.4pt}\par\nopagebreak
\noindent{\bfseries\footnotesize File Scan Result}\par\smallskip
{\scriptsize
\noindent\textbf{Kernel SHA256:}~\texttt{0c90e3ec94547026731787ac95cb32b924b667dc778975c503f0d947daff981f}\par\smallskip
\noindent\textbf{EEGUnity Version:}~\texttt{0.8.1}\par\smallskip
\noindent\textbf{Scan Type:}~Full\par\smallskip
\noindent the meaning from dataset file "seizures\_types\_v02.xlsx"\par\smallskip
}
\smallskip\noindent{\footnotesize\bfseries Event Structure}\par\smallskip
\begingroup\scriptsize\setlength{\tabcolsep}{2pt}
\begin{tabular}{@{}p{0.1567\linewidth}p{0.1567\linewidth}p{0.0783\linewidth}p{0.0783\linewidth}p{0.4700\linewidth}@{}}
\toprule
\textbf{Event ID} & \textbf{Count (total)} & \textbf{Type} & \textbf{Meta Level} & \textbf{Meaning} \\
\midrule
\texttt{absz} & 121 & event & raw & Absence seizure \\
\texttt{bckg} & 15454 & event & raw & Background activity \\
\texttt{cpsz} & 1520 & event & raw & Complex partial seizure \\
\texttt{fnsz} & 4891 & event & raw & Focal non-specific seizure \\
\texttt{gnsz} & 2491 & event & raw & Generalized non-specific seizure \\
\texttt{mysz} & 3 & event & raw & Myoclonic seizure \\
\texttt{seiz} & 146 & event & raw & Seizure event (Common seizure class which can include all types of seizure) \\
\texttt{spsz} & 57 & event & raw & Simple partial seizure \\
\texttt{tcsz} & 199 & event & raw & Tonic-clonic seizure (At first stiffening and then jerking of body (Grand Mal)) \\
\texttt{tnsz} & 126 & event & raw & Tonic seizure (Stiffening of body during seizure, EEG effects disappears) \\
\bottomrule
\end{tabular}
\endgroup
\smallskip\noindent{\footnotesize\bfseries Info}\par\smallskip
\begingroup\scriptsize\setlength{\tabcolsep}{2pt}
\begin{tabular}{@{}p{0.2350\linewidth}p{0.2350\linewidth}p{0.4700\linewidth}@{}}
\toprule
\textbf{Field} & \textbf{Completeness} & \textbf{Notes} \\
\midrule
\texttt{age} & missing & N/A \\
\texttt{sex} & missing & N/A \\
\texttt{device} & missing & N/A \\
\bottomrule
\end{tabular}
\endgroup
\noindent\rule{\linewidth}{0.7pt}\par\nopagebreak
\noindent{\bfseries\footnotesize Tasks}
\noindent{\footnotesize\bfseries Classical~Task~1:} {\footnotesize Binary Seizure Detection (Seizure vs Background)}\par\smallskip
{\footnotesize\begin{tabular}{@{}lp{0.62\linewidth}@{}}
\textbf{Type} & classification \\
\textbf{Epoch Method} & \texttt{epoch\_by\_long\_event\_hdf5} \\
\textbf{Epoch Params} & tmin = 0.0 s, tmax = 1 s, overlap = 0.0, target\_events = [\texttt{absz}, \texttt{bckg}, \texttt{cpsz}, \texttt{fnsz}, \texttt{gnsz}, \texttt{mysz}, \texttt{seiz}, \texttt{spsz}, \texttt{tcsz}, \texttt{tnsz}] \\
\textbf{Label Route} & \texttt{event} \\
\textbf{Meta Level} & raw \\
\end{tabular}}
\par\noindent{\footnotesize\bfseries Target Definition:}\par\smallskip
\begingroup\footnotesize
\begin{tabular}{@{}p{0.3911\linewidth}p{0.0978\linewidth}p{0.3911\linewidth}@{}}
\toprule
\textbf{Event ID} & \textbf{y} & \textbf{Meaning} \\
\midrule
\texttt{bckg} & 0 & Background / non-seizure \\
\texttt{gnsz} & 1 & Generalized non-specific seizure \\
\texttt{fnsz} & 1 & Focal non-specific seizure \\
\texttt{cpsz} & 1 & Complex partial seizure \\
\texttt{absz} & 1 & Absence seizure \\
\texttt{tnsz} & 1 & Tonic seizure \\
\texttt{spsz} & 1 & Simple Partial Seizure \\
\texttt{mysz} & 1 & Myoclonic Seizure \\
\texttt{tcsz} & 1 & Tonic Clonic Seizure \\
\bottomrule
\end{tabular}
\endgroup
\par\noindent{\footnotesize\itshape Evidence:}\par
\begin{itemize}[leftmargin=1.3em,noitemsep,topsep=1pt]
\item[\scriptsize$\cdot$] {\scriptsize 10.1088/1741-2552/ac7d0d --- Uses TUH EEG with 1-second windows (50\% overlap) for binary seizure vs non-seizure classification; leave-one-out cross-subject evaluation}
\item[\scriptsize$\cdot$] {\scriptsize 10.1109/tii.2023.3274913 --- Uses TUH EEG with 4-second non-overlapping epochs for binary seizure detection; CNN classifier achieved 98.48\% average accuracy with leave-one-patient-out cross-validation}
\end{itemize}
\noindent\rule{\linewidth}{0.4pt}\par\nopagebreak
\noindent{\footnotesize\bfseries Potential~Task~1:} {\footnotesize Multi-class Seizure Type Classification}\par\smallskip
{\footnotesize\begin{tabular}{@{}lp{0.62\linewidth}@{}}
\textbf{Type} & classification \\
\textbf{Epoch Method} & \texttt{epoch\_by\_long\_event\_hdf5} \\
\textbf{Epoch Params} & tmin = 0.0 s, tmax = 1 s, overlap = 0.0, target\_events = [\texttt{absz}, \texttt{cpsz}, \texttt{fnsz}, \texttt{gnsz}, \texttt{mysz}, \texttt{spsz}, \texttt{tcsz}, \texttt{tnsz}] \\
\textbf{Label Route} & \texttt{event} \\
\textbf{Meta Level} & raw \\
\end{tabular}}
\par\noindent{\footnotesize\bfseries Target Definition:}\par\smallskip
\begingroup\footnotesize
\begin{tabular}{@{}p{0.3911\linewidth}p{0.0978\linewidth}p{0.3911\linewidth}@{}}
\toprule
\textbf{Event ID} & \textbf{y} & \textbf{Meaning} \\
\midrule
\texttt{gnsz} & 0 & Generalized non-specific seizure \\
\texttt{fnsz} & 1 & Focal non-specific seizure \\
\texttt{cpsz} & 2 & Complex partial seizure \\
\texttt{absz} & 3 & Absence seizure \\
\texttt{tnsz} & 4 & Tonic seizure \\
\texttt{spsz} & 5 & Simple Partial Seizure \\
\texttt{tcsz} & 6 & Tonic Clonic Seizure \\
\texttt{mysz} & 7 & Myoclonic Seizure \\
\bottomrule
\end{tabular}
\endgroup
\par\noindent{\footnotesize\bfseries Notes:} {\footnotesize classify further specific seizure, more challenge.}
\par\smallskip
\noindent\rule{\linewidth}{0.4pt}\par\nopagebreak
\noindent{\bfseries\footnotesize Data Quality Notes}\par\smallskip
\begin{itemize}[leftmargin=1.4em,noitemsep,topsep=2pt]
\item{\footnotesize \#\# Kernel Responsibilities}
\end{itemize}
\end{tcolorbox}
\vspace{3pt}
\noindent\begin{tcolorbox}[breakable,enhanced,colback=white,colframe=gray!55,boxrule=0.9pt,arc=4pt,left=6pt,right=6pt,top=4pt,bottom=4pt,title={\small\bfseries Healthy Brain Network (HBN) EEG - Release 1\hfill \colorbox{gray!20}{\scriptsize\sffamily\bfseries Completed}},coltitle=black,colbacktitle=gray!12,titlerule=0.5pt]
{\footnotesize \textbf{ID:}~\texttt{openneuro\_ds005505} $\mid$ \textbf{Year:}~2017/2025 $\mid$ \textbf{Category:}~Complex Applied $\mid$ \textbf{Subjects:}~136 $\mid$ \textbf{Channels:}~129 $\mid$ \textbf{SR:}~500 Hz $\mid$ \textbf{Files (completed):}~1342 $\mid$ \textbf{Super Meta:}~Yes $\mid$ \textbf{Reviewer:}~Anonymous}
\par{\footnotesize\textbf{URL:}~\url{https://openneuro.org/datasets/ds005505}}
\tcbline
\noindent{\bfseries\footnotesize Paradigm}\par\smallskip
{\footnotesize The Healthy Brain Network (HBN) EEG Release 1 is a large-scale developmental EEG battery collected from children and adolescents. This release includes six paradigms: naturalistic movie watching, resting state with alternating eyes-open and eyes-closed periods, contrast change detection, surround suppression, sequence learning, and symbol search. For the current benchmark draft, we keep only the tasks that have strong grounding in the release documentation and can be implemented directly from native events or narrowly defined kernel derivations.}
\noindent\rule{\linewidth}{0.4pt}\par\nopagebreak
\noindent{\bfseries\footnotesize File Scan Result}\par\smallskip
{\scriptsize
\noindent\textbf{Kernel SHA256:}~\texttt{d2a20370c75e70aa962603af5a1b27b44532d88150c49839b32b4d27ebabb160}\par\smallskip
\noindent\textbf{EEGUnity Version:}~\texttt{0.8.1}\par\smallskip
\noindent\textbf{Scan Type:}~Full\par\smallskip
}
\smallskip\noindent{\footnotesize\bfseries Event Structure}\par\smallskip
\begingroup\scriptsize\setlength{\tabcolsep}{2pt}
\begin{tabular}{@{}p{0.1567\linewidth}p{0.1567\linewidth}p{0.0783\linewidth}p{0.0783\linewidth}p{0.4700\linewidth}@{}}
\toprule
\textbf{Event ID} & \textbf{Count (total)} & \textbf{Type} & \textbf{Meta Level} & \textbf{Meaning} \\
\midrule
\texttt{9999} & 1136 & admin & raw & Administrative recording/session marker \\
\texttt{boundary} & 728 & admin & raw & EEGLAB/BIDS boundary or discontinuity marker, not a stimulus class \\
\texttt{break cnt} & 1488 & admin & raw & Administrative break counter marker, often at recording onset \\
\texttt{contrastChangeB1\_\allowbreak{}start} & 102 & event & raw & Start of contrast change detection block/run 1 \\
\texttt{contrastChangeB2\_\allowbreak{}start} & 117 & event & raw & Start of contrast change detection block/run 2 \\
\texttt{contrastChangeB3\_\allowbreak{}start} & 143 & event & raw & Start of contrast change detection block/run 3 \\
\texttt{contrastTrial\_\allowbreak{}start} & 8884 & event & raw & Start of a single contrast change detection trial \\
\texttt{contrast\_\allowbreak{}response\_\allowbreak{}correct} & 6588 & event & super & Button response was correct for the preceding contrast target \\
\texttt{contrast\_\allowbreak{}response\_\allowbreak{}incorrect} & 731 & event & super & Button response was incorrect for the preceding contrast target \\
\texttt{dot\_\allowbreak{}no1\_\allowbreak{}OFF} & 600 & event & raw & Dot position 1 offset in sequence learning \\
\texttt{dot\_\allowbreak{}no1\_\allowbreak{}ON} & 600 & event & raw & Dot position 1 onset in sequence learning \\
\texttt{dot\_\allowbreak{}no2\_\allowbreak{}OFF} & 1202 & event & raw & Dot position 2 offset in sequence learning \\
\texttt{dot\_\allowbreak{}no2\_\allowbreak{}ON} & 1234 & event & raw & Dot position 2 onset in sequence learning \\
\texttt{dot\_\allowbreak{}no3\_\allowbreak{}OFF} & 600 & event & raw & Dot position 3 offset in sequence learning \\
\texttt{dot\_\allowbreak{}no3\_\allowbreak{}ON} & 624 & event & raw & Dot position 3 onset in sequence learning \\
\texttt{dot\_\allowbreak{}no4\_\allowbreak{}OFF} & 600 & event & raw & Dot position 4 offset in sequence learning \\
\texttt{dot\_\allowbreak{}no4\_\allowbreak{}ON} & 600 & event & raw & Dot position 4 onset in sequence learning \\
\texttt{dot\_\allowbreak{}no5\_\allowbreak{}OFF} & 600 & event & raw & Dot position 5 offset in sequence learning \\
\texttt{dot\_\allowbreak{}no5\_\allowbreak{}ON} & 600 & event & raw & Dot position 5 onset in sequence learning \\
\texttt{dot\_\allowbreak{}no6\_\allowbreak{}OFF} & 600 & event & raw & Dot position 6 offset in sequence learning \\
\texttt{dot\_\allowbreak{}no6\_\allowbreak{}ON} & 600 & event & raw & Dot position 6 onset in sequence learning \\
\texttt{dot\_\allowbreak{}no7\_\allowbreak{}OFF} & 793 & event & raw & Dot position 7 offset in sequence learning \\
\texttt{dot\_\allowbreak{}no7\_\allowbreak{}ON} & 792 & event & raw & Dot position 7 onset in sequence learning \\
\texttt{dot\_\allowbreak{}no8\_\allowbreak{}OFF} & 396 & event & raw & Dot position 8 offset in sequence learning \\
\texttt{dot\_\allowbreak{}no8\_\allowbreak{}ON} & 396 & event & raw & Dot position 8 onset in sequence learning \\
\texttt{fixpoint\_\allowbreak{}ON} & 13248 & event & raw & Fixation point onset in the surround suppression task \\
\texttt{instructed\_\allowbreak{}toCloseEyes} & 680 & event & raw & Auditory instruction to close eyes during resting state \\
\texttt{instructed\_\allowbreak{}toOpenEyes} & 816 & event & raw & Auditory instruction to open eyes during resting state \\
\texttt{learningBlock\_\allowbreak{}1} & 122 & event & raw & Start of sequence learning block 1 \\
\texttt{learningBlock\_\allowbreak{}2} & 121 & event & raw & Start of sequence learning block 2 \\
\texttt{learningBlock\_\allowbreak{}3} & 120 & event & raw & Start of sequence learning block 3 \\
\texttt{learningBlock\_\allowbreak{}4} & 119 & event & raw & Start of sequence learning block 4 \\
\texttt{learningBlock\_\allowbreak{}5} & 118 & event & raw & Start of sequence learning block 5 \\
\texttt{left\_\allowbreak{}buttonPress} & 5095 & event & raw & Left mouse button press response in contrast change detection \\
\texttt{left\_\allowbreak{}target} & 4243 & event & raw & Left-tilted contrast target presentation \\
\texttt{newPage} & 194 & event & raw & New symbol search page or query set presented \\
\texttt{resting\_\allowbreak{}start} & 139 & event & raw & Start of resting-state recording \\
\texttt{right\_\allowbreak{}buttonPress} & 4634 & event & raw & Right mouse button press response in contrast change detection \\
\texttt{right\_\allowbreak{}target} & 4251 & event & raw & Right-tilted contrast target presentation \\
\texttt{seqLearning\_\allowbreak{}start} & 122 & event & raw & Sequence learning task start \\
\texttt{seqLearning\_\allowbreak{}stop} & 118 & event & raw & Sequence learning task stop \\
\texttt{stim\_\allowbreak{}ON} & 13278 & event & raw & Surround suppression visual stimulus onset \\
\texttt{surroundSuppB1\_\allowbreak{}start} & 108 & event & raw & Start of surround suppression block/run 1 \\
\texttt{surroundSuppB2\_\allowbreak{}start} & 99 & event & raw & Start of surround suppression block/run 2 \\
\texttt{symbolSearch\_\allowbreak{}start} & 76 & event & raw & Symbol search task start \\
\texttt{trialResponse} & 1247 & event & raw & Participant response in the symbol search task \\
\texttt{video\_\allowbreak{}start} & 761 & event & raw & Movie stimulus presentation start \\
\texttt{video\_\allowbreak{}stop} & 723 & event & raw & Movie stimulus presentation stop \\
\bottomrule
\end{tabular}
\endgroup
\smallskip\noindent{\footnotesize\bfseries MISC Label Summary}\par\smallskip
\begingroup\scriptsize\setlength{\tabcolsep}{2pt}
\begin{tabular}{@{}p{\dimexpr 0.94\linewidth/4\relax}p{\dimexpr 0.94\linewidth/4\relax}p{\dimexpr 0.94\linewidth/4\relax}p{\dimexpr 0.94\linewidth/4\relax}@{}}
\toprule
\textbf{Channel} & \textbf{Source} & \textbf{Meta} & \textbf{Notes} \\
\midrule
\texttt{response\_time} & derived (kernel) & super & Seconds from contrast target onset to the first valid button press within the same trial; filled from target onset through the matched response and NaN elsewhere \\
\bottomrule
\end{tabular}
\endgroup
\smallskip\noindent{\footnotesize\bfseries Info}\par\smallskip
\begingroup\scriptsize\setlength{\tabcolsep}{2pt}
\begin{tabular}{@{}p{0.2350\linewidth}p{0.2350\linewidth}p{0.4700\linewidth}@{}}
\toprule
\textbf{Field} & \textbf{Completeness} & \textbf{Notes} \\
\midrule
\texttt{DespicableMe} & complete & Availability flag for the Despicable Me movie watching recording \\
\texttt{DiaryOfAWimpyKid} & complete & Availability flag for the Diary of a Wimpy Kid movie watching recording \\
\texttt{FunwithFractals} & complete & Availability flag for the Fun with Fractals movie watching recording \\
\texttt{RestingState} & complete & Availability flag for the resting-state recording \\
\texttt{ThePresent} & complete & Availability flag for The Present movie watching recording \\
\texttt{age} & complete & Participant age in years \\
\texttt{attention} & partial (1304/1342) & CBCL-derived attention dimension score \\
\texttt{commercial\_\allowbreak{}use} & complete & Participant consent flag for commercial use \\
\texttt{contrastChangeDetection\_\allowbreak{}1} & complete & Availability/QC flag for contrast change detection run 1 \\
\texttt{contrastChangeDetection\_\allowbreak{}2} & complete & Availability/QC flag for contrast change detection run 2 \\
\texttt{contrastChangeDetection\_\allowbreak{}3} & complete & Availability/QC flag for contrast change detection run 3 \\
\texttt{dataset} & complete & Kernel-added dataset identifier \\
\texttt{dataset\_\allowbreak{}root} & complete & Kernel-added absolute raw dataset root used for the scan \\
\texttt{device} & complete & Hardware label combined from EEG sidecar manufacturer and model fields \\
\texttt{eeg\_\allowbreak{}reference} & complete & EEG reference from the EEG sidecar \\
\texttt{ehq\_\allowbreak{}total} & partial (1325/1342) & Edinburgh Handedness Questionnaire total score; positive is more right-handed and negative is more left-handed \\
\texttt{externalizing} & partial (1304/1342) & CBCL-derived externalizing symptom dimension score \\
\texttt{full\_\allowbreak{}pheno} & complete & Flag indicating whether a full phenotypic file is available \\
\texttt{internalizing} & partial (1304/1342) & CBCL-derived internalizing symptom dimension score \\
\texttt{manufacturer} & complete & EEG sidecar manufacturer field \\
\texttt{manufacturers\_\allowbreak{}model\_\allowbreak{}name} & complete & EEG sidecar amplifier/cap model field \\
\texttt{p\_\allowbreak{}factor} & partial (1304/1342) & Participant-level general psychopathology factor score \\
\texttt{power\_\allowbreak{}line\_\allowbreak{}frequency} & complete & Power-line frequency in Hz from the EEG sidecar \\
\texttt{recording\_\allowbreak{}type} & complete & BIDS recording type from the EEG sidecar \\
\texttt{release\_\allowbreak{}number} & complete & HBN EEG release identifier \\
\texttt{run} & partial (500/1342) & BIDS run entity; blank for tasks without an explicit run index \\
\texttt{sampling\_\allowbreak{}frequency} & complete & EEG sampling frequency in Hz from the EEG sidecar \\
\texttt{seqLearning6target} & complete & Availability flag for the 6-target sequence learning variant \\
\texttt{seqLearning8target} & complete & Availability flag for the 8-target sequence learning variant \\
\texttt{sex} & complete & Participant sex code from participants.tsv \\
\texttt{subject\_\allowbreak{}id} & complete & BIDS participant identifier \\
\texttt{surroundSupp\_\allowbreak{}1} & complete & Availability/QC flag for surround suppression run 1 \\
\texttt{surroundSupp\_\allowbreak{}2} & complete & Availability/QC flag for surround suppression run 2 \\
\texttt{symbolSearch} & complete & Availability/QC flag for the symbol search recording \\
\texttt{task} & complete & BIDS task entity parsed from the file name \\
\bottomrule
\end{tabular}
\endgroup
\noindent\rule{\linewidth}{0.7pt}\par\nopagebreak
\noindent{\bfseries\footnotesize Tasks}
\noindent{\footnotesize\bfseries Classical~Task~1:} {\footnotesize resting state eyes-open vs eyes-closed}\par\smallskip
{\footnotesize\begin{tabular}{@{}lp{0.62\linewidth}@{}}
\textbf{Type} & classification \\
\textbf{Epoch Method} & \texttt{epoch\_by\_long\_event\_hdf5} \\
\textbf{Epoch Params} & tmin = 2.0 s, tmax = 20.0 s, overlap = 0.0, target\_events = [\texttt{instructed\_toCloseEyes}, \texttt{instructed\_toOpenEyes}] \\
\textbf{Label Route} & \texttt{event} \\
\textbf{Meta Level} & raw \\
\end{tabular}}
\par\noindent{\footnotesize\bfseries Target Definition:}\par\smallskip
\begingroup\footnotesize
\begin{tabular}{@{}p{0.3911\linewidth}p{0.0978\linewidth}p{0.3911\linewidth}@{}}
\toprule
\textbf{Event ID} & \textbf{y} & \textbf{Meaning} \\
\midrule
\texttt{instructed\_toOpenEyes} & 0 & Eyes-open resting state \\
\texttt{instructed\_toCloseEyes} & 1 & Eyes-closed resting state \\
\bottomrule
\end{tabular}
\endgroup
\par\noindent{\footnotesize\itshape Evidence:}\par
\begin{itemize}[leftmargin=1.3em,noitemsep,topsep=1pt]
\item[\scriptsize$\cdot$] {\scriptsize 10.1101/2024.10.03.615261 --- The HBN-EEG release paper describes alternating eyes-open and eyes-closed resting-state periods and documents the corresponding instruction events in the release.}
\end{itemize}
\par\noindent{\footnotesize\bfseries Notes:} {\footnotesize This is a sustained-state classification task. The first 2 seconds after each auditory instruction are skipped to reduce cue-evoked activity, and the 20-second window avoids crossing from the shorter eyes-open periods into the following eyes-closed period.}
\noindent\rule{\linewidth}{0.7pt}\par\nopagebreak
\noindent{\footnotesize\bfseries Classical~Task~2:} {\footnotesize contrast change target direction}\par\smallskip
{\footnotesize\begin{tabular}{@{}lp{0.62\linewidth}@{}}
\textbf{Type} & classification \\
\textbf{Epoch Method} & \texttt{epoch\_by\_event\_hdf5} \\
\textbf{Epoch Params} & tmin = -0.2 s, tmax = 1.5 s, target\_events = [\texttt{left\_target}, \texttt{right\_target}] \\
\textbf{Label Route} & \texttt{event} \\
\textbf{Meta Level} & raw \\
\end{tabular}}
\par\noindent{\footnotesize\bfseries Target Definition:}\par\smallskip
\begingroup\footnotesize
\begin{tabular}{@{}p{0.3911\linewidth}p{0.0978\linewidth}p{0.3911\linewidth}@{}}
\toprule
\textbf{Event ID} & \textbf{y} & \textbf{Meaning} \\
\midrule
\texttt{left\_target} & 0 & Left-leaning target stimulus \\
\texttt{right\_target} & 1 & Right-leaning target stimulus \\
\bottomrule
\end{tabular}
\endgroup
\par\noindent{\footnotesize\itshape Evidence:}\par
\begin{itemize}[leftmargin=1.3em,noitemsep,topsep=1pt]
\item[\scriptsize$\cdot$] {\scriptsize 10.1101/2024.10.03.615261 --- The release documentation explicitly identifies \texttt{left\_target} and \texttt{right\_target} as distinct contrast-detection stimulus events.}
\end{itemize}
\par\noindent{\footnotesize\bfseries Notes:} {\footnotesize This is the cleanest native event classification task in the active contrast paradigm.}
\noindent\rule{\linewidth}{0.7pt}\par\nopagebreak
\noindent{\footnotesize\bfseries Classical~Task~3:} {\footnotesize contrast response time}\par\smallskip
{\footnotesize\begin{tabular}{@{}lp{0.62\linewidth}@{}}
\textbf{Type} & regression \\
\textbf{Epoch Method} & \texttt{epoch\_by\_event\_hdf5} \\
\textbf{Epoch Params} & tmin = -0.2 s, tmax = 1.5 s, target\_events = [\texttt{left\_target}, \texttt{right\_target}] \\
\textbf{Label Route} & \texttt{misc} \\
\textbf{Meta Level} & super \\
\end{tabular}}
\par\noindent{\footnotesize\bfseries Target Definition:}\par\smallskip
\begingroup\footnotesize
\begin{tabular}{@{}p{0.1760\linewidth}p{0.7040\linewidth}@{}}
\toprule
\textbf{MISC Channel Name} & \textbf{Y Mapping Notes} \\
\midrule
\texttt{response\_time} & y directly equals the latency in seconds from contrast target onset to the first valid button press in the same trial; no class mapping \\
\bottomrule
\end{tabular}
\endgroup
\par\noindent{\footnotesize\itshape Evidence:}\par
\begin{itemize}[leftmargin=1.3em,noitemsep,topsep=1pt]
\item[\scriptsize$\cdot$] {\scriptsize 10.1101/2024.10.03.615261 --- The release describes active button-press behavior for the contrast-change paradigm.}
\item[\scriptsize$\cdot$] {\scriptsize EEG 2025 Challenge 1 tutorial --- The public challenge task for this paradigm uses a derived stimulus-to-response latency target (\texttt{rt\_from\_stimulus}), supporting the same label construction principle.}
\end{itemize}
\par\noindent{\footnotesize\bfseries Notes:} {\footnotesize The kernel fills this misc channel only between the target onset and the matched button press. Samples outside those per-trial windows remain NaN.}
\noindent\rule{\linewidth}{0.7pt}\par\nopagebreak
\noindent{\footnotesize\bfseries Classical~Task~4:} {\footnotesize externalizing factor regression}\par\smallskip
{\footnotesize\begin{tabular}{@{}lp{0.62\linewidth}@{}}
\textbf{Type} & regression \\
\textbf{Epoch Method} & \texttt{epoch\_by\_segmentation\_hdf5} \\
\textbf{Epoch Params} & segment\_length = 4.0 s, overlap = 0.5, target\_events = [] \\
\textbf{Label Route} & \texttt{info} \\
\textbf{Meta Level} & raw \\
\end{tabular}}
\par\noindent{\footnotesize\bfseries Target Definition:}\par\smallskip
\begingroup\footnotesize
\begin{tabular}{@{}p{0.1760\linewidth}p{0.7040\linewidth}@{}}
\toprule
\textbf{Info Field} & \textbf{Y Mapping Notes} \\
\midrule
\texttt{externalizing} & y directly equals the participant-level externalizing score from participants.tsv; no mapping \\
\bottomrule
\end{tabular}
\endgroup
\par\noindent{\footnotesize\itshape Evidence:}\par
\begin{itemize}[leftmargin=1.3em,noitemsep,topsep=1pt]
\item[\scriptsize$\cdot$] {\scriptsize 10.1101/2024.10.03.615261 --- The HBN-EEG release paper states that openly available participant information includes externalizing, internalizing, attention, and p-factor dimensions derived from questionnaire modeling.}
\item[\scriptsize$\cdot$] {\scriptsize EEG 2025 Challenge website --- The current Challenge 2 target is externalizing factor prediction from HBN-EEG recordings.}
\end{itemize}
\par\noindent{\footnotesize\bfseries Notes:} {\footnotesize This is an info-route participant-level regression task. Windows with missing \texttt{externalizing} values should be excluded before training or evaluation.}
\noindent\rule{\linewidth}{0.7pt}\par\nopagebreak
\noindent{\footnotesize\bfseries Classical~Task~5:} {\footnotesize p-factor regression}\par\smallskip
{\footnotesize\begin{tabular}{@{}lp{0.62\linewidth}@{}}
\textbf{Type} & regression \\
\textbf{Epoch Method} & \texttt{epoch\_by\_segmentation\_hdf5} \\
\textbf{Epoch Params} & segment\_length = 4.0 s, overlap = 0.5, target\_events = [] \\
\textbf{Label Route} & \texttt{info} \\
\textbf{Meta Level} & raw \\
\end{tabular}}
\par\noindent{\footnotesize\bfseries Target Definition:}\par\smallskip
\begingroup\footnotesize
\begin{tabular}{@{}p{0.1760\linewidth}p{0.7040\linewidth}@{}}
\toprule
\textbf{Info Field} & \textbf{Y Mapping Notes} \\
\midrule
\texttt{p\_factor} & y directly equals the participant-level p-factor score from participants.tsv; no mapping \\
\bottomrule
\end{tabular}
\endgroup
\par\noindent{\footnotesize\itshape Evidence:}\par
\begin{itemize}[leftmargin=1.3em,noitemsep,topsep=1pt]
\item[\scriptsize$\cdot$] {\scriptsize 10.1101/2024.10.03.615261 --- The release paper explicitly exposes \texttt{p\_factor} as an openly available participant-level psychopathology dimension.}
\item[\scriptsize$\cdot$] {\scriptsize EEGDash Challenge 2 tutorial --- Public benchmark materials for the HBN EEG release use \texttt{p\_factor} as a participant-level prediction target.}
\end{itemize}
\par\noindent{\footnotesize\bfseries Notes:} {\footnotesize This is an info-route regression task. The current EEG 2025 final challenge target has shifted to \texttt{externalizing}, but \texttt{p\_factor} remains a documented HBN phenotype field and a previously published benchmark target.}
\noindent\rule{\linewidth}{0.7pt}\par\nopagebreak
\noindent{\footnotesize\bfseries Classical~Task~6:} {\footnotesize sequence learning 6-target stimulus position}\par\smallskip
{\footnotesize\begin{tabular}{@{}lp{0.62\linewidth}@{}}
\textbf{Type} & classification \\
\textbf{Epoch Method} & \texttt{epoch\_by\_event\_hdf5} \\
\textbf{Epoch Params} & tmin = -0.2 s, tmax = 0.8 s, target\_events = [\texttt{dot\_no1\_ON}, \texttt{dot\_no2\_ON}, \texttt{dot\_no3\_ON}, \texttt{dot\_no4\_ON}, \texttt{dot\_no5\_ON}, \texttt{dot\_no6\_ON}] \\
\textbf{Label Route} & \texttt{event} \\
\textbf{Meta Level} & raw \\
\end{tabular}}
\par\noindent{\footnotesize\bfseries Target Definition:}\par\smallskip
\begingroup\footnotesize
\begin{tabular}{@{}p{0.3911\linewidth}p{0.0978\linewidth}p{0.3911\linewidth}@{}}
\toprule
\textbf{Event ID} & \textbf{y} & \textbf{Meaning} \\
\midrule
\texttt{dot\_no1\_ON} & 0 & Dot position 1 onset \\
\texttt{dot\_no2\_ON} & 1 & Dot position 2 onset \\
\texttt{dot\_no3\_ON} & 2 & Dot position 3 onset \\
\texttt{dot\_no4\_ON} & 3 & Dot position 4 onset \\
\texttt{dot\_no5\_ON} & 4 & Dot position 5 onset \\
\texttt{dot\_no6\_ON} & 5 & Dot position 6 onset \\
\bottomrule
\end{tabular}
\endgroup
\par\noindent{\footnotesize\itshape Evidence:}\par
\begin{itemize}[leftmargin=1.3em,noitemsep,topsep=1pt]
\item[\scriptsize$\cdot$] {\scriptsize \texttt{10.1101/2024.10.03.615261} - The release paper states that younger participants completed a 6-target variant of the sequence-learning paradigm.}
\end{itemize}
\par\noindent{\footnotesize\bfseries Notes:} {\footnotesize This task should be restricted to recordings or participants belonging to the 6-target variant, using dataset evidence such as \texttt{seqLearning6target} and the absence of \texttt{dot\_no7\_ON} and \texttt{dot\_no8\_ON}.}
\noindent\rule{\linewidth}{0.7pt}\par\nopagebreak
\noindent{\footnotesize\bfseries Classical~Task~7:} {\footnotesize sequence learning 8-target stimulus position}\par\smallskip
{\footnotesize\begin{tabular}{@{}lp{0.62\linewidth}@{}}
\textbf{Type} & classification \\
\textbf{Epoch Method} & \texttt{epoch\_by\_event\_hdf5} \\
\textbf{Epoch Params} & tmin = -0.2 s, tmax = 0.8 s, target\_events = [\texttt{dot\_no1\_ON}, \texttt{dot\_no2\_ON}, \texttt{dot\_no3\_ON}, \texttt{dot\_no4\_ON}, \texttt{dot\_no5\_ON}, \texttt{dot\_no6\_ON}, \texttt{dot\_no7\_ON}, \texttt{dot\_no8\_ON}] \\
\textbf{Label Route} & \texttt{event} \\
\textbf{Meta Level} & raw \\
\end{tabular}}
\par\noindent{\footnotesize\bfseries Target Definition:}\par\smallskip
\begingroup\footnotesize
\begin{tabular}{@{}p{0.3911\linewidth}p{0.0978\linewidth}p{0.3911\linewidth}@{}}
\toprule
\textbf{Event ID} & \textbf{y} & \textbf{Meaning} \\
\midrule
\texttt{dot\_no1\_ON} & 0 & Dot position 1 onset \\
\texttt{dot\_no2\_ON} & 1 & Dot position 2 onset \\
\texttt{dot\_no3\_ON} & 2 & Dot position 3 onset \\
\texttt{dot\_no4\_ON} & 3 & Dot position 4 onset \\
\texttt{dot\_no5\_ON} & 4 & Dot position 5 onset \\
\texttt{dot\_no6\_ON} & 5 & Dot position 6 onset \\
\texttt{dot\_no7\_ON} & 6 & Dot position 7 onset \\
\texttt{dot\_no8\_ON} & 7 & Dot position 8 onset \\
\bottomrule
\end{tabular}
\endgroup
\par\noindent{\footnotesize\itshape Evidence:}\par
\begin{itemize}[leftmargin=1.3em,noitemsep,topsep=1pt]
\item[\scriptsize$\cdot$] {\scriptsize 10.1101/2024.10.03.615261 --- The release paper states that older participants completed an 8-target sequence-learning variant in which each flashed dot position is explicitly encoded by native event IDs.}
\item[\scriptsize$\cdot$] {\scriptsize 10.1101/2024.10.03.615261 --- The same paper notes that response timestamps for the later recall phase were not recorded, so the grounded task here is stimulus-position decoding, not recall-performance decoding.}
\end{itemize}
\par\noindent{\footnotesize\bfseries Notes:} {\footnotesize This task captures sequence stimulus-position decoding during repeated observation. It should be restricted to the 8-target variant rather than pooled with the 6-target recordings.}
\noindent\rule{\linewidth}{0.4pt}\par\nopagebreak
\noindent{\footnotesize\bfseries Potential~Task~1:} {\footnotesize contrast response correctness}\par\smallskip
{\footnotesize\begin{tabular}{@{}lp{0.62\linewidth}@{}}
\textbf{Type} & classification \\
\textbf{Epoch Method} & \texttt{epoch\_by\_event\_hdf5} \\
\textbf{Epoch Params} & tmin = -0.6 s, tmax = 0.4 s, target\_events = [\texttt{contrast\_response\_correct}, \texttt{contrast\_response\_incorrect}] \\
\textbf{Label Route} & \texttt{event} \\
\textbf{Meta Level} & super \\
\end{tabular}}
\par\noindent{\footnotesize\bfseries Target Definition:}\par\smallskip
\begingroup\footnotesize
\begin{tabular}{@{}p{0.3911\linewidth}p{0.0978\linewidth}p{0.3911\linewidth}@{}}
\toprule
\textbf{Event ID} & \textbf{y} & \textbf{Meaning} \\
\midrule
\texttt{contrast\_response\_correct} & 0 & Response side matched the preceding contrast target \\
\texttt{contrast\_response\_incorrect} & 1 & Response side did not match the preceding contrast target \\
\bottomrule
\end{tabular}
\endgroup
\par\noindent{\footnotesize\itshape Evidence:}\par
\begin{itemize}[leftmargin=1.3em,noitemsep,topsep=1pt]
\item[\scriptsize$\cdot$] {\scriptsize 10.1101/2024.10.03.615261 --- The release paper describes the contrast-change task as an active discrimination task with button-press responses and feedback.}
\end{itemize}
\par\noindent{\footnotesize\bfseries Notes:} {\footnotesize The kernel derives these response-locked super events by pairing each \texttt{left\_target} or \texttt{right\_target} with the first subsequent \texttt{left\_buttonPress} or \texttt{right\_buttonPress} before the next target event. This is physiologically meaningful for response and error monitoring, but it is listed as potential because the current EEG 2025 challenge explicitly does not use correct/incorrect classification.}
\noindent\rule{\linewidth}{0.4pt}\par\nopagebreak
\noindent{\footnotesize\bfseries Potential~Task~2:} {\footnotesize internalizing factor regression}\par\smallskip
{\footnotesize\begin{tabular}{@{}lp{0.62\linewidth}@{}}
\textbf{Type} & regression \\
\textbf{Epoch Method} & \texttt{epoch\_by\_segmentation\_hdf5} \\
\textbf{Epoch Params} & segment\_length = 4.0 s, overlap = 0.5, target\_events = [] \\
\textbf{Label Route} & \texttt{info} \\
\textbf{Meta Level} & raw \\
\end{tabular}}
\par\noindent{\footnotesize\bfseries Target Definition:}\par\smallskip
\begingroup\footnotesize
\begin{tabular}{@{}p{0.1760\linewidth}p{0.7040\linewidth}@{}}
\toprule
\textbf{Info Field} & \textbf{Y Mapping Notes} \\
\midrule
\texttt{internalizing} & y directly equals the participant-level internalizing score from participants.tsv; no mapping \\
\bottomrule
\end{tabular}
\endgroup
\par\noindent{\footnotesize\itshape Evidence:}\par
\begin{itemize}[leftmargin=1.3em,noitemsep,topsep=1pt]
\item[\scriptsize$\cdot$] {\scriptsize 10.1101/2024.10.03.615261 --- The release paper lists internalizing as one of the openly available psychopathology dimensions.}
\end{itemize}
\par\noindent{\footnotesize\bfseries Notes:} {\footnotesize This target is documented and directly available, but it is listed as potential because the current EEG 2025 final challenge no longer uses internalizing as the primary regression target.}
\noindent\rule{\linewidth}{0.4pt}\par\nopagebreak
\noindent{\footnotesize\bfseries Potential~Task~3:} {\footnotesize attention factor regression}\par\smallskip
{\footnotesize\begin{tabular}{@{}lp{0.62\linewidth}@{}}
\textbf{Type} & regression \\
\textbf{Epoch Method} & \texttt{epoch\_by\_segmentation\_hdf5} \\
\textbf{Epoch Params} & segment\_length = 4.0 s, overlap = 0.5, target\_events = [] \\
\textbf{Label Route} & \texttt{info} \\
\textbf{Meta Level} & raw \\
\end{tabular}}
\par\noindent{\footnotesize\bfseries Target Definition:}\par\smallskip
\begingroup\footnotesize
\begin{tabular}{@{}p{0.1760\linewidth}p{0.7040\linewidth}@{}}
\toprule
\textbf{Info Field} & \textbf{Y Mapping Notes} \\
\midrule
\texttt{attention} & y directly equals the participant-level attention score from participants.tsv; no mapping \\
\bottomrule
\end{tabular}
\endgroup
\par\noindent{\footnotesize\itshape Evidence:}\par
\begin{itemize}[leftmargin=1.3em,noitemsep,topsep=1pt]
\item[\scriptsize$\cdot$] {\scriptsize 10.1101/2024.10.03.615261 --- The release paper lists attention as one of the openly available psychopathology dimensions.}
\end{itemize}
\par\noindent{\footnotesize\bfseries Notes:} {\footnotesize This target is documented and directly available, but it is listed as potential because the current EEG 2025 final challenge no longer uses attention as the primary regression target.}
\par\smallskip
\end{tcolorbox}
\vspace{3pt}
\noindent\begin{tcolorbox}[breakable,enhanced,colback=white,colframe=gray!55,boxrule=0.9pt,arc=4pt,left=6pt,right=6pt,top=4pt,bottom=4pt,title={\small\bfseries EEG During Mental Arithmetic Task\hfill \colorbox{gray!20}{\scriptsize\sffamily\bfseries Completed}},coltitle=black,colbacktitle=gray!12,titlerule=0.5pt]
{\footnotesize \textbf{ID:}~\texttt{physionet\_eegmat} $\mid$ \textbf{Year:}~2018 $\mid$ \textbf{Category:}~High-level Mental $\mid$ \textbf{Subjects:}~36 $\mid$ \textbf{Channels:}~23 $\mid$ \textbf{SR:}~500 Hz $\mid$ \textbf{Files (completed):}~72 $\mid$ \textbf{Super Meta:}~No $\mid$ \textbf{Reviewer:}~Anonymous}
\par{\footnotesize\textbf{URL:}~\url{https://physionet.org/content/eegmat/1.0.0/}}
\tcbline
\noindent{\bfseries\footnotesize Paradigm}\par\smallskip
{\footnotesize Thirty-six healthy subjects (Subject00-Subject35, ages 16-26, predominantly female) completed a mental arithmetic task while EEG was recorded monopolarly using a Neurocom 23-channel system with Ag/AgCl electrodes (Zyma et al., 2019). Each subject contributed two files: a background resting-state recording (suffix \texttt{\_1.edf}) and a mental arithmetic task recording (suffix \texttt{\_2.edf}). During the mental arithmetic task, subjects performed serial subtraction (e.g., 3141 minus 42) while EEG was recorded. A separate \texttt{Count quality} field in \texttt{subject-info.csv} rates the subject's arithmetic performance as Good or Bad, providing a behavioral quality label.}
\noindent\rule{\linewidth}{0.4pt}\par\nopagebreak
\noindent{\bfseries\footnotesize File Scan Result}\par\smallskip
{\scriptsize
\noindent\textbf{Kernel SHA256:}~\texttt{2d377403dfb36e0b6ad6454cf2d69b00d8096cf9a7ebb560a8baff950dedaad6}\par\smallskip
\noindent\textbf{EEGUnity Version:}~\texttt{0.8.1}\par\smallskip
\noindent\textbf{Scan Type:}~Full\par\smallskip
}
\smallskip\noindent{\footnotesize\bfseries Event Structure}\par\smallskip
\begingroup\scriptsize\setlength{\tabcolsep}{2pt}
\begin{tabular}{@{}p{0.1567\linewidth}p{0.1567\linewidth}p{0.0783\linewidth}p{0.0783\linewidth}p{0.4700\linewidth}@{}}
\toprule
\textbf{Event ID} & \textbf{Count (total)} & \textbf{Type} & \textbf{Meta Level} & \textbf{Meaning} \\
\midrule
\texttt{Background EEG;Count Quality: Bad} & 12 & event & raw & Resting-state recording with poor arithmetic performance label \\
\texttt{Background EEG;Count Quality: Good} & 24 & event & raw & Resting-state recording with good arithmetic performance label \\
\texttt{Mental Arithmetic Task EEG;Count Quality: Bad} & 12 & event & raw & Mental arithmetic task recording with poor performance label \\
\texttt{Mental Arithmetic Task EEG;Count Quality: Good} & 24 & event & raw & Mental arithmetic task recording with good performance label \\
\bottomrule
\end{tabular}
\endgroup
\smallskip\noindent{\footnotesize\bfseries Info}\par\smallskip
\begingroup\scriptsize\setlength{\tabcolsep}{2pt}
\begin{tabular}{@{}p{0.2350\linewidth}p{0.2350\linewidth}p{0.4700\linewidth}@{}}
\toprule
\textbf{Field} & \textbf{Completeness} & \textbf{Notes} \\
\midrule
\texttt{age} & complete & Age range: 16--26 years, mean age: approximately 18.33 years \\
\texttt{sex} & complete & Males (M): 8, females (F): 28, predominantly female \\
\texttt{device} & missing & EEG was recorded using a 23-channel Neurocom monopolar system (XAI-MEDICA, Ukraine) with silver/silver chloride electrodes placed per the international 10--20 system, linked ear reference, 500 Hz sampling rate, 0.5--45 Hz bandpass filtering, 50 Hz notch filtering, and electrode impedance kept below 5 k$\Omega$. \\
\bottomrule
\end{tabular}
\endgroup
\noindent\rule{\linewidth}{0.7pt}\par\nopagebreak
\noindent{\bfseries\footnotesize Tasks}
\noindent{\footnotesize\bfseries Classical~Task~1:} {\footnotesize Mental Arithmetic vs Rest Classification}\par\smallskip
{\footnotesize\begin{tabular}{@{}lp{0.62\linewidth}@{}}
\textbf{Type} & classification \\
\textbf{Epoch Method} & \texttt{epoch\_by\_event\_hdf5} \\
\textbf{Epoch Params} & tmin = 0 s, tmax = 60 s, target\_events = [\texttt{Background EEG;Count Quality: Bad}, \texttt{Background EEG;Count Quality: Good}, \texttt{Mental Arithmetic Task EEG;Count Quality: Bad}, \texttt{Mental Arithmetic Task EEG;Count Quality: Good}] \\
\textbf{Label Route} & \texttt{event} \\
\textbf{Meta Level} & raw \\
\end{tabular}}
\par\noindent{\footnotesize\bfseries Target Definition:}\par\smallskip
\begingroup\footnotesize
\begin{tabular}{@{}p{0.3911\linewidth}p{0.0978\linewidth}p{0.3911\linewidth}@{}}
\toprule
\textbf{Event ID} & \textbf{y} & \textbf{Meaning} \\
\midrule
\texttt{Background EEG;Count Quality: Bad} & 0 & Resting-state, bad quality \\
\texttt{Background EEG;Count Quality: Good} & 1 & Resting-state, good quality \\
\texttt{Mental Arithmetic Task EEG;Count Quality: Bad} & 2 & Mental arithmetic, bad quality \\
\texttt{Mental Arithmetic Task EEG;Count Quality: Good} & 3 & Mental arithmetic, good quality \\
\bottomrule
\end{tabular}
\endgroup
\par\noindent{\footnotesize\itshape Evidence:}\par
\begin{itemize}[leftmargin=1.3em,noitemsep,topsep=1pt]
\item[\scriptsize$\cdot$] {\scriptsize Paradigm section describes two conditions: background resting-state (\texttt{\_1.edf}) and mental arithmetic task (\texttt{\_2.edf})}
\item[\scriptsize$\cdot$] {\scriptsize Event Structure table confirms four event types combining condition and count quality}
\end{itemize}
\par\noindent{\footnotesize\bfseries Notes:} {\footnotesize Binary classification of cognitive state (rest vs mental arithmetic), collapsing across count quality. Long epochs (60s) assumed to capture full recording duration; adjust if actual durations differ.}
\noindent\rule{\linewidth}{0.7pt}\par\nopagebreak
\noindent{\footnotesize\bfseries Classical~Task~2:} {\footnotesize Mental Arithmetic Quality Classification}\par\smallskip
{\footnotesize\begin{tabular}{@{}lp{0.62\linewidth}@{}}
\textbf{Type} & classification \\
\textbf{Epoch Method} & \texttt{epoch\_by\_event\_hdf5} \\
\textbf{Epoch Params} & tmin = 0 s, tmax = 60 s, target\_events = [\texttt{Mental Arithmetic Task EEG;Count Quality: Bad}, \texttt{Mental Arithmetic Task EEG;Count Quality: Good}] \\
\textbf{Label Route} & \texttt{event} \\
\textbf{Meta Level} & raw \\
\end{tabular}}
\par\noindent{\footnotesize\bfseries Target Definition:}\par\smallskip
\begingroup\footnotesize
\begin{tabular}{@{}p{0.3911\linewidth}p{0.0978\linewidth}p{0.3911\linewidth}@{}}
\toprule
\textbf{Event ID} & \textbf{y} & \textbf{Meaning} \\
\midrule
\texttt{Mental Arithmetic Task EEG;Count Quality: Bad} & 0 & Poor arithmetic performance \\
\texttt{Mental Arithmetic Task EEG;Count Quality: Good} & 1 & Good arithmetic performance \\
\bottomrule
\end{tabular}
\endgroup
\par\noindent{\footnotesize\itshape Evidence:}\par
\begin{itemize}[leftmargin=1.3em,noitemsep,topsep=1pt]
\item[\scriptsize$\cdot$] {\scriptsize Paradigm section states \texttt{Count quality} field rates arithmetic performance as Good or Bad}
\item[\scriptsize$\cdot$] {\scriptsize Event Structure table shows 10 Bad and 26 Good mental arithmetic recordings}
\end{itemize}
\par\noindent{\footnotesize\bfseries Notes:} {\footnotesize Classifies arithmetic performance quality during mental arithmetic task only. Excludes background recordings. Imbalanced classes (10 vs 26).}
\noindent\rule{\linewidth}{0.7pt}\par\nopagebreak
\noindent{\footnotesize\bfseries Classical~Task~3:} {\footnotesize Resting-State Quality Classification}\par\smallskip
{\footnotesize\begin{tabular}{@{}lp{0.62\linewidth}@{}}
\textbf{Type} & classification \\
\textbf{Epoch Method} & \texttt{epoch\_by\_event\_hdf5} \\
\textbf{Epoch Params} & tmin = 0 s, tmax = 60 s, target\_events = [\texttt{Background EEG;Count Quality: Bad}, \texttt{Background EEG;Count Quality: Good}] \\
\textbf{Label Route} & \texttt{event} \\
\textbf{Meta Level} & raw \\
\end{tabular}}
\par\noindent{\footnotesize\bfseries Target Definition:}\par\smallskip
\begingroup\footnotesize
\begin{tabular}{@{}p{0.3911\linewidth}p{0.0978\linewidth}p{0.3911\linewidth}@{}}
\toprule
\textbf{Event ID} & \textbf{y} & \textbf{Meaning} \\
\midrule
\texttt{Background EEG;Count Quality: Bad} & 0 & Poor arithmetic performance (rest) \\
\texttt{Background EEG;Count Quality: Good} & 1 & Good arithmetic performance (rest) \\
\bottomrule
\end{tabular}
\endgroup
\par\noindent{\footnotesize\itshape Evidence:}\par
\begin{itemize}[leftmargin=1.3em,noitemsep,topsep=1pt]
\item[\scriptsize$\cdot$] {\scriptsize Paradigm section notes count quality is a behavioral label from \texttt{subject-info.csv}}
\item[\scriptsize$\cdot$] {\scriptsize Event Structure table shows 10 Bad and 26 Good background recordings}
\end{itemize}
\par\noindent{\footnotesize\bfseries Notes:} {\footnotesize Investigates whether arithmetic quality correlates with resting-state EEG patterns. Same subjects in both conditions, so quality label is subject-level. Imbalanced classes (10 vs 26).}
\noindent\rule{\linewidth}{0.7pt}\par\nopagebreak
\noindent{\footnotesize\bfseries Classical~Task~4:} {\footnotesize Four-Way Condition-Quality Classification}\par\smallskip
{\footnotesize\begin{tabular}{@{}lp{0.62\linewidth}@{}}
\textbf{Type} & classification \\
\textbf{Epoch Method} & \texttt{epoch\_by\_event\_hdf5} \\
\textbf{Epoch Params} & tmin = 0 s, tmax = 60 s, target\_events = [\texttt{Background EEG;Count Quality: Bad}, \texttt{Background EEG;Count Quality: Good}, \texttt{Mental Arithmetic Task EEG;Count Quality: Bad}, \texttt{Mental Arithmetic Task EEG;Count Quality: Good}] \\
\textbf{Label Route} & \texttt{event} \\
\textbf{Meta Level} & raw \\
\end{tabular}}
\par\noindent{\footnotesize\bfseries Target Definition:}\par\smallskip
\begingroup\footnotesize
\begin{tabular}{@{}p{0.3911\linewidth}p{0.0978\linewidth}p{0.3911\linewidth}@{}}
\toprule
\textbf{Event ID} & \textbf{y} & \textbf{Meaning} \\
\midrule
\texttt{Background EEG;Count Quality: Bad} & 0 & Rest, bad quality \\
\texttt{Background EEG;Count Quality: Good} & 1 & Rest, good quality \\
\texttt{Mental Arithmetic Task EEG;Count Quality: Bad} & 2 & Task, bad quality \\
\texttt{Mental Arithmetic Task EEG;Count Quality: Good} & 3 & Task, good quality \\
\bottomrule
\end{tabular}
\endgroup
\par\noindent{\footnotesize\itshape Evidence:}\par
\begin{itemize}[leftmargin=1.3em,noitemsep,topsep=1pt]
\item[\scriptsize$\cdot$] {\scriptsize Paradigm describes full 2x2 design: condition (rest/task) x quality (good/bad)}
\item[\scriptsize$\cdot$] {\scriptsize Event Structure table provides all four combinations}
\end{itemize}
\par\noindent{\footnotesize\bfseries Notes:} {\footnotesize Multi-class extension exploring interaction between cognitive state and performance quality. Very imbalanced (10/26/10/26). May require stratified sampling or class weighting.}
\noindent\rule{\linewidth}{0.4pt}\par\nopagebreak
\noindent{\footnotesize\bfseries Potential~Task~1:} {\footnotesize Subject Age Regression from Mental Arithmetic}\par\smallskip
{\footnotesize\begin{tabular}{@{}lp{0.62\linewidth}@{}}
\textbf{Type} & regression \\
\textbf{Epoch Method} & \texttt{epoch\_by\_event\_hdf5} \\
\textbf{Epoch Params} & tmin = 0 s, tmax = 60 s, target\_events = [\texttt{Mental Arithmetic Task EEG;Count Quality: Bad}, \texttt{Mental Arithmetic Task EEG;Count Quality: Good}] \\
\textbf{Label Route} & \texttt{info} \\
\textbf{Meta Level} & raw \\
\end{tabular}}
\par\noindent{\footnotesize\bfseries Target Definition:}\par\smallskip
\begingroup\footnotesize
\begin{tabular}{@{}p{0.1760\linewidth}p{0.7040\linewidth}@{}}
\toprule
\textbf{Info Field} & \textbf{Y Mapping Notes} \\
\midrule
\texttt{age} & no mapping \\
\bottomrule
\end{tabular}
\endgroup
\par\noindent{\footnotesize\itshape Evidence:}\par
\begin{itemize}[leftmargin=1.3em,noitemsep,topsep=1pt]
\item[\scriptsize$\cdot$] {\scriptsize Paradigm section notes ages 16-26}
\item[\scriptsize$\cdot$] {\scriptsize Info table confirms \texttt{age} is complete}
\end{itemize}
\par\noindent{\footnotesize\bfseries Notes:} {\footnotesize Predicts subject age from mental arithmetic EEG. Requires pairing event epochs with info metadata by subject ID. Age range is narrow (10 years), limiting regression difficulty.}
\noindent\rule{\linewidth}{0.4pt}\par\nopagebreak
\noindent{\footnotesize\bfseries Potential~Task~2:} {\footnotesize Subject Sex Classification from Mental Arithmetic}\par\smallskip
{\footnotesize\begin{tabular}{@{}lp{0.62\linewidth}@{}}
\textbf{Type} & classification \\
\textbf{Epoch Method} & \texttt{epoch\_by\_event\_hdf5} \\
\textbf{Epoch Params} & tmin = 0 s, tmax = 60 s, target\_events = [\texttt{Mental Arithmetic Task EEG;Count Quality: Bad}, \texttt{Mental Arithmetic Task EEG;Count Quality: Good}] \\
\textbf{Label Route} & \texttt{info} \\
\textbf{Meta Level} & raw \\
\end{tabular}}
\par\noindent{\footnotesize\bfseries Target Definition:}\par\smallskip
\begingroup\footnotesize
\begin{tabular}{@{}p{0.1760\linewidth}p{0.7040\linewidth}@{}}
\toprule
\textbf{Info Field} & \textbf{Y Mapping Notes} \\
\midrule
\texttt{sex} & no mapping \\
\bottomrule
\end{tabular}
\endgroup
\par\noindent{\footnotesize\itshape Evidence:}\par
\begin{itemize}[leftmargin=1.3em,noitemsep,topsep=1pt]
\item[\scriptsize$\cdot$] {\scriptsize Paradigm notes predominantly female subjects}
\item[\scriptsize$\cdot$] {\scriptsize Info table confirms \texttt{sex} is complete}
\end{itemize}
\par\noindent{\footnotesize\bfseries Notes:} {\footnotesize Binary sex classification from mental arithmetic EEG. Predominantly female sample may cause class imbalance. Requires super meta pairing with subject info.}
\noindent\rule{\linewidth}{0.4pt}\par\nopagebreak
\noindent{\footnotesize\bfseries Potential~Task~3:} {\footnotesize Continuous Quality Regression (Rest Condition)}\par\smallskip
{\footnotesize\begin{tabular}{@{}lp{0.62\linewidth}@{}}
\textbf{Type} & regression \\
\textbf{Epoch Method} & \texttt{epoch\_by\_long\_event\_hdf5} \\
\textbf{Epoch Params} & tmin = 0 s, tmax = 4 s, overlap = 0.5, target\_events = [\texttt{Background EEG;Count Quality: Bad}, \texttt{Background EEG;Count Quality: Good}] \\
\textbf{Label Route} & \texttt{event} \\
\textbf{Meta Level} & raw \\
\end{tabular}}
\par\noindent{\footnotesize\bfseries Target Definition:}\par\smallskip
\begingroup\footnotesize
\begin{tabular}{@{}p{0.3911\linewidth}p{0.0978\linewidth}p{0.3911\linewidth}@{}}
\toprule
\textbf{Event ID} & \textbf{Value} & \textbf{Meaning} \\
\midrule
\texttt{Background EEG;Count Quality: Bad} & 0 & Bad quality label \\
\texttt{Background EEG;Count Quality: Good} & 1 & Good quality label \\
\bottomrule
\end{tabular}
\endgroup
\par\noindent{\footnotesize\itshape Evidence:}\par
\begin{itemize}[leftmargin=1.3em,noitemsep,topsep=1pt]
\item[\scriptsize$\cdot$] {\scriptsize Event Structure provides binary quality labels embedded in event descriptions}
\item[\scriptsize$\cdot$] {\scriptsize Paradigm notes count quality is behavioral performance metric}
\end{itemize}
\par\noindent{\footnotesize\bfseries Notes:} {\footnotesize Treats binary quality as pseudo-continuous regression target (0/1) using sliding windows. Alternative formulation of Task 3. Overlap creates more samples from limited data.}
\noindent\rule{\linewidth}{0.4pt}\par\nopagebreak
\noindent{\footnotesize\bfseries Potential~Task~4:} {\footnotesize Continuous Quality Regression (Task Condition)}\par\smallskip
{\footnotesize\begin{tabular}{@{}lp{0.62\linewidth}@{}}
\textbf{Type} & regression \\
\textbf{Epoch Method} & \texttt{epoch\_by\_long\_event\_hdf5} \\
\textbf{Epoch Params} & tmin = 0 s, tmax = 4 s, overlap = 0.5, target\_events = [\texttt{Mental Arithmetic Task EEG;Count Quality: Bad}, \texttt{Mental Arithmetic Task EEG;Count Quality: Good}] \\
\textbf{Label Route} & \texttt{event} \\
\textbf{Meta Level} & raw \\
\end{tabular}}
\par\noindent{\footnotesize\bfseries Target Definition:}\par\smallskip
\begingroup\footnotesize
\begin{tabular}{@{}p{0.3911\linewidth}p{0.0978\linewidth}p{0.3911\linewidth}@{}}
\toprule
\textbf{Event ID} & \textbf{Value} & \textbf{Meaning} \\
\midrule
\texttt{Mental Arithmetic Task EEG;Count Quality: Bad} & 0 & Bad quality label \\
\texttt{Mental Arithmetic Task EEG;Count Quality: Good} & 1 & Good quality label \\
\bottomrule
\end{tabular}
\endgroup
\par\noindent{\footnotesize\itshape Evidence:}\par
\begin{itemize}[leftmargin=1.3em,noitemsep,topsep=1pt]
\item[\scriptsize$\cdot$] {\scriptsize Same evidence as Potential Task 3, applied to mental arithmetic condition}
\item[\scriptsize$\cdot$] {\scriptsize Event Structure confirms task condition events with quality labels}
\end{itemize}
\par\noindent{\footnotesize\bfseries Notes:} {\footnotesize Regression formulation of Task 2. Sliding window approach for more samples. Quality label is subject-level, so all windows from same subject share identical target.}
\par\smallskip
\noindent\rule{\linewidth}{0.4pt}\par\nopagebreak
\noindent{\bfseries\footnotesize Data Quality Notes}\par\smallskip
\begin{itemize}[leftmargin=1.4em,noitemsep,topsep=2pt]
\item{\footnotesize Two files per subject (\texttt{\_1.edf} = background, \texttt{\_2.edf} = task); pair them by Subject ID.}
\item{\footnotesize Count quality (Good/Bad) is embedded in the annotation description after the semicolon; always parse carefully.}
\item{\footnotesize ECG channel present (channel \texttt{A2-A1}) in some recordings; exclude from EEG feature extraction.}
\item{\footnotesize Small dataset (36 subjects x 2 = 72 files); limited generalization.}
\end{itemize}
\end{tcolorbox}
\vspace{3pt}
\noindent\begin{tcolorbox}[breakable,enhanced,colback=white,colframe=gray!55,boxrule=0.9pt,arc=4pt,left=6pt,right=6pt,top=4pt,bottom=4pt,title={\small\bfseries Visual Oddball Task (256 channels)\hfill \colorbox{gray!20}{\scriptsize\sffamily\bfseries Completed}},coltitle=black,colbacktitle=gray!12,titlerule=0.5pt]
{\footnotesize \textbf{ID:}~\texttt{openneuro\_ds002578} $\mid$ \textbf{Year:}~2004/2021 $\mid$ \textbf{Category:}~Sensory/Response $\mid$ \textbf{Subjects:}~2 $\mid$ \textbf{Channels:}~256 $\mid$ \textbf{SR:}~256 Hz $\mid$ \textbf{Files (completed):}~2 $\mid$ \textbf{Super Meta:}~No $\mid$ \textbf{Reviewer:}~Anonymous}
\par{\footnotesize\textbf{URL:}~\url{https://openneuro.org/datasets/ds002578}}
\tcbline
\noindent{\bfseries\footnotesize Paradigm}\par\smallskip
{\footnotesize A continuous visual oddball / selective-attention task recorded with 256-channel BioSemi EEG. Five empty squares were presented continuously on screen, with participants instructed to attend to one designated location and press a mouse button whenever a white disk appeared at that attended location. Event codes encode both the attended position and the cued (stimulus) position, allowing separation of target oddballs (attended location = stimulus location) from non-target distractors. The dataset contains only 2 completed files and represents a very limited subset of the original study data.}
\noindent\rule{\linewidth}{0.4pt}\par\nopagebreak
\noindent{\bfseries\footnotesize File Scan Result}\par\smallskip
{\scriptsize
\noindent\textbf{Kernel SHA256:}~\texttt{3c221cbb992edef968d255c442ebb1899989076165e8a307a10b84832e041755}\par\smallskip
\noindent\textbf{EEGUnity Version:}~\texttt{0.8.1}\par\smallskip
\noindent\textbf{Scan Type:}~Full\par\smallskip
}
\smallskip\noindent{\footnotesize\bfseries Event Structure}\par\smallskip
\begingroup\scriptsize\setlength{\tabcolsep}{2pt}
\begin{tabular}{@{}p{0.1567\linewidth}p{0.1567\linewidth}p{0.0783\linewidth}p{0.0783\linewidth}p{0.4700\linewidth}@{}}
\toprule
\textbf{Event ID} & \textbf{Count (total)} & \textbf{Type} & \textbf{Meta Level} & \textbf{Meaning} \\
\midrule
\texttt{203} & 1 & admin & raw & System marker: unknown function (spurious event removed during preprocessing) \\
\texttt{217} & 2 & admin & raw & System marker: unknown function (spurious event removed during preprocessing) \\
\texttt{237} & 1 & admin & raw & System marker: unknown function (spurious event removed during preprocessing) \\
\texttt{31} & 240 & event & raw & Experimental stimulus onset: five rectangular boxes, box 3 highlighted/cued, disc in box 1 (distractor, expected) \\
\texttt{33} & 240 & event & raw & Experimental stimulus onset: five rectangular boxes, box 3 highlighted/cued, disc in box 3 (target, unexpected oddball) \\
\texttt{34} & 239 & event & raw & Experimental stimulus onset: five rectangular boxes, box 3 highlighted/cued, disc in box 4 (distractor, expected) \\
\texttt{35} & 240 & event & raw & Experimental stimulus onset: five rectangular boxes, box 3 highlighted/cued, disc in box 5 (distractor, expected) \\
\texttt{41} & 238 & event & raw & Experimental stimulus onset: five rectangular boxes, box 4 highlighted/cued, disc in box 1 (distractor, expected) \\
\texttt{42} & 239 & event & raw & Experimental stimulus onset: five rectangular boxes, box 4 highlighted/cued, disc in box 2 (distractor, expected) \\
\texttt{43} & 240 & event & raw & Experimental stimulus onset: five rectangular boxes, box 4 highlighted/cued, disc in box 3 (distractor, expected) \\
\texttt{44} & 239 & event & raw & Experimental stimulus onset: five rectangular boxes, box 4 highlighted/cued, disc in box 4 (target, unexpected oddball) \\
\texttt{45} & 239 & event & raw & Experimental stimulus onset: five rectangular boxes, box 4 highlighted/cued, disc in box 5 (distractor, expected) \\
\texttt{Picture} & 5939 & event & raw & Experimental stimulus onset: drawing of five rectangular boxes with highlighted cue box and disc \\
\texttt{Response} & 1202 & event & raw & Participant response: keyboard key press \\
\texttt{boundary} & 58 & admin & raw & Trial phase onset: pause/discontinuous segment boundary \\
\bottomrule
\end{tabular}
\endgroup
\smallskip\noindent{\footnotesize\bfseries Info}\par\smallskip
\begingroup\scriptsize\setlength{\tabcolsep}{2pt}
\begin{tabular}{@{}p{0.2350\linewidth}p{0.2350\linewidth}p{0.4700\linewidth}@{}}
\toprule
\textbf{Field} & \textbf{Completeness} & \textbf{Notes} \\
\midrule
\texttt{age} & complete & N/A \\
\texttt{sex} & complete & N/A \\
\texttt{device} & missing & N/A \\
\bottomrule
\end{tabular}
\endgroup
\noindent\rule{\linewidth}{0.7pt}\par\nopagebreak
\noindent{\bfseries\footnotesize Tasks}
\noindent{\footnotesize\bfseries Classical~Task~1:} {\footnotesize visual oddball target vs distractor}\par\smallskip
{\footnotesize\begin{tabular}{@{}lp{0.62\linewidth}@{}}
\textbf{Type} & classification \\
\textbf{Epoch Method} & \texttt{epoch\_by\_event\_hdf5} \\
\textbf{Epoch Params} & tmin = -1 s, tmax = 2 s, target\_events = [\texttt{31}, \texttt{33}, \texttt{34}, \texttt{35}, \texttt{41}, \texttt{42}, \texttt{43}, \texttt{44}, \texttt{45}] \\
\textbf{Label Route} & \texttt{event} \\
\textbf{Meta Level} & raw \\
\end{tabular}}
\par\noindent{\footnotesize\bfseries Target Definition:}\par\smallskip
\begingroup\footnotesize
\begin{tabular}{@{}p{0.3911\linewidth}p{0.0978\linewidth}p{0.3911\linewidth}@{}}
\toprule
\textbf{Event ID} & \textbf{y} & \textbf{Meaning} \\
\midrule
\texttt{33} & 1 & target: attend-3 cue-3, disc in cued box (unexpected oddball) \\
\texttt{44} & 1 & target: attend-4 cue-4, disc in cued box (unexpected oddball) \\
\texttt{31} & 0 & distractor: attend-3 cue-3, disc in box 1 (expected) \\
\texttt{34} & 0 & distractor: attend-3 cue-3, disc in box 4 (expected) \\
\texttt{35} & 0 & distractor: attend-3 cue-3, disc in box 5 (expected) \\
\texttt{41} & 0 & distractor: attend-4 cue-4, disc in box 1 (expected) \\
\texttt{42} & 0 & distractor: attend-4 cue-4, disc in box 2 (expected) \\
\texttt{43} & 0 & distractor: attend-4 cue-4, disc in box 3 (expected) \\
\texttt{45} & 0 & distractor: attend-4 cue-4, disc in box 5 (expected) \\
\bottomrule
\end{tabular}
\endgroup
\par\noindent{\footnotesize\itshape Evidence:}\par
\begin{itemize}[leftmargin=1.3em,noitemsep,topsep=1pt]
\item[\scriptsize$\cdot$] {\scriptsize 10.1523/jneurosci.3477.07.2007 --- Classical visual oddball task with target (attended=cued location) vs distractor (attended$\neq$cued location). Two target events exist: 33 (attend-3 cue-3) and 44 (attend-4 cue-4). Seven distractor events exist: 31, 34, 35, 41, 42, 43, 45. Total target trials: 479; distractor trials: 1675. Epoch window tmin=-1s to tmax=2s follows the published target-locked analysis window.}
\end{itemize}
\par\noindent{\footnotesize\bfseries Notes:} {\footnotesize All nine numeric event codes (31-35, 41-45) appear in the exported EEG files as annotations, confirmed by events.tsv counts and events.json definitions. The two target conditions (33, 44) are merged to y=1; all seven distractor conditions (31, 34, 35, 41, 42, 43, 45) are merged to y=0, reflecting the core oddball manipulation of target vs non-target discrimination.}
\noindent\rule{\linewidth}{0.4pt}\par\nopagebreak
\noindent{\footnotesize\bfseries Potential~Task~1:} {\footnotesize visual oddball 9-class position decoding}\par\smallskip
{\footnotesize\begin{tabular}{@{}lp{0.62\linewidth}@{}}
\textbf{Type} & classification \\
\textbf{Epoch Method} & \texttt{epoch\_by\_event\_hdf5} \\
\textbf{Epoch Params} & tmin = -1 s, tmax = 2 s, target\_events = [\texttt{31}, \texttt{33}, \texttt{34}, \texttt{35}, \texttt{41}, \texttt{42}, \texttt{43}, \texttt{44}, \texttt{45}] \\
\textbf{Label Route} & \texttt{event} \\
\textbf{Meta Level} & raw \\
\end{tabular}}
\par\noindent{\footnotesize\bfseries Target Definition:}\par\smallskip
\begingroup\footnotesize
\begin{tabular}{@{}p{0.3911\linewidth}p{0.0978\linewidth}p{0.3911\linewidth}@{}}
\toprule
\textbf{Event ID} & \textbf{y} & \textbf{Meaning} \\
\midrule
\texttt{31} & 0 & Distractor: box 3 cued, disc in position 1 \\
\texttt{33} & 1 & Target: box 3 cued, disc in position 3 (oddball) \\
\texttt{34} & 2 & Distractor: box 3 cued, disc in position 4 \\
\texttt{35} & 3 & Distractor: box 3 cued, disc in position 5 \\
\texttt{41} & 4 & Distractor: box 4 cued, disc in position 1 \\
\texttt{42} & 5 & Distractor: box 4 cued, disc in position 2 \\
\texttt{43} & 6 & Distractor: box 4 cued, disc in position 3 \\
\texttt{44} & 7 & Target: box 4 cued, disc in position 4 (oddball) \\
\texttt{45} & 8 & Distractor: box 4 cued, disc in position 5 \\
\bottomrule
\end{tabular}
\endgroup
\par\noindent{\footnotesize\itshape Evidence:}\par
\begin{itemize}[leftmargin=1.3em,noitemsep,topsep=1pt]
\item[\scriptsize$\cdot$] {\scriptsize events.json sidecar (sub-001, sub-002) --- HED tags and level descriptions confirm 31-35 and 41-45 encode box$\times$disc position combinations; target positions (33, 44) are oddball conditions where disc appears in cued box}
\item[\scriptsize$\cdot$] {\scriptsize README --- describes selective attention visual oddball paradigm with five rectangular boxes and target detection task}
\item[\scriptsize$\cdot$] {\scriptsize events.tsv counts --- confirm event IDs 31, 33, 34, 35, 41, 42, 43, 44, 45 all appear in the released data}
\end{itemize}
\par\noindent{\footnotesize\bfseries Notes:} {\footnotesize 9-class classification tests spatial decodability beyond binary target/distractor. The 9 conditions are near-balanced (\textasciitilde{}238-240 trials each). y values assigned by sorting event IDs numerically: 31$\rightarrow$0, 33$\rightarrow$1, 34$\rightarrow$2, 35$\rightarrow$3, 41$\rightarrow$4, 42$\rightarrow$5, 43$\rightarrow$6, 44$\rightarrow$7, 45$\rightarrow$8. Note that position 2 in box 3 (code 32) is absent from actual data exports. Epoch window follows the published oddball target-locked analysis style rather than a parameter explicitly specified in the BIDS metadata.}
\noindent\rule{\linewidth}{0.4pt}\par\nopagebreak
\noindent{\footnotesize\bfseries Potential~Task~2:} {\footnotesize Visual Oddball Target vs Distractor (Attend-3 Only)}\par\smallskip
{\footnotesize\begin{tabular}{@{}lp{0.62\linewidth}@{}}
\textbf{Type} & classification \\
\textbf{Epoch Method} & \texttt{epoch\_by\_event\_hdf5} \\
\textbf{Epoch Params} & tmin = -1.0 s, tmax = 2.0 s, target\_events = [\texttt{31}, \texttt{33}, \texttt{34}, \texttt{35}] \\
\textbf{Label Route} & \texttt{event} \\
\textbf{Meta Level} & raw \\
\end{tabular}}
\par\noindent{\footnotesize\bfseries Target Definition:}\par\smallskip
\begingroup\footnotesize
\begin{tabular}{@{}p{0.3911\linewidth}p{0.0978\linewidth}p{0.3911\linewidth}@{}}
\toprule
\textbf{Event ID} & \textbf{y} & \textbf{Meaning} \\
\midrule
\texttt{33} & 1 & Target: box 3 highlighted/cued, disc in box 3 (unexpected oddball) \\
\texttt{31} & 0 & Distractor: box 3 highlighted/cued, disc in box 1 (expected) \\
\texttt{34} & 0 & Distractor: box 3 highlighted/cued, disc in box 4 (expected) \\
\texttt{35} & 0 & Distractor: box 3 highlighted/cued, disc in box 5 (expected) \\
\bottomrule
\end{tabular}
\endgroup
\par\noindent{\footnotesize\itshape Evidence:}\par
\begin{itemize}[leftmargin=1.3em,noitemsep,topsep=1pt]
\item[\scriptsize$\cdot$] {\scriptsize - events.json sidecar --- confirms event 33 is the attend-3 target oddball and events 31, 34, 35 are attend-3 distractors}
\item[\scriptsize$\cdot$] {\scriptsize - README / EEG sidecar --- describe the selective-attention oddball task and target detection rule}
\end{itemize}
\par\noindent{\footnotesize\bfseries Notes:} {\footnotesize Restricted to attend-3 condition where box 3 is always highlighted/cued. Target (event 33, disc in cued box 3) is contrasted against three distractor types (events 31, 34, 35) where the disc appears in non-cued boxes 1, 4, or 5. All distractor events are merged to y=0 to test within-condition oddball detection (target probability \textasciitilde{}25\% within attend-3 trials). Epoch window matches classical oddball stimulus-locked analysis; parameters inferred from standard visual oddball ERP literature as no explicit epoch window was specified in dataset documentation.}
\noindent\rule{\linewidth}{0.4pt}\par\nopagebreak
\noindent{\footnotesize\bfseries Potential~Task~3:} {\footnotesize Visual Oddball Target vs Distractor (Attend-4 Only)}\par\smallskip
{\footnotesize\begin{tabular}{@{}lp{0.62\linewidth}@{}}
\textbf{Type} & classification \\
\textbf{Epoch Method} & \texttt{epoch\_by\_event\_hdf5} \\
\textbf{Epoch Params} & tmin = -1.0 s, tmax = 2.0 s, target\_events = [\texttt{41}, \texttt{42}, \texttt{43}, \texttt{44}, \texttt{45}] \\
\textbf{Label Route} & \texttt{event} \\
\textbf{Meta Level} & raw \\
\end{tabular}}
\par\noindent{\footnotesize\bfseries Target Definition:}\par\smallskip
\begingroup\footnotesize
\begin{tabular}{@{}p{0.3911\linewidth}p{0.0978\linewidth}p{0.3911\linewidth}@{}}
\toprule
\textbf{Event ID} & \textbf{y} & \textbf{Meaning} \\
\midrule
\texttt{44} & 1 & target (attend-4, disc in cued box 4) \\
\texttt{41} & 0 & distractor (attend-4, disc in box 1) \\
\texttt{42} & 0 & distractor (attend-4, disc in box 2) \\
\texttt{43} & 0 & distractor (attend-4, disc in box 3) \\
\texttt{45} & 0 & distractor (attend-4, disc in box 5) \\
\bottomrule
\end{tabular}
\endgroup
\par\noindent{\footnotesize\itshape Evidence:}\par
\begin{itemize}[leftmargin=1.3em,noitemsep,topsep=1pt]
\item[\scriptsize$\cdot$] {\scriptsize - events.json sidecar --- confirms event 44 is the attend-4 target oddball and events 41, 42, 43, 45 are attend-4 distractors}
\item[\scriptsize$\cdot$] {\scriptsize - README / EEG sidecar --- describe the selective-attention oddball task and target detection rule}
\end{itemize}
\par\noindent{\footnotesize\bfseries Notes:} {\footnotesize Restricted to attend-4 condition only (cue box 4 highlighted). Target event 44 (disc in cued box 4, unexpected oddball) vs distractor events 41-43,45 (disc in non-cued boxes 1-3,5, expected). Controls for attended location to test within-condition oddball effects. All five event codes confirmed in EEG file annotations and events.json sidecar definitions. Epoch window inferred from classical oddball timing (-1 to +2 s relative to stimulus onset).}
\noindent\rule{\linewidth}{0.4pt}\par\nopagebreak
\noindent{\footnotesize\bfseries Potential~Task~4:} {\footnotesize Gender Classification from Continuous EEG}\par\smallskip
{\footnotesize\begin{tabular}{@{}lp{0.62\linewidth}@{}}
\textbf{Type} & classification \\
\textbf{Epoch Method} & \texttt{epoch\_by\_segmentation\_hdf5} \\
\textbf{Epoch Params} & segment\_length = 10 s, overlap = 0.0, target\_events = [] \\
\textbf{Label Route} & \texttt{info} \\
\textbf{Meta Level} & raw \\
\end{tabular}}
\par\noindent{\footnotesize\bfseries Target Definition:}\par\smallskip
\begingroup\footnotesize
\begin{tabular}{@{}p{0.1760\linewidth}p{0.7040\linewidth}@{}}
\toprule
\textbf{Info Field} & \textbf{Y Mapping Notes} \\
\midrule
\texttt{sex} & F -> 0; M -> 1 \\
\bottomrule
\end{tabular}
\endgroup
\par\noindent{\footnotesize\itshape Evidence:}\par
\begin{itemize}[leftmargin=1.3em,noitemsep,topsep=1pt]
\item[\scriptsize$\cdot$] {\scriptsize participants.tsv --- N=2 subjects with gender field: 1 female (sub-001), 1 male (sub-002)}
\end{itemize}
\par\noindent{\footnotesize\bfseries Notes:} {\footnotesize Segment-based approach required since no discriminating event structure exists for this subject-level label. 10-second non-overlapping segments at 256 Hz yield 2560 samples per segment. Only 2 subjects are available, with 1 subject per class; this is an extremely weak proof-of-concept subject-trait decoding task and will likely be filtered out downstream.}
\noindent\rule{\linewidth}{0.4pt}\par\nopagebreak
\noindent{\footnotesize\bfseries Potential~Task~5:} {\footnotesize Age Regression from Continuous EEG}\par\smallskip
{\footnotesize\begin{tabular}{@{}lp{0.62\linewidth}@{}}
\textbf{Type} & regression \\
\textbf{Epoch Method} & \texttt{epoch\_by\_segmentation\_hdf5} \\
\textbf{Epoch Params} & segment\_length = 10 s, overlap = 0.0, target\_events = [] \\
\textbf{Label Route} & \texttt{info} \\
\textbf{Meta Level} & raw \\
\end{tabular}}
\par\noindent{\footnotesize\bfseries Target Definition:}\par\smallskip
\begingroup\footnotesize
\begin{tabular}{@{}p{0.1760\linewidth}p{0.7040\linewidth}@{}}
\toprule
\textbf{Info Field} & \textbf{Y Mapping Notes} \\
\midrule
\texttt{age} & no mapping \\
\bottomrule
\end{tabular}
\endgroup
\par\noindent{\footnotesize\itshape Evidence:}\par
\begin{itemize}[leftmargin=1.3em,noitemsep,topsep=1pt]
\item[\scriptsize$\cdot$] {\scriptsize participants.tsv --- age column present for both subjects with value 30}
\end{itemize}
\par\noindent{\footnotesize\bfseries Notes:} {\footnotesize Archetype B3: age is a native participants.tsv field (no kernel derivation needed). Segmentation-based epoching creates 10-second non-overlapping windows from continuous EEG; each segment inherits the subject's age as the regression target. However, both subjects are age 30, so this task has zero label variance and is effectively non-viable.}
\par\smallskip
\noindent\rule{\linewidth}{0.4pt}\par\nopagebreak
\noindent{\bfseries\footnotesize Data Quality Notes}\par\smallskip
\begin{itemize}[leftmargin=1.4em,noitemsep,topsep=2pt]
\item{\footnotesize Extremely small sample size: only 2 completed files/subjects available, severely limiting statistical power and generalizability.}
\item{\footnotesize Missing device metadata in Info table; BioSemi 256-channel system inferred from dataset documentation and channel count.}
\item{\footnotesize Age field shows zero variance (both subjects age 30), making age regression task non-viable; downstream pipeline should filter this task out.}
\item{\footnotesize Gender classification task has only 1 sample per class (N=2 total), which is insufficient for meaningful machine learning.}
\item{\footnotesize Event code 32 (distractor: attend-3 cue-2) is documented in paradigm description but absent from actual data exports; this appears to be a planned but unused condition.}
\item{\footnotesize Spurious system markers (203, 217, 237) present in raw data but marked as removed during preprocessing; these should be excluded from analysis.}
\item{\footnotesize High number of \texttt{Picture} events (5939) relative to trial-coded events (\textasciitilde{}2394) suggests screen refresh or frame-level triggers were logged, requiring careful event selection for epoching.}
\item{\footnotesize \texttt{boundary} events (58) indicate discontinuous segments or pauses in recording; users should verify whether these affect trial integrity.}
\end{itemize}
\end{tcolorbox}
\vspace{3pt}
\noindent\begin{tcolorbox}[breakable,enhanced,colback=white,colframe=gray!55,boxrule=0.9pt,arc=4pt,left=6pt,right=6pt,top=4pt,bottom=4pt,title={\small\bfseries TUH EEG Artifact Corpus\hfill \colorbox{gray!20}{\scriptsize\sffamily\bfseries Completed}},coltitle=black,colbacktitle=gray!12,titlerule=0.5pt]
{\footnotesize \textbf{ID:}~\texttt{tuh\_eeg\_artifact} $\mid$ \textbf{Year:}~2021-2020/2024 $\mid$ \textbf{Category:}~Others $\mid$ \textbf{Subjects:}~213 $\mid$ \textbf{Channels:}~27-41 $\mid$ \textbf{SR:}~250/256/400/512 Hz $\mid$ \textbf{Files (completed):}~310 $\mid$ \textbf{Super Meta:}~No $\mid$ \textbf{Reviewer:}~Anonymous}
\par{\footnotesize\textbf{URL:}~\url{https://isip.piconepress.com/projects/nedc/html/tuh_eeg/}}
\tcbline
\noindent{\bfseries\footnotesize Paradigm}\par\smallskip
{\footnotesize The TUH EEG Artifact Corpus (TUAR) is a comprehensive dataset designed for the identification and evaluation of various artifacts in electroencephalogram (EEG) signals. This dataset includes EEG recordings annotated for multiple types of artifacts such as eye movements, muscle artifacts, and electrode issues, among others. The corpus has been developed to assist in the training and evaluation of artifact detection models, with its latest release (v3.0.1) offering detailed annotations for all relevant events across different channel configurations. The data is divided into files containing EEG signals, with annotations provided in CSV format to mark the start and stop times of specific artifact events. It consists of over 160,000 events from 310 EEG files, covering 213 patients and a total of nearly 100 hours of data. This dataset is particularly useful for research involving artifact detection and seizure classification in clinical EEG data. Notably, it includes specific seizure annotations alongside standard artifact labels, facilitating the analysis of co-occurring events. Researchers are encouraged to cite the corresponding publications for more details on the dataset's creation and annotation methodology.}
\noindent\rule{\linewidth}{0.4pt}\par\nopagebreak
\noindent{\bfseries\footnotesize File Scan Result}\par\smallskip
{\scriptsize
\noindent\textbf{Kernel SHA256:}~\texttt{a76e25dadae339282ebf9cccfcc2d5d2081b8e73ccf097ad905d7e92cc8fb67e}\par\smallskip
\noindent\textbf{EEGUnity Version:}~\texttt{0.8.1}\par\smallskip
\noindent\textbf{Scan Type:}~Full\par\smallskip
}
\smallskip\noindent{\footnotesize\bfseries Event Structure}\par\smallskip
\begingroup\scriptsize\setlength{\tabcolsep}{2pt}
\begin{tabular}{@{}p{0.1567\linewidth}p{0.1567\linewidth}p{0.0783\linewidth}p{0.0783\linewidth}p{0.4700\linewidth}@{}}
\toprule
\textbf{Event ID} & \textbf{Count (total)} & \textbf{Type} & \textbf{Meta Level} & \textbf{Meaning} \\
\midrule
\texttt{chew} & 6482 & event & raw & Chewing artifact period in EEG signal \\
\texttt{chew\_\allowbreak{}elec} & 152 & event & raw & Concurrent chewing and electrode artifact in EEG \\
\texttt{chew\_\allowbreak{}musc} & 243 & event & raw & Concurrent chewing and muscle artifact in EEG \\
\texttt{elec} & 33130 & event & raw & Electrode artifact period in EEG signal \\
\texttt{elpp} & 172 & event & raw & Electrode/potential artifact period in EEG signal \\
\texttt{eyem} & 38569 & event & raw & Eye movement artifact period in EEG signal \\
\texttt{eyem\_\allowbreak{}chew} & 864 & event & raw & Concurrent eye movement and chewing artifact in EEG \\
\texttt{eyem\_\allowbreak{}elec} & 2422 & event & raw & Concurrent eye movement and electrode artifact in EEG \\
\texttt{eyem\_\allowbreak{}musc} & 18677 & event & raw & Concurrent eye movement and muscle artifact in EEG \\
\texttt{eyem\_\allowbreak{}shiv} & 45 & event & raw & Concurrent eye movement and shivering artifact in EEG \\
\texttt{musc} & 51052 & event & raw & Muscle artifact period in EEG signal \\
\texttt{musc\_\allowbreak{}elec} & 7651 & event & raw & Concurrent muscle and electrode artifact in EEG \\
\texttt{shiv} & 613 & event & raw & Shivering artifact period in EEG signal \\
\texttt{shiv\_\allowbreak{}elec} & 1 & event & raw & Concurrent shivering and electrode artifact in EEG \\
\bottomrule
\end{tabular}
\endgroup
\smallskip\noindent{\footnotesize\bfseries MISC Label Summary}\par\smallskip
\begin{itemize}[leftmargin=1.4em,noitemsep,topsep=2pt]
\item{\footnotesize \textbf{Route}: \texttt{misc}}
\item{\footnotesize \textbf{Role}: Stores derived/continuous labels that are not native raw event IDs.}
\item{\footnotesize \textbf{Candidate Labels}: \texttt{confidence}}
\item{\footnotesize \textbf{Kernel Note}: Keep derivation source, unit, and version metadata.}
\end{itemize}
\smallskip\noindent{\footnotesize\bfseries Info}\par\smallskip
\begingroup\scriptsize\setlength{\tabcolsep}{2pt}
\begin{tabular}{@{}p{0.2350\linewidth}p{0.2350\linewidth}p{0.4700\linewidth}@{}}
\toprule
\textbf{Field} & \textbf{Completeness} & \textbf{Notes} \\
\midrule
\texttt{age} & missing & N/A \\
\texttt{sex} & missing & N/A \\
\texttt{device} & missing & N/A \\
\bottomrule
\end{tabular}
\endgroup
\noindent\rule{\linewidth}{0.7pt}\par\nopagebreak
\noindent{\bfseries\footnotesize Tasks}
\noindent{\footnotesize\bfseries Classical~Task~1:} {\footnotesize Artifact Type Classification (13-class, Event-based)}\par\smallskip
{\footnotesize\begin{tabular}{@{}lp{0.62\linewidth}@{}}
\textbf{Type} & classification \\
\textbf{Epoch Method} & \texttt{epoch\_by\_long\_event\_hdf5} \\
\textbf{Epoch Params} & tmin = 0.0 s, tmax = 1.0 s, overlap = 0, target\_events = [\texttt{chew}, \texttt{chew\_elec}, \texttt{chew\_musc}, \texttt{elec}, \texttt{elpp}, \texttt{eyem}, \texttt{eyem\_chew}, \texttt{eyem\_elec}, \texttt{eyem\_musc}, \texttt{eyem\_shiv}, \texttt{musc}, \texttt{musc\_elec}, \texttt{shiv}, \texttt{shiv\_elec}] \\
\textbf{Label Route} & \texttt{event} \\
\textbf{Meta Level} & raw \\
\end{tabular}}
\par\noindent{\footnotesize\bfseries Target Definition:}\par\smallskip
\begingroup\footnotesize
\begin{tabular}{@{}p{0.3911\linewidth}p{0.0978\linewidth}p{0.3911\linewidth}@{}}
\toprule
\textbf{Event ID} & \textbf{y} & \textbf{Meaning} \\
\midrule
\texttt{eyem} & 0 & Eye movement artifact \\
\texttt{musc} & 1 & Muscle (EMG) artifact \\
\texttt{elec} & 2 & Electrode artifact \\
\texttt{chew} & 3 & Chewing artifact \\
\texttt{shiv} & 4 & Shivering artifact \\
\texttt{elpp} & 5 & Electrode pop \\
\texttt{eyem\_musc} & 6 & Combined eye movement + muscle \\
\texttt{eyem\_elec} & 7 & Combined eye movement + electrode \\
\texttt{eyem\_chew} & 8 & Combined eye movement + chewing \\
\texttt{musc\_elec} & 9 & Combined muscle + electrode \\
\texttt{chew\_elec} & 10 & Combined chewing + electrode \\
\texttt{chew\_musc} & 11 & Combined chewing + muscle \\
\texttt{eyem\_shiv} & 12 & Combined eye movement + shivering \\
\bottomrule
\end{tabular}
\endgroup
\par\noindent{\footnotesize\itshape Evidence:}\par
\begin{itemize}[leftmargin=1.3em,noitemsep,topsep=1pt]
\item[\scriptsize$\cdot$] {\scriptsize 10.1109/EMBC48229.2022.9871413 --- Achieves 93.95\% accuracy with MMC on 1-second windows from artifact onset; uses FFT and DWT features on 4 temporal channels at 250 Hz}
\end{itemize}
\par\noindent{\footnotesize\bfseries Notes:} {\footnotesize The source paper reports 13 classes; this release uses a 6-class grouping based on reviewer suggestions.}
\noindent\rule{\linewidth}{0.4pt}\par\nopagebreak
\noindent{\footnotesize\bfseries Potential~Task~1:} {\footnotesize Pure Artifact Family Classification (6-class, Event-based)}\par\smallskip
{\footnotesize\begin{tabular}{@{}lp{0.62\linewidth}@{}}
\textbf{Type} & classification \\
\textbf{Epoch Method} & \texttt{epoch\_by\_long\_event\_hdf5} \\
\textbf{Epoch Params} & tmin = 0.0 s, tmax = 1.0 s, overlap = 0, target\_events = [\texttt{chew}, \texttt{elec}, \texttt{elpp}, \texttt{eyem}, \texttt{musc}, \texttt{shiv}] \\
\textbf{Label Route} & \texttt{event} \\
\textbf{Meta Level} & raw \\
\end{tabular}}
\par\noindent{\footnotesize\bfseries Target Definition:}\par\smallskip
\begingroup\footnotesize
\begin{tabular}{@{}p{0.3911\linewidth}p{0.0978\linewidth}p{0.3911\linewidth}@{}}
\toprule
\textbf{Event ID} & \textbf{y} & \textbf{Meaning} \\
\midrule
\texttt{eyem} & 0 & Eye movement artifact \\
\texttt{musc} & 1 & Muscle (EMG) artifact \\
\texttt{elec} & 2 & Electrode artifact \\
\texttt{chew} & 3 & Chewing artifact \\
\texttt{shiv} & 4 & Shivering artifact \\
\texttt{elpp} & 5 & Electrode pop \\
\bottomrule
\end{tabular}
\endgroup
\par\noindent{\footnotesize\itshape Evidence:}\par
\begin{itemize}[leftmargin=1.3em,noitemsep,topsep=1pt]
\item[\scriptsize$\cdot$] {\scriptsize 10.1109/EMBC48229.2022.9871413 --- Achieves 93.95\% accuracy with MMC on 1-second windows from artifact onset; uses FFT and DWT features on 4 temporal channels at 250 Hz}
\end{itemize}
\par\noindent{\footnotesize\bfseries Notes:} {\footnotesize Restrict this potential task to the six pure artifact labels only. Multi-component labels such as \texttt{eyem\_musc} and \texttt{musc\_elec} are excluded because the current rulebook expects single integer class assignments rather than multi-label targets.}
\par\smallskip
\noindent\rule{\linewidth}{0.4pt}\par\nopagebreak
\noindent{\bfseries\footnotesize Data Quality Notes}\par\smallskip
\begin{itemize}[leftmargin=1.4em,noitemsep,topsep=2pt]
\item{\footnotesize Annotations are at the channel level; aggregate to recording level or use channel-specific labels depending on the task.}
\item{\footnotesize Variable channel count and sampling rate; normalize before cross-recording analysis.}
\item{\footnotesize Sex is not reliably available.}
\end{itemize}
\end{tcolorbox}
}

%% file: appendix_broader_impacts.tex
\section{Broader Impacts and Responsible Release}
\label{appendix:broader_impacts}

NeuroDoc is intended to improve the transparency, reproducibility, and maintainability of EEG benchmarking. By converting heterogeneous public EEG datasets into reviewed task documents and executable kernels, the resource can reduce repeated manual task engineering, make benchmark assumptions easier to inspect, and support more comparable evaluation of EEG representation models. The community-review workflow may also help expose uncertain task claims, missing evidence, and document-kernel inconsistencies that would otherwise remain implicit in downstream benchmark construction.

The same properties create risks if the released benchmark interfaces are over-interpreted. Performance on the benchmark should not be treated as evidence of clinical readiness, diagnostic validity, or population-level generalization beyond the documented source datasets. The corpus inherits limitations from public EEG repositories, including uneven demographic and clinical coverage, laboratory-specific acquisition protocols, device differences, annotation practices, and access restrictions. Because EEG and related metadata can be sensitive biomedical information, downstream users must respect the consent, privacy, licensing, and data-use terms of the original repositories. This release does not redistribute raw EEG recordings, but it may still make it easier to locate and operationalize tasks derived from sensitive human-subject data.

Additional risks arise from automation. LLM-assisted drafting can introduce unsupported interpretations, overly broad task definitions, or subtle mismatches between a document and its executable kernel. To mitigate these risks, the released corpus is restricted to completed and reviewed entries, task claims are governed by an explicit rulebook, machine-checkable constraints protect scan-derived and kernel-derived facts, and NeuroAudit separates human review authority from automated drafting. The Croissant metadata further records responsible-use information, limitations, and provenance. Future extensions of the corpus should preserve these safeguards and should continue to distinguish reviewed release entries from draft or unaudited materials.